\documentclass[11pt]{article} 
\PassOptionsToPackage{table}{xcolor}
\usepackage[preprint]{acl}
\usepackage{booktabs}
\usepackage{pifont}
\usepackage{makecell}
\providecommand{\checkmark}{\ding{51}}
\providecommand{\cmark}{\checkmark}

\usepackage{float}
\usepackage{hyperref}
\usepackage{url}
\usepackage{graphicx}
\usepackage{enumitem}
\usepackage{subcaption} 
\usepackage[most]{tcolorbox}
\usepackage{amsmath}
\usepackage{amssymb}
\usepackage{tabularx}      
\newcolumntype{C}{>{\centering\arraybackslash}X} 
\usepackage{soul}
\sethlcolor{yellow}
\usepackage{ifthen}
\newboolean{showrebuttal}
\setboolean{showrebuttal}{false}  
\newcommand{\rebuttal}[1]{%
  \ifthenelse{\boolean{showrebuttal}}%
    {\hl{#1}}
    {}
}

\title{Mind's Eye: A Benchmark of Visual Abstraction, Transformation and Composition for Multimodal LLMs}

\author{
  Rohit Sinha\thanks{Work done at Microsoft Research India} \\
  CSE Dept., IIT Hyderabad \\
  Hyderabad, India \\
  \texttt{rohit.sinha@prjt.cse.iith.ac.in} \\\And
  Aditya Kanade \\
  Microsoft Research \\
  Bengaluru, India \\
  \texttt{kanade850@gmail.com} \\\AND
  Sai Srinivas Kancheti\footnotemark[1] \\
  CSE Dept., IIT Hyderabad \\
  Hyderabad, India \\
  \texttt{cs21resch01004@iith.ac.in} \\\And
  Vineeth N Balasubramanian\thanks{Corresponding author} \\
  Microsoft Research \\
  Bengaluru, India \\
  \texttt{vineeth.nb@microsoft.com} \\\AND
  Tanuja Ganu\footnotemark[2] \\
  Microsoft Research \\
  Bengaluru, India \\
  \texttt{tanuja.ganu@microsoft.com} \\
}

\begin{document}
\maketitle


\begin{abstract}

Multimodal large language models (MLLMs) have achieved impressive progress on vision language benchmarks, yet their capacity for visual cognitive and visuospatial reasoning remains less understood. We introduce \textsc{Mind’s Eye}, a multiple-choice benchmark of eight visuo-cognitive tasks inspired by classic human intelligence tests and organized under a novel \textbf{A–R–T} taxonomy: \textit{Abstraction}, \textit{Relation}, and \textit{Transformation}. The tasks probe core processes of fluid intelligence such as pattern induction, analogical relation mapping, and mental transformation. We evaluate a diverse suite of closed-source and open-source MLLMs and compare their performance with human participants. Humans achieve  \textbf{80\%} accuracy, while top performing MLLMs remain below \textbf{50\%}. Error analysis reveals failures in: (i) visual attention allocation, (ii) internal perceptual manipulation, and (iii) weak abstraction of underlying visual concepts. Our findings suggest that current MLLMs exhibit limited visuospatial reasoning capabilities, when compared with human participants, highlighting the need for more cognitively grounded evaluation frameworks\footnote{Code and Benchmark are available at: \url{https://github.com/microsoft/Mind-s-Eye}}.

\end{abstract}

\section{Introduction}

Multimodal Large Language Models (MLLMs) have demonstrated compelling visual understanding in recent years: identifying objects, reading text, or describing spatial relationships in presented scenes \citep{li2023blip2}. These tasks primarily test whether models can encode visual inputs and map them to linguistic outputs. As MLLMs become stronger, it becomes imperative to study their performance on increasingly complex visual tasks that are natural to humans. To this end, visuospatial transformation tasks require such models to generate novel visual states not present in the input such as mentally rotating a 3D object, predicting how a surface unfolds, or visualizing shape compositions. Such transformations form the core of spatial reasoning but demand a capability beyond perceptual encoding: the construction and manipulation of implicit spatial representations \citep{ vandenberg1978mental}. The capability of MLLMs to possess such generative spatial understanding remains an open empirical question.

Existing evaluations of MLLMs can be broadly categorized into three kinds: broad evaluations that test surface perception and prioritize scale \citep{li2023mmbench, yue2023mmmu}, synthetic diagnostics that probe compositional reasoning through pattern matching (rather than mental simulation) \citep{zhang2019raven}, and cognitive benchmarks that test spatial rule learning \citep{ chollet2025arcagi2newchallengefrontier}. We point out two key gaps across these studies: (1) Firstly, existing evaluations do not isolate and study visuospatial transformation, the capability (that comes to humans naturally) to mentally rotate, fold, or recompose shapes \citep{shepard1971mental, fleuret2011svrt}; and (2) Most existing studies often conflate visual evidence with linguistic priors, leaving it unclear on whether models reason from images or exploit language shortcuts\citep{suhr2019nlvr2}.

Keeping these in mind, we herein introduce \textsc{Mind's Eye}, a cognitively grounded benchmark derived from classic cognitive psychology tests such as mental rotation and paper folding \citep{ Ekstrom1976ManualFK}. Eight visuocognitive tasks are organized under an Abstraction-Relation-Transformation (ART) taxonomy: Abstraction tests study pattern induction, Relation tests study analogical mapping, Transformation tests study mental manipulation of shapes. Our generation process allows us to isolate visuospatial reasoning from world knowledge and linguistic priors. Each item includes diagnostic distractors targeting specific error types, enabling fine-grained analysis of where and why models fail. In particular, we organize our investigation around three questions: (1) \textit{How do MLLMs perform and compare with human performance on controlled visuospatial tests?} (2) \textit{Which cognitive factors drive the largest deficits in the performance of MLLMs on these tasks?} (3) \textit{Can prompting interventions help improve the performance of MLLMs on the considered tasks, or do failures reflect general model limitations?}


Evaluation of 18 MLLMs on \textsc{Mind's Eye} reveals significant underperformance relative to humans in visuospatial reasoning. Humans average 80\% accuracy across tasks, while the best models remain below 50\%. The largest deficits appear on Transformation and Abstraction tasks, both of which require mental simulation rather than surface pattern matching. Notably, while human accuracy degrades from easy to hard instances, MLLM performance remains flat across difficulty levels, indicating the absence of foundational visuo-cognitive operations rather than mere struggles with complexity. Prompting interventions yield dimension-dependent effects: structured scaffolding benefits Abstraction tasks but consistently impairs Transformation performance, suggesting that prompting facilitates rule derivation yet fails to elicit procedural visuospatial operations. Attention analysis further reveals that while models can localize relevant answer regions, they fail to reason reliably over this information—they identify where to look but not how to reason over what they see.

Our work makes the following contributions:

\begin{itemize}
    \item A new benchmark, \textsc{Mind's Eye}, for visuo-cognitive understanding of MLLMs, grounded in cognitive constructs of Abstraction-Relation-Transformation (ART) which includes diagnostic distractors.
    
    \item Evaluation of 18 MLLMs on the benchmark and comparison with a human baseline; the study includes prompting strategies, as well as fine-tuning and reinforcement learning-based alignment on a strong open-source model.
    
    \item Diagnostic analyses showing attention misalignment, difficulty-invariant failure patterns, and reasoning trace errors.
\end{itemize}

\section{Related Work}
\label{related_work}
\vspace{-3pt}

\begin{table*}[t]
\caption{\textbf{Closest benchmarks vs.\ \textsc{Mind's Eye} along diagnostic axes}: A comparative evaluation of \textsc{Mind's Eye} against other benchmarks on key diagnostic criteria. The table highlights the unique features that make \textsc{Mind's Eye} a more controlled and cognitively grounded diagnostic tool for assessing fluid visuospatial intelligence. 
\ding{51}=explicit support; \ding{114}=partial; \ding{55}=absent.}
\resizebox{\textwidth}{!}{
\begin{tabular}{lcccccc}
\toprule
Dataset & Formal Psychometric  & Psychometric      & Distractors        & No Knowledge     &  Parametric & Scalability\\
        & Taxonomy             & Task Derivation   & Keyed to Confounds & Reliance         &  Control  \\
\midrule
RAVEN \cite{zhang2019raven} & \textbf{\ding{55}} & \textbf{\ding{51}} & \textbf{\ding{55}} & \textbf{\ding{51}} & \textbf{\ding{55}} & \textbf{\ding{51}} \\
Bongard-LOGO \cite{nie2020bongard} & \textbf{\ding{55}} & \textbf{\ding{55}} & \textbf{\ding{55}} & \textbf{\ding{51}} & \textbf{\ding{55}} & \textbf{\ding{51}} \\
CLEVR \cite{johnson2017clevr} & \textbf{\ding{55}} & \textbf{\ding{55}} & \textbf{\ding{55}} & \textbf{\ding{55}} & \textbf{\ding{51}} & \textbf{\ding{51}} \\
VGRP-Bench \cite{vgrp} & \textbf{\ding{55}} & \textbf{\ding{55}} & \textbf{\ding{55}} & \textbf{\ding{114}} & \textbf{\ding{51}} & \textbf{\ding{51}}\\ 
VisualPuzzles \cite{visualpuzzlesdecouplingmultimodalreasoning} & \textbf{\ding{55}} & \textbf{\ding{55}} & \textbf{\ding{55}} & \textbf{\ding{51}} & \textbf{\ding{51}} & \textbf{\ding{55}} \\
AlgopuzzleVQA \cite{algopuzzlevqa} & \textbf{\ding{55}} & \textbf{\ding{55}} & \textbf{\ding{55}} & \textbf{\ding{114}} & \textbf{\ding{55}} & \textbf{\ding{51}} \\
VisFactor \cite{visfactor} & \textbf{\ding{55}} & \textbf{\ding{51}} & \textbf{\ding{55}} & \textbf{\ding{51}} & \textbf{\ding{55}} & \textbf{\ding{55}} \\
IQBench \cite{iqbench} & \textbf{\ding{55}} & \textbf{\ding{55}} & \textbf{\ding{55}} & \textbf{\ding{51}} & \textbf{\ding{55}} & \textbf{\ding{55}} \\
NTSEBench \cite{ntsebench} & \textbf{\ding{55}} & \textbf{\ding{55}} & \textbf{\ding{55}} & \textbf{\ding{114}} & \textbf{\ding{55}} & \textbf{\ding{55}} \\
SpatialVisBench \cite{spatialvizbench} & \textbf{\ding{55}} & \textbf{\ding{55}} & \textbf{\ding{55}} & \textbf{\ding{51}} & \textbf{\ding{51}} & \textbf{\ding{55}} \\
\midrule
\rowcolor{green!20}
\textbf{Mind's Eye (ours)} & \textbf{\ding{51}} & \textbf{\ding{51}} & \textbf{\ding{51}} & \textbf{\ding{51}} & \textbf{\ding{51}} & \textbf{\ding{51}} \\
\bottomrule
\end{tabular}}
\label{tab:rw-contrast-rw}
\end{table*}

\paragraph{Multimodal and visual reasoning benchmarks.}
General-purpose benchmarks such as MMBench, SEED-Bench, MathVista, and MMMU \citep{liu2024mmbench,li2023seed,li2023seed2,lu2023mathvista,yue2023mmmu,yu2023mmvet} measure breadth across Visual QA, OCR, and mathematical reasoning, but lack parametric control for studying visuospatial understanding. Compositional reasoning benchmarks including CLEVR, RAVEN, and CV-Bench \citep{johnson2017clevr,zhang2019raven,grit2023,hudson2019gqa,suhr2019nlvr2} target attribute binding and relational comparison over single-frame perception, yet offer limited control over geometric transformations (e.g., rotation angle, fold parity) or mental manipulation.

\paragraph{Cognitive and analogical reasoning benchmarks.}
Recent efforts have moved toward cognitive testing: \textit{Mind the Gap} \citep{mindthegap2024} for spatial completion, \textit{Bongard-LOGO} \citep{nie2020bongard} and \textit{VisuLogic} \citep{visulogic2024} for rule induction, and \textit{Do You See Me} \citep{doyouseeme2024} for perception grounding. However, benchmarks like ARC, Bongard-LOGO, and SVRT \citep{arcagi2019,fleuret2011svrt,nie2020bongard} emphasize rule discovery and analogy over stepwise geometric simulation. VisFactor \citep{visfactor} evaluates basic perceptual factors by digitizing FRCT-style tests, while IQBench \citep{iqbench} assesses broader IQ-style reasoning including RPMs and analogies.

\paragraph{Cognitive science foundations.}
Our benchmark draws from classical studies on mental rotation \citep{shepard1971mental}, the Vandenberg \& Kuse MRT \citep{vandenberg1978mrt}, and CogAT paper-folding tests \citep{cogat2009paperfolding}, as well as Hofstadter's theory of analogy \citep{hoffstadter1979godel} and Newell's unified cognition framework \citep{newell1994unified}. While existing multimodal benchmarks emphasize either \emph{recognition} or \emph{abstraction}, they under-specify \emph{internal simulation}. \textsc{Mind's Eye} bridges this gap through programmatic generation of tasks probing whether MLLMs can perform internal transformations—rotation, folding, composition, symmetry recognition—central to human visuospatial intelligence.

\noindent\textbf{Positioning Mind's Eye.}
Table~\ref{tab:rw-contrast-rw} compares benchmarks along six diagnostic axes. \emph{Formal Psychometric Taxonomy} asks whether tasks are organized under a structured cognitive framework with explicit construct-coverage mappings (e.g., a q-matrix), not merely motivated by cognitive science. \emph{Task Derivation from Established Assessments} indicates whether tasks are adapted from validated psychometric instruments. \emph{Distractors Keyed to Confounds} captures whether wrong-answer options are designed to diagnose specific reasoning errors rather than sampled randomly. \emph{No Knowledge Reliance} marks benchmarks solvable without domain knowledge or linguistic priors. \emph{Parametric Control} indicates whether independent generation parameters enable systematic difficulty manipulation, and \emph{Scalability} indicates whether new items can be produced programmatically at negligible cost. To our knowledge, Mind's Eye is the first benchmark in this space to satisfy all six criteria simultaneously.

\section{\textsc{Mind's Eye:} The Benchmark}
\label{benchmark_design}

Going beyond assessing \textit{what} models perceive (such as in object recognition or scene description of images), our proposed benchmark seeks to study the capabilities of models when one focuses on \textit{how} models reason over visual input. Core capacities of human visual intelligence, such as mentally rotating objects, tracking structure through spatial transformations, or inducing abstract rules from visual patterns together constitute \textit{visuocognitive reasoning}, i.e. cognitive operations performed over visual representations, encompassing not only spatial manipulation but also pattern abstraction and relational inference. \textsc{Mind's Eye}, our proposed benchmark, is grounded in an \textbf{Abstraction–Relation–Transformation (ART)} taxonomy that isolates these visuocognitive processes. The taxonomy draws on Carroll's construct of fluid intelligence \citep{carroll1993human} to decompose visual reasoning into three complementary facets: inferring abstract patterns, mapping relational correspondences, and mentally manipulating spatial structure. We now detail the conceptual foundations of ART (\S\ref{subsec:art}), as well as the benchmark's design principles and task suite (\S\ref{subsec:design}).

\subsection{The ART Taxonomy}
\label{subsec:art}
\textsc{Mind's Eye} organizes visuocognitive reasoning along three dimensions: \textbf{Abstraction}, \textbf{Relation}, and \textbf{Transformation} (ART). These are complementary facets of fluid visual reasoning, viz. the capacity to solve novel problems through deliberate, knowledge-independent thought \citep{carroll1993human,schneider2018intelligence}. Each dimension isolates a distinct cognitive operation over visual input:
\begin{itemize}[leftmargin=*]
\item \textbf{Abstraction} requires \textit{inducing latent structure from surface variation}. The solver must identify an underlying rule, pattern, or category that unifies disparate visual instances; for example, recognizing that two differently oriented configurations share identical hierarchical organization. This corresponds to inductive reasoning in psychometric models of fluid intelligence\citep{mcgrew2005cattell}.
\item \textbf{Relation} requires \textit{mapping correspondences across visual structures}. The solver must detect how elements in one configuration align with elements in another, supporting analogical transfer and structural comparison. This corresponds to the relational reasoning central to analogy-making and fluid intelligence \citep{halford2010relational,nie2020bongard}.
\item \textbf{Transformation} requires \textit{mentally simulating spatial operations}. The solver must internally rotate, fold, compose, or otherwise manipulate visual representations to predict outcomes—engaging spatial working memory and figural reasoning \citep{ekstrom1976kit,vandenberg1978mrt}.
\end{itemize}
This framework draws on Carroll's Three Stratum Theory, which situates fluid intelligence (Gf) as a broad factor underlying performance on novel reasoning tasks \citep{carroll1993human}. Crucially, fluid intelligence manifests not only in verbal or symbolic reasoning but also in figural and spatial domains; tasks such as Raven's Progressive Matrices are canonical measures precisely because they require abstract rule induction over visual patterns \citep{raven2000raven}. The ART taxonomy makes explicit the component processes conflating such tasks, enabling targeted diagnosis of where models succeed or fail.

The ART taxonomy provides the theoretical scaffold; the benchmark instantiates 
it through tasks that satisfy three design principles. (1) \textit{Cognitive 
isolation}: tasks require reasoning over visual structure, not retrieval of 
world knowledge, ensuring that performance reflects visuocognitive capacity 
rather than domain familiarity. (2) \textit{Diagnostic precision}: each 
item includes carefully constructed distractors tied to specific reasoning 
errors (e.g., mirrored transformations, parity mistakes), enabling fine-grained 
failure analysis. \textit{Psychometric rigor}: stimulus generation 
follows factorial designs with calibrated difficulty, and all items use 
standardized multiple-choice format to permit reliable comparison across 
models and against human baselines. The following subsection details how 
these principles are realized in the benchmark's eight tasks.

\begin{figure*}[t] 
    \centering
    \includegraphics[width=0.8\linewidth]{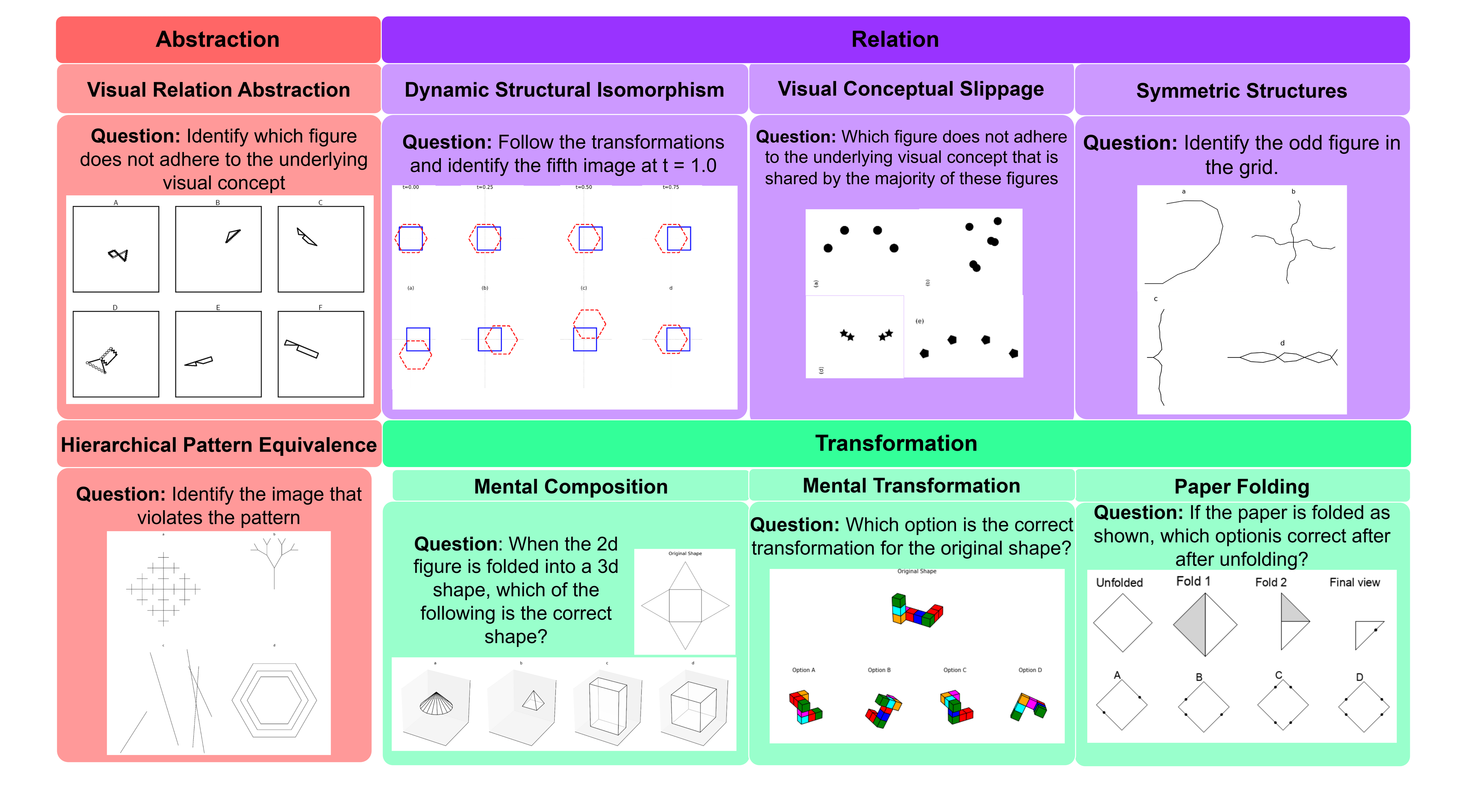}
    \caption{\textbf{Overview of the eight tasks in the proposed \textsc{Mind’s Eye} benchmark}: Each panel shows an example image-question pair of the benchmark}
    \label{fig:overview}
\end{figure*}

\subsection{Benchmark Design}
\label{subsec:design}

\paragraph{Task Suite.}
Our benchmark comprises eight tasks, distributed across the ART dimensions 
(Figure~\ref{fig:overview}). \textit{Abstraction} is probed through 
Visual Relation Abstraction (VRA) and Hierarchical Pattern Equivalence (HPE), 
which require inducing latent rules or detecting recursive structure from 
visual exemplars. \textit{Relation} is probed through Dynamic Structural 
Correspondence (DSC), Visual Conceptual Slippage (VCS), and Symmetric 
Structures (SS), which require mapping correspondences across configurations 
or detecting violations of relational invariants. \textit{Transformation} 
is probed through Mental Transformation (MT), Paper Folding (PF), and 
Mental Composition (MC), classic tests of spatial manipulation adapted from 
the psychometric literature \citep{vandenberg1978mrt,cogat2009paperfolding}. 
Each task is operationalized as a multiple-choice problem with four or six 
options. The formal task–construct mapping is specified via a q-matrix 
(Table~\ref{tab:qmatrix} in the Appendix), following psychometric design 
standards for construct coverage \citep{embretson2013item}.

\paragraph{Stimulus Generation.}
All stimuli are programmatically generated as scalable vector graphics, 
enabling control over geometric parameters and ensuring perceptual 
uniformity across items. Generation follows a factorial design: structural 
parameters that determine task difficulty (e.g., rotation magnitude, fold 
count, hierarchy depth) are varied independently of nuisance parameters 
(e.g., color, spatial layout, surface texture) that should be 
task-irrelevant. This separation serves two purposes. Firstly, it permits 
a priori difficulty calibration based on structural complexity \citep{embretson1983construct,ekstrom1976kit}. Secondly, it mitigates shortcut learning: models cannot exploit incidental correlations between surface features and correct answers. Full generation 
specifications for each task are provided in Appendix~\ref{app:benchmark}.

\noindent \textbf{Rationale for Synthetic Stimulus Generation:} The use of synthetic, programmatically generated SVG images in our benchmark follows established practices in cognitive psychology. Synthetic stimulus generation enables precise control over confounding variables while isolating specific cognitive abilities, an approach employed by foundational assessments that remain the gold standard for measuring human cognition, including the Kit of Factor-Referenced Cognitive Tests \cite{Ekstrom1976ManualFK,Vandenberg1978MentalRotations,Thurstone1938PrimaryMentalAbilities,Guilford1948TheGA,  McFall1993TVPSTestRetest}. Critically, evidence demonstrates that performance on synthetic reasoning tasks correlates with general visual cognition and real-world capabilities across domains \cite{Burton2003VisualImagerySpatialAbility, Moen2020StrengtheningSpatialReasoning, Kunda2012ReasoningRavensAPM}. This approach also aligns with how the community has adopted benchmarks like ARC-AGI \cite{chollet2025arcagi2newchallengefrontier} as measures of progress toward general intelligence. By grounding each visuospatial reasoning task in established cognitive science assessments (Section 3.1), our benchmark provides diagnostic insights into whether failures in visual understanding of models stem from fundamental cognitive limitations versus superficial perception gaps, a distinction crucial for understanding and improving MLLM capabilities. Similar contemporary work has validated this approach for evaluating visual reasoning and abstract reasoning \cite{visulogic2024, mindthegap2025} in vision-language models.

\paragraph{Diagnostic Distractors.}
Each item includes distractor choices in the answer options designed to capture the granularity of a model's reasoning error. For Transformation tasks, distractors include reflections mistaken for rotations, incorrect fold parity, and off-by-$\theta$ rotation errors. For Relation tasks, distractors swap corresponding elements or preserve surface similarity while violating structural correspondence. For Abstraction tasks, distractors match superficial features (e.g., shape, color) but violate the latent rule. This design helps analyze errors beyond a binary (correct/incorrect) observation: the distractor chosen reveals the model's understanding and approach to the solution, enabling a relatively more fine-grained comparison across models. Detailed distractor generations are discussed in Appendix \ref{app:benchmark_design}.

\begin{table*}[t!]
\centering
\rowcolors{2}{blue!10}{white}
\resizebox{\textwidth}{!}{
\begin{tabular}{l|cc|ccc|ccc}
\hiderowcolors
\toprule
& \multicolumn{8}{c}{\textbf{Accuracy} $ \uparrow $ } \\
\midrule
 & \multicolumn{2}{c|}{\textbf{Abstraction}} & \multicolumn{3}{c|}{\textbf{Relation}} 
& \multicolumn{3}{c}{\textbf{Transformation}} \\
 & VRA & HPE & DSC & VCS & SS & MT & PF & MC \\
\midrule
Random Choice & 16.0 & 25.0 & 25.0 & 16.0 & 25.0 & 25.0 & 25.0 & 25.0 \\
\rowcolor{green!10}
Human         & 68.0 & 88.0 & 81.2 & 87.0 & 78.0 & 81.0 & 80.1 & 82.0 \\
\hiderowcolors
\hline
\rowcolor{gray!30}
\multicolumn{9}{l}{\textbf{Open-source multimodal LLMs}} \\
\showrowcolors
Idefics - 8B                & 24.0\scriptsize$\pm$0.02 & 25.1\scriptsize$\pm$0.08 & 32.3\scriptsize$\pm$0.08 & 12.2\scriptsize$\pm$0.01 & 25.0\scriptsize$\pm$0.01 & 32.0\scriptsize$\pm$0.09 & \textbf{41.5\scriptsize$\pm$0.00} & 20.0\scriptsize$\pm$0.02 \\
InternVL3 - 8B              & 22.0\scriptsize$\pm$0.01 & \textbf{29.1\scriptsize$\pm$0.21} & 31.0\scriptsize$\pm$0.03 & 23.7\scriptsize$\pm$0.06 & 29.1\scriptsize$\pm$0.01 & 29.1\scriptsize$\pm$0.05 & 24.6\scriptsize$\pm$0.09 & 28.0\scriptsize$\pm$0.05 \\
LLaMa-3.2 - 11B             & 22.0\scriptsize$\pm$0.06 & 29.0\scriptsize$\pm$0.02 & 31.2\scriptsize$\pm$0.05 & 23.1\scriptsize$\pm$0.02 & 29.3\scriptsize$\pm$0.03 & 29.2\scriptsize$\pm$0.08 & 24.5\scriptsize$\pm$0.02 & 28.0\scriptsize$\pm$0.02 \\
Llava-1.6-Mistral - 7B      & 16.0\scriptsize$\pm$0.01 & 23.7\scriptsize$\pm$0.02 & \textbf{32.4\scriptsize$\pm$0.04} & 30.6\scriptsize$\pm$0.01 & 24.5\scriptsize$\pm$0.02 & \textbf{35.8\scriptsize$\pm$0.04} & 24.1\scriptsize$\pm$0.04 & 29.1\scriptsize$\pm$0.02 \\
Phi3.5-vision-instruct - 8B & 22.0\scriptsize$\pm$0.01 & \textbf{29.1\scriptsize$\pm$0.02} & 31.0\scriptsize$\pm$0.04 & 23.5\scriptsize$\pm$0.01 & 29.7\scriptsize$\pm$0.01 & 29.3\scriptsize$\pm$0.03 & 24.3\scriptsize$\pm$0.01 & 28.7\scriptsize$\pm$0.02 \\
Qwen-2.5-VL - 3B            & 20.0\scriptsize$\pm$0.00 & 26.2\scriptsize$\pm$0.02 & 31.0\scriptsize$\pm$0.09 & 21.0\scriptsize$\pm$0.01 & 21.2\scriptsize$\pm$0.02 & 22.4\scriptsize$\pm$0.01 & 25.0\scriptsize$\pm$0.01 & 27.9\scriptsize$\pm$0.01 \\
Qwen-2.5-VL - 7B            & 19.1\scriptsize$\pm$0.01 & 24.2\scriptsize$\pm$0.01 & 30.4\scriptsize$\pm$0.01 & 22.7\scriptsize$\pm$0.01 & 20.2\scriptsize$\pm$0.04 & 25.7\scriptsize$\pm$0.02 & 25.1\scriptsize$\pm$0.02 & 36.4\scriptsize$\pm$0.01 \\
Qwen-2.5-VL - 32B           & 25.1\scriptsize$\pm$0.01 & 18.3\scriptsize$\pm$0.01 & 22.6\scriptsize$\pm$0.04 & 30.2\scriptsize$\pm$0.07 & 26.3\scriptsize$\pm$0.02 & 27.6\scriptsize$\pm$0.01 & 32.0\scriptsize$\pm$0.02 & 49.5\scriptsize$\pm$0.02 \\
Blip - 2.7B                  & 11.2\scriptsize$\pm$0.07 & 22.7\scriptsize$\pm$0.02 & 18.3\scriptsize$\pm$0.02 & 09.1\scriptsize$\pm$0.04 & 17.0\scriptsize$\pm$0.01 & 10.1\scriptsize$\pm$0.05 & 21.4\scriptsize$\pm$0.02 & 24.0\scriptsize$\pm$0.02 \\
InstructBlip - 4B            & 16.3\scriptsize$\pm$0.01 & 26.4\scriptsize$\pm$0.02 & 19.1\scriptsize$\pm$0.02 & 12.3\scriptsize$\pm$0.04 & 15.0\scriptsize$\pm$0.05 & 28.1\scriptsize$\pm$0.02 & 11.3\scriptsize$\pm$0.01 & 13.0\scriptsize$\pm$0.07 \\
Paligemma - 3B              & 12.5\scriptsize$\pm$0.02 & 17.2\scriptsize$\pm$0.02 & 12.7\scriptsize$\pm$0.03 & 34.7\scriptsize$\pm$0.03 & 13.1\scriptsize$\pm$0.01 & 14.0\scriptsize$\pm$0.02 & 26.4\scriptsize$\pm$0.02 & 29.6\scriptsize$\pm$0.03 \\
Smol - 2.2B                & 11.3\scriptsize$\pm$0.03 & 21.2\scriptsize$\pm$0.03 & 19.2\scriptsize$\pm$0.06 & 21.2\scriptsize$\pm$0.21 & 15.1\scriptsize$\pm$0.01 & 23.5\scriptsize$\pm$0.01 & 26.0\scriptsize$\pm$0.02 & 28.2\scriptsize$\pm$0.31 \\
\hline
\rowcolor{gray!30}
\multicolumn{9}{l}{\textbf{Open-source multimodal LRMs}} \\
\showrowcolors
Vision-G1 - 7B & 22.3\scriptsize$\pm$0.05& 24.2\scriptsize$\pm$0.17& 29.7\scriptsize$\pm$0.11& 24.1\scriptsize$\pm$0.16& 23.6\scriptsize$\pm$0.07 & 25.1\scriptsize$\pm$0.01 & 28.1\scriptsize$\pm$0.17& 38.1\scriptsize$\pm$0.12 \\
GT-Thinker - 7B & 23.1\scriptsize$\pm$0.11& 25.5\scriptsize$\pm$0.09 & 28.1\scriptsize$\pm$0.04 & 25.7\scriptsize$\pm$0.07 & 24.1\scriptsize$\pm$0.14 & 26.7\scriptsize$\pm$0.01 & 27.9\scriptsize$\pm$0.04 & 39.6\scriptsize$\pm$0.17 \\
V-Thinker - 8B & 21.5\scriptsize$\pm$0.15 & 22.6\scriptsize$\pm$0.08 & 27.1\scriptsize$\pm$0.61 & 25.5\scriptsize$\pm$0.71 & 22.9\scriptsize$\pm$0.16 & 24.5\scriptsize$\pm$0.10 & 29.2\scriptsize$\pm$0.18 & 32.2\scriptsize$\pm$0.21 \\

\hline
\rowcolor{gray!30}
\multicolumn{9}{l}{\textbf{API-based models}} \\
\rowcolor{blue!10}
GPT-o3	 & 21.1 & 22.4 & 11.2 & 23.7 & 22.3 & 25.1 & 25.6 & 43.1 \\	
\hiderowcolors
GPT-4o	 & 28.4 & 26.6 & 30.3 & 25.2 & 19.1 & 32.7 & 29.0 & 35.0 \\
\rowcolor{blue!10}
Gemini-2.5& \textbf{29.0} & 20.2 & 30.0 & \textbf{35.3} & \textbf{31.4} & 35.6 & 31.1 & \textbf{51.8} \\

\bottomrule
\end{tabular}
}
\vspace{-4pt}
\caption{\textbf{Task-wise results of MLLMs}: \textbf{\textit{Abstraction:}} VRA (Visual Relation Abstraction), HPE (Hierarchical Pattern Equivalence). \textbf{\textit{Relation:}} DSC (Dynamic Structural Correspondence), VCS (Visual Conceptual Slippage), SS (Symmetric Structures). \textbf{\textit{Transformation:}} MT (Mental Transformation), PF (Paper Folding), MC (Mental Composition).}
\vspace{-13pt}
\label{tab:results}
\end{table*}

\paragraph{Benchmark Scale.}
Our evaluation set comprises 800 items: 100 per task, balanced across the three ART dimensions. The size of our benchmark follows the scale of existing synthetic benchmarks for targeted capability assessment \cite{gadz2024visulogic,mindthegap2025}. We however note that while our objective in this work is diagnostic assessment of MLLMs, since our stimuli are programmatically generated, our dataset can scale without additional annotation cost, if required for training purposes. We provide an extended set of 2{,}500 items per task (20{,}000 total), generated with 
identical templates and difficulty controls, for fine-tuning or representation learning. Both the diagnostic and extended versions of our benchmark will be made public on acceptance. 
The diagnostic and extended partitions do not have data overlap, and hence are maintained separately to support both fair comparison. Full generation procedures, per-task specifications, and dataset statistics are detailed in Appendix~\ref{app:benchmark}.

\paragraph{Human Evaluation.}
To study how humans perform on the benchmark, we recruited 30 participants of the age group ranging from 20 to 40, with a gender distribution of 19 males and 11 females. Each participant was presented with 5 questions from each task, sampled via inverse-frequency weighting from a pool of 20 questions per task (total of 40 questions across tasks). To minimize bias and ensure consistency, all participants first completed an identical calibration phase consisting of 8 examples spanning all tasks. 
Human accuracy for each subtask was measured by comparing participant responses against the ground truth. For more details on the human evaluation protocol, see Appendix \ref{humanproto}.

\vspace{-3pt}

\section{Experiments and Results}
\label{experimets}

\paragraph{Experimental Setup.}
We evaluate \textsc{Mind's Eye} on a wide range of recent MLLMs, including GPT-4o, GPT-o3 and Gemini-2.5 pro which are accessed via their respective proprietary APIs, as well as open-source models: 
LLaVA-1.6-7B, Llama-3.2-11B-Vision, phi-4-multimodal-instruct-5.7B, Qwen2.5-VL-Instruct (3B, 7B and 32B) and InternVL3.5-8B. To ensure fair comparison, all models are evaluated on identical visual inputs and standardized textual prompts. Since modern MLLMs often produce long, free-form outputs, rule-based answer extraction can be unreliable \citep{vlmevalkit2023,li2024blink}. Following recent practice \citep{lu2024mathvistaevaluatingmathematicalreasoning,zhang2024mathversedoesmultimodalllm}, we adopt an \textit{expert LLM evaluation pipeline} comprising three stages:  
(1) candidate model receives the image and question in a fixed template;  
(2) an answer extractor, Gemma-3 \citep{gemmateam2025gemma3technicalreport}, parses the raw output into a concise response; and  (3) the parsed response is mapped to standardized task-specific labels for accuracy computation across all eight tasks. This approach leverages robust semantic extraction via a large model, while maintaining fully automated, reproducible scoring (details in Appendix~\ref{ap:eval_setup}). To prevent positional bias, correct answer options were randomly rotated across positions following standard MCQ evaluation practice.

\paragraph{Prompting strategies.}  
Since multimodal reasoning can be sensitive to prompt phrasing \citep{wei2022cot,kojima2022zerocot}, we evaluate four structured prompting paradigms: Chain-of-Thought (CoT), Meta-Task Framing, Step-by-Step Instruction (SBS), and Hint-based prompting. Full prompt templates and examples for each strategy are provided in Appendix~\ref{ap:eval_setup} and ~\ref{app:prompt_strategies}. Results for CoT-based prompts are in Table \ref{tab:results}; results of other prompting strategies are in Appendix~\ref{app:prompt_style_performance}.

\paragraph{Main Results.} Our primary results are reported in Table \ref{tab:results}. The results reveal a general weakness of MLLMs on the considered visuospatial reasoning tasks, especially when compared to human performance. Although these models can often identify 3D arrangements or object correspondences, they struggle to integrate this perception into consistent reasoning, frequently selecting implausible answers (Fig~\ref{fig:mrtgpt4o} in Appendix). Models particularly struggle with tasks requiring interpreting temporal sequences and tracking visual elements across transformations. For example, \textit{Dynamic Structural Correspondence} tests unidirectional tracking of changes across a sequence, while \textit{Paper Folding} requires not only forward tracking but also mentally reversing the process; in both cases, models frequently misinterpret the visual dynamics. Scaling analysis (Fig~\ref{fig:model_task_heatmap}) shows that performance generally increases with model size, reinforcing the role of model parameter scale; however, performance also increases when moving from abstraction-heavy tasks to transformation-oriented ones, underscoring persistent weaknesses in reasoning abilities based on mental manipulation. A likely root cause of these failures may be the models' limited capability to \textit{cognitively group figures and concepts} into coherent representations. We also observe a \textit{strong dependence on surface perception}: in \textit{Mental Composition}, models succeed when the unfolded net visually resembles a cube but fail when correct inference requires mentally folding a shape into a nontrivial 3D structure (Fig~\ref{fig:mcgpt4o} in Appendix).
We note that while the specific tasks are different, related previous efforts \cite{huang2025humancognitivebenchmarksreveal,stogiannidis2025mindgapbenchmarkingspatial,urgun2025analysisarchitecturalimpactllmbased} also report similar performance numbers. Our studies with prompt variations corroborate our above observations (see Appendix \ref{app:encoding}).



\vspace{-4pt}
\section{Analysis and Discussion}
\label{error_analysis}

\noindent \textbf{Attention Alignment and Accuracy.}
We analyze whether option-directed attention predicts reasoning success. For each item, we compute an \emph{Option-Specific Attention Score} ($\mathrm{OAS}_{\text{correct}}$): the mean normalized attention mass directed toward the correct option's spatial region during reasoning-token generation. Across 200 items (25 per task, stratified by difficulty), $\mathrm{OAS}_{\text{correct}}$ correlates positively with accuracy (point-biserial $r_{\text{pb}} = 0.34$, $p < 0.001$). Yet even in the highest-attention quartile, accuracy remains well below human performance ($>$80\%), indicating that attention alignment is necessary but not sufficient for correct reasoning. This dissociation is further supported by a paired analysis: on correct predictions, attention to the correct option exceeds attention to distractors ($0.24$ vs.\ $0.16$; $t(87) = 4.32$, $p < 0.001$), whereas incorrect predictions show no such preference ($0.18$ vs.\ $0.17$; $t(112) = 0.84$, $p = 0.40$). Thus, while models are able to localize the relevant information but fail to reason over it reliably (see Appendix \ref{app:attention_performance}).

\noindent \textbf{Robustness Under Relative Attention Normalization.}
A natural concern with the preceding analysis is that raw softmax attention may be confounded by register-token artifacts~\citep{darcet2024vision}, which can inflate diffuse background attention and obscure genuine spatial focus. To address this, we follow \citet{zhang2025mllms} and recompute all attention metrics using \emph{relative attention}.

Reassuringly, the findings reported above not only hold but become somewhat sharper under this normalization. The point-biserial correlation between $\text{OAS}_{\text{correct}}$ and accuracy increases from $r_{pb} = 0.34$ to $r_{pb} = 0.41$ ($p < 0.001$), suggesting that relative attention provides a cleaner predictor once register noise is factored out. On correct predictions, the effect size for attention preference toward the correct option over distractors grows from Cohen's $d = 1.15$ to $d = 1.41$ ($t(86) = 5.89$, $p < 0.001$). On incorrect predictions, relative attention to the selected wrong option remains statistically indistinguishable from attention to the correct option ($p = 0.31$), confirming that the dissociation is not an artifact of noisy attention extraction. Mean Region-Aligned Attention (RAA) increases modestly from $0.18$ to $0.21$, yet even this improved grounding remains far below what would be needed for reliable reasoning and well below human performance ($>$80\%).

This reinforces our earlier conclusion: the bottleneck appears to lie not in \emph{where} models attend, but in their limited ability to perform the cognitive operations required to reason over correctly localized information.

\begin{tcolorbox}[colback=gray!10,colframe=gray!70]
Attention to relevant regions correlates with accuracy, yet localization alone proves insufficient: \textit{models can identify where to look but not how to reason over what they see}.
\end{tcolorbox}
\noindent \textbf{Reasoning Stability under Prompt Variations.}

\begin{figure}[ht] 
\centering
    \centering
    \includegraphics[width=\columnwidth]{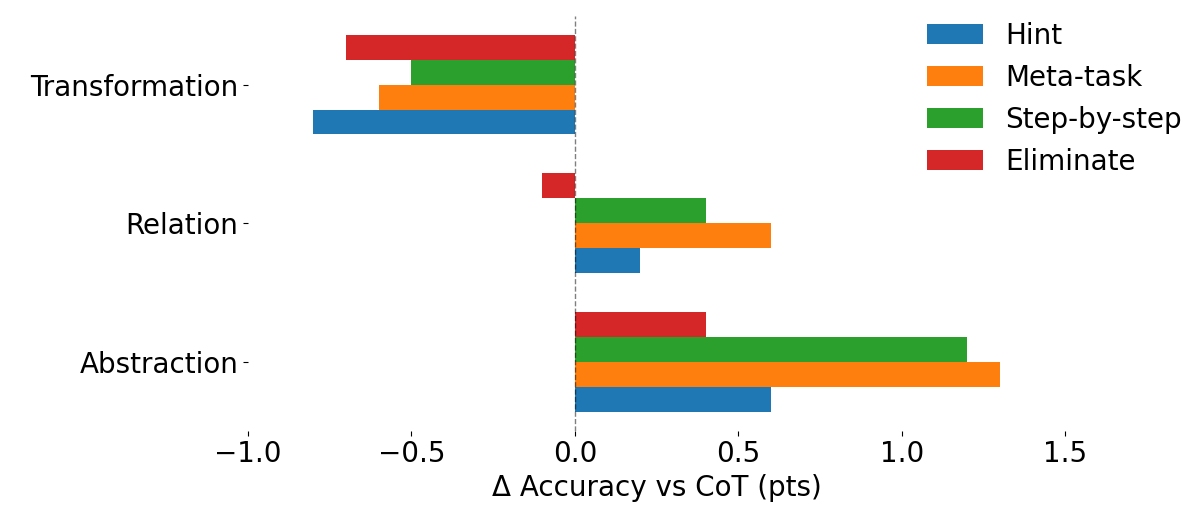}
    \vspace{-22pt}
    \caption{Change in accuracy ($\Delta$ Accuracy) of different prompt variations w.r.t. CoT performance across ART dimensions}
\label{fig:prompt_deltas}
\end{figure}
Figure \ref{fig:prompt_deltas} reports the differential effects of prompt variations on model performance when compared with baseline CoT performance. The results reveal that prompting effects are dimension-dependent rather than uniformly beneficial. Transformation tasks exhibit consistent performance degradation across all alternative prompting strategies, with Hint prompting showing the largest drop (approximately $-0.9$ pts), suggesting that tasks requiring internal simulation are particularly sensitive to instruction framing. In contrast, Abstraction tasks benefit from structured guidance, with Meta-task and Step-by-step prompting yielding gains of approximately $+1.3$ pts, indicating that explicit scaffolding facilitates latent rule derivation. Relation tasks show intermediate behavior, with modest improvements under Meta-task prompting but near-baseline performance otherwise. These asymmetric effects suggest that while prompting can enhance pattern recognition and abstraction, it fails to induce the procedural operations underlying robust transformation reasoning (see Appendix ~\ref{app:cotvncot}).

\begin{tcolorbox}[colback=gray!10,colframe=gray!70]
Prompting benefits abstraction but impairs transformation performance: \textit{scaffolding facilitates rule derivation yet fails to elicit procedural visuocognitive operations}.
\end{tcolorbox}

\begin{figure}[ht]
    \centering
    \includegraphics[width=\columnwidth]{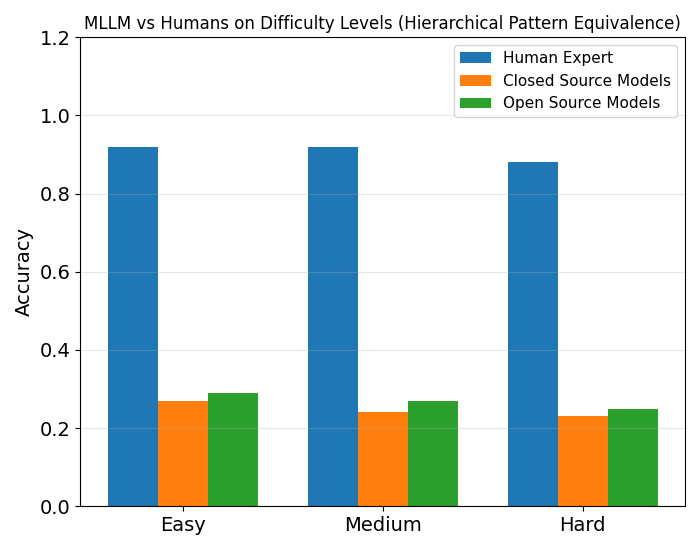}
    \hfill
    \includegraphics[width=\columnwidth]{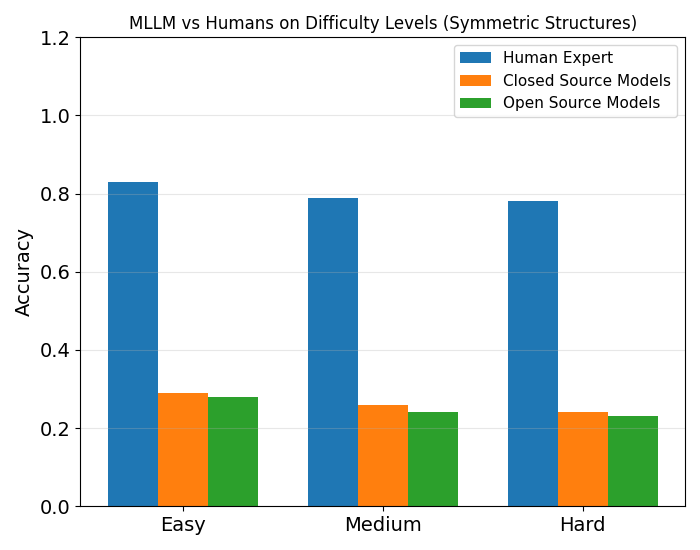}
    \caption{Human-model performance gap across ART taxonomy dimensions stratified by difficulty. Each bar represents the macro-average accuracy for a task across all models in that category (see Table~\ref{tab:results}).}
    \label{fig:difficulty}
\end{figure}

\noindent \textbf{Performance Across ART dimension by Difficulty.}
To examine how model capabilities vary across the three core dimensions of our ART taxonomy, we analyzed performance stratified by difficulty level (Easy, Medium, Hard). Figure \ref{fig:difficulty} reveals a consistent pattern: human accuracy degrades predictably with difficulty (from $>0.80$ on Easy to $~0.25$ on Hard), while both open and closed-source models exhibit flat performance curves ($0.20–0.45$) regardless of difficulty. On certain hard instances, particularly Mental Transformation and Visual Relation Abstraction, human accuracy drops to model-level performance. However, this convergence is asymmetric: humans fail because the task is hard; models fail because they cannot perform the underlying operation at any difficulty level. This difficulty-invariant failure across all three dimensions suggests that current MLLMs lack foundational visual-cognitive operations beyond merely struggling with complex instances (refer Appendix \ref{app:difficulty}).

\begin{tcolorbox}[colback=gray!10,colframe=gray!70]
MLLMs fail uniformly across difficulty levels: \textit{they lack foundational visuo-cognitive operations rather than merely struggling with complexity}.
\end{tcolorbox}
\vspace{-3pt}
\section{Conclusions}
\label{discussion}
\vspace{-3pt}

We present \textsc{Mind's Eye}, a visuocognitive benchmark for evaluating MLLMs on visual intelligence tasks, organized along three axes inspired by Carroll's three-stratum theory: Abstraction, Relation, and Transformation. Our evaluation reveals a persistent human-model gap: non-expert humans achieve 80\% mean accuracy while top MLLMs remain below 50\%. Prompting strategies yield task-dependent but modest improvements without altering error profiles. Key findings include: (i) MLLMs rely heavily on perceptual cues with limited coupling between textual reasoning and visual evidence; (ii) scaling improves surface-matching tasks more than those requiring internal simulation; and (iii) our ART-aligned, parametric design exposes specific failure modes, suggesting directions for advances in grounded attention, spatial working memory, and transformation-aware representations. Future work includes open-ended responses, 3D perception tasks, and human studies across expertise levels.

\section*{Limitations}
As stated earlier, our Mind's Eye benchmark focuses on using a \emph{multiple\mbox{-}choice} scoring for reliability and objectivity of comparison; however, open\mbox{-}ended generation may bring about its own unique set of insights. Secondly, our tasks herein center on 2D renderings with controlled 3D implications; fully 3D inputs and interactions remain a focus of future work. Thirdly, our human baseline uses non\mbox{-}expert adults in a single language setting; cross\mbox{-}lingual and expert cohorts may shift absolute levels of performance (we hypothesize though that relative gaps are likely to remain, based on our observations in this work).

\paragraph{Threats to Validity.}
\textit{Construct validity.} While tasks target Abstraction/Relation/Transformation, they are proxies for broader visuo\mbox{-}cognition; we limit language priors but cannot eliminate all heuristics. 
\textit{External validity.} Findings on synthetic, controlled items may not transfer to natural images; we release generators to enable domain shifts. 
\textit{Reliability.} Item difficulty and distractor quality are controlled parametrically; bootstrap CI's and mixed\mbox{-}effects models quantify uncertainty.

\paragraph{Risks of Anthropomorphism.}
Cognitive-style performance can invite anthropomorphic interpretations, for e.g., ascribing 'mental rotation,' 'working memory,' or 'attention' in the human sense to models. This risks conflating \emph{functional} success on a narrowly specified task with \emph{mechanistic} equivalence to human cognition. Over-interpretation can also invert causality: improvements from prompt engineering or data exposure may be mistaken for emergent cognitive faculties. To mitigate this, we treat model outputs as \emph{behavioral signatures} under controlled stimuli, avoid mentalistic language, and separate construct-level claims (what is measured) from implementation claims (how models compute).

\paragraph{Broader Impact/Ethics.}
Our benchmark uses synthetic, knowledge-minimal stimuli designed to reduce privacy, content, and demographic risks; nevertheless, we report a few broader limitations that hold for almost all benchmarks. Firstly, publishing leaderboards may encourage narrow optimization, masking real-world limitations in safety-critical contexts (education, assessment, medical imaging). Secondly, human baselines reflect a specific population (age, language, interface); results should not be used to rank individuals or groups. Thirdly, cognitive-style tests could be misapplied as gatekeeping tools in hiring or education; our license and documentation explicitly prohibit human evaluation or selection use. We release generators, seeds, and scoring code to enable transparent replication and stress-testing, and we encourage researchers to report uncertainty, disclose inference settings, and evaluate interventions (e.g., grounded-attention or working-memory modules) for safety as well as performance.

\newpage
\bibliography{refs}

@inproceedings{
darcet2024vision,
title={Vision Transformers Need Registers},
author={Timoth{\'e}e Darcet and Maxime Oquab and Julien Mairal and Piotr Bojanowski},
booktitle={The Twelfth International Conference on Learning Representations},
year={2024},
url={https://openreview.net/forum?id=2dnO3LLiJ1}
}

@inproceedings{
zhang2025mllms,
title={{MLLM}s Know Where to Look: Training-free Perception of Small Visual Details with Multimodal {LLM}s},
author={Jiarui Zhang and Mahyar Khayatkhoei and Prateek Chhikara and Filip Ilievski},
booktitle={The Thirteenth International Conference on Learning Representations},
year={2025},
url={https://openreview.net/forum?id=DgaY5mDdmT}
}

@misc{urgun2025analysisarchitecturalimpactllmbased,
      title={An Analysis of Architectural Impact on LLM-based Abstract Visual Reasoning: A Systematic Benchmark on RAVEN-FAIR}, 
      author={Sinan Urgun and Seçkin Arı},
      year={2025},
      eprint={2511.11916},
      archivePrefix={arXiv},
      primaryClass={cs.AI},
      url={https://arxiv.org/abs/2511.11916}, 
}

@misc{stogiannidis2025mindgapbenchmarkingspatial,
      title={Mind the Gap: Benchmarking Spatial Reasoning in Vision-Language Models}, 
      author={Ilias Stogiannidis and Steven McDonagh and Sotirios A. Tsaftaris},
      year={2025},
      eprint={2503.19707},
      archivePrefix={arXiv},
      primaryClass={cs.CV},
      url={https://arxiv.org/abs/2503.19707}, 
}

@misc{huang2025humancognitivebenchmarksreveal,
      title={Human Cognitive Benchmarks Reveal Foundational Visual Gaps in MLLMs}, 
      author={Jen-Tse Huang and Dasen Dai and Jen-Yuan Huang and Youliang Yuan and Xiaoyuan Liu and Wenxuan Wang and Wenxiang Jiao and Pinjia He and Zhaopeng Tu and Haodong Duan},
      year={2025},
      eprint={2502.16435},
      archivePrefix={arXiv},
      primaryClass={cs.CV},
      url={https://arxiv.org/abs/2502.16435}, 
}

@article{landis1977measurement,
  title={The measurement of observer agreement for categorical data},
  author={Landis, J Richard and Koch, Gary G},
  journal={Biometrics},
  volume={33},
  number={1},
  pages={159--174},
  year={1977},
  month={mar},
  publisher={JSTOR},
  pmid={843571}
}

@misc{chollet2025arcagi2newchallengefrontier,
      title={ARC-AGI-2: A New Challenge for Frontier AI Reasoning Systems}, 
      author={Francois Chollet and Mike Knoop and Gregory Kamradt and Bryan Landers and Henry Pinkard},
      year={2025},
      eprint={2505.11831},
      archivePrefix={arXiv},
      primaryClass={cs.AI},
      url={https://arxiv.org/abs/2505.11831}, 
}

@inproceedings{Kunda2012ReasoningRavensAPM,
  author    = {Kunda, Maithilee and McGreggor, Keith and Goel, Ashok},
  title     = {Reasoning on the Raven's Advanced Progressive Matrices Test with Iconic Visual Representations},
  booktitle = {Proceedings of the Cognitive Science Society},
  year      = {2012},
  address   = {Sapporo, Japan}
}

@article{Moen2020StrengtheningSpatialReasoning,
  author  = {Moen, Kevin C. and Beck, Michael R. and Saltzmann, Samuel M. and others},
  title   = {Strengthening spatial reasoning: elucidating the attentional and neural mechanisms associated with mental rotation skill development},
  journal = {Cognitive Research: Principles and Implications},
  year    = {2020},
  volume  = {5},
  number  = {1},
  pages   = {20},
  doi     = {10.1186/s41235-020-00211-y}
}

@article{Burton2003VisualImagerySpatialAbility,
  author  = {Burton, Lorelle},
  title   = {Examining the Relation Between Visual Imagery and Spatial Ability Tests},
  journal = {International Journal of Testing},
  year    = {2003},
  volume  = {3},
  number  = {3},
  pages   = {277--291},
  doi     = {10.1207/S15327574IJT0303_6}
}

@article{McFall1993TVPSTestRetest,
  author  = {McFall, Suzanne A. and Deitz, Jeanette C. and Crowe, Terry K.},
  title   = {Test-retest reliability of the Test of Visual Perceptual Skills with children with learning disabilities},
  journal = {American Journal of Occupational Therapy},
  year    = {1993},
  volume  = {47},
  number  = {9},
  pages   = {819--824},
  doi     = {10.5014/ajot.47.9.819}
}

@article{Guilford1948TheGA,
  title={The Guilford-Zimmerman Aptitude Survey.},
  author={J. P. Guilford and Wayne S. Zimmerman},
  journal={Journal of Applied Psychology},
  year={1948},
  volume={32},
  pages={24-34},
  url={https://api.semanticscholar.org/CorpusID:145008409}
}

@book{Thurstone1938PrimaryMentalAbilities,
  author    = {Thurstone, Louis Leon},
  title     = {Primary Mental Abilities},
  publisher = {University of Chicago Press},
  address   = {Chicago},
  year      = {1938}
}

@article{Vandenberg1978MentalRotations,
  author  = {Vandenberg, Steven G. and Kuse, Allan R.},
  title   = {Mental rotations, a group test of three-dimensional spatial visualization},
  journal = {Perceptual and Motor Skills},
  year    = {1978},
  volume  = {47},
  number  = {2},
  pages   = {599--604},
  doi     = {10.2466/pms.1978.47.2.599},
  pmid    = {724398}
}

@inproceedings{Ekstrom1976ManualFK,
  title={Manual for kit of factor-referenced cognitive tests},
  author={Ruth B. Ekstrom and John W. French and Harry H. Harman},
  year={1976},
  url={https://api.semanticscholar.org/CorpusID:141329865}
}

@misc{vgrp,
      title={VGRP-Bench: Visual Grid Reasoning Puzzle Benchmark for Large Vision-Language Models}, 
      author={Yufan Ren and Konstantinos Tertikas and Shalini Maiti and Junlin Han and Tong Zhang and Sabine Süsstrunk and Filippos Kokkinos},
      year={2025},
      eprint={2503.23064},
      archivePrefix={arXiv},
      primaryClass={cs.CV},
      url={https://arxiv.org/abs/2503.23064}, 
}

@inproceedings{ntsebench,
    title = "{NTSEBENCH}: Cognitive Reasoning Benchmark for Vision Language Models",
    author = "Pandya, Pranshu  and
      Gupta, Vatsal  and
      Talwarr, Agney S  and
      Kataria, Tushar  and
      Roth, Dan  and
      Gupta, Vivek",
    editor = "Chiruzzo, Luis  and
      Ritter, Alan  and
      Wang, Lu",
    booktitle = "Findings of the Association for Computational Linguistics: NAACL 2025",
    month = apr,
    year = "2025",
    address = "Albuquerque, New Mexico",
    publisher = "Association for Computational Linguistics",
    url = "https://aclanthology.org/2025.findings-naacl.204/",
    doi = "10.18653/v1/2025.findings-naacl.204",
    pages = "3680--3708"
}

@misc{spatialvizbench,
      title={SpatialViz-Bench: An MLLM Benchmark for Spatial Visualization}, 
      author={Siting Wang and Minnan Pei and Luoyang Sun and Cheng Deng and Kun Shao and Zheng Tian and Haifeng Zhang and Jun Wang},
      year={2025},
      eprint={2507.07610},
      archivePrefix={arXiv},
      primaryClass={cs.CV},
      url={https://arxiv.org/abs/2507.07610}, 
}

@inproceedings{algopuzzlevqa,
    title = "{A}lgo{P}uzzle{VQA}: Diagnosing Multimodal Reasoning Challenges of Language Models with Algorithmic Multimodal Puzzles",
    author = "Ghosal, Deepanway  and
      Toh, Vernon  and
      Chia, Yew Ken  and
      Poria, Soujanya",
    editor = "Chiruzzo, Luis  and
      Ritter, Alan  and
      Wang, Lu",
    booktitle = "Proceedings of the 2025 Conference of the Nations of the Americas Chapter of the Association for Computational Linguistics: Human Language Technologies (Volume 1: Long Papers)",
    month = apr,
    year = "2025",
    address = "Albuquerque, New Mexico",
    publisher = "Association for Computational Linguistics",
    url = "https://aclanthology.org/2025.naacl-long.486/",
    doi = "10.18653/v1/2025.naacl-long.486",
    pages = "9615--9632",
    ISBN = "979-8-89176-189-6",
}

@misc{visualpuzzlesdecouplingmultimodalreasoning,
      title={VisualPuzzles: Decoupling Multimodal Reasoning Evaluation from Domain Knowledge}, 
      author={Yueqi Song and Tianyue Ou and Yibo Kong and Zecheng Li and Graham Neubig and Xiang Yue},
      year={2025},
      eprint={2504.10342},
      archivePrefix={arXiv},
      primaryClass={cs.CL},
      url={https://arxiv.org/abs/2504.10342}, 
}

@misc{visfactor,
      title={Human Cognitive Benchmarks Reveal Foundational Visual Gaps in MLLMs}, 
      author={Jen-Tse Huang and Dasen Dai and Jen-Yuan Huang and Youliang Yuan and Xiaoyuan Liu and Wenxuan Wang and Wenxiang Jiao and Pinjia He and Zhaopeng Tu and Haodong Duan},
      year={2025},
      eprint={2502.16435},
      archivePrefix={arXiv},
      primaryClass={cs.CV},
      url={https://arxiv.org/abs/2502.16435}, 
}

@misc{iqbench,
      title={IQBench: How "Smart'' Are Vision-Language Models? A Study with Human IQ Tests}, 
      author={Tan-Hanh Pham and Phu-Vinh Nguyen and Dang The Hung and Bui Trong Duong and Vu Nguyen Thanh and Chris Ngo and Tri Quang Truong and Truong-Son Hy},
      year={2025},
      eprint={2505.12000},
      archivePrefix={arXiv},
      primaryClass={cs.CV},
      url={https://arxiv.org/abs/2505.12000}, 
}

@inproceedings{suhr2019nlvr2,
  title     = {A Corpus for Reasoning about Natural Language Grounded in Photographs},
  author    = {Suhr, Alane and Trischler, Adam and Cheung, Jackie Chi Kit and Artzi, Yoav},
  booktitle = {ACL},
  year      = {2019}
}

@article{fleuret2011svrt,
  title   = {Comparing Machines and Humans on a Visual Categorization Test},
  author  = {Fleuret, Fran\c{c}ois and Li, Tingting and Dubout, Charles and Wampler, Eric K. and Yantis, Steven and Geman, Donald},
  journal = {PNAS},
  volume  = {108},
  number  = {43},
  pages   = {17621--17625},
  year    = {2011}
}

@article{agarwal2024promptwizard,
  title        = {PromptWizard: Task-Aware Prompt Optimization Framework},
  author       = {Eshaan Agarwal and Joykirat Singh and Vivek Dani and Raghav Magazine and Tanuja Ganu and Akshay Nambi},
  journal      = {arXiv preprint arXiv:2405.18369},
  year         = {2024},
  url          = {https://arxiv.org/abs/2405.18369}
}

@misc{bogdan2025thoughtanchorsllmreasoning,
      title={Thought Anchors: Which LLM Reasoning Steps Matter?}, 
      author={Paul C. Bogdan and Uzay Macar and Neel Nanda and Arthur Conmy},
      year={2025},
      eprint={2506.19143},
      archivePrefix={arXiv},
      primaryClass={cs.LG},
      url={https://arxiv.org/abs/2506.19143}, 
}

@misc{narrowgate2024,
  title={The Narrow Gate: Localized Image-Text Communication in Native Multimodal Models}, 
      author={Alessandro Serra and Francesco Ortu and Emanuele Panizon and Lucrezia Valeriani and Lorenzo Basile and Alessio Ansuini and Diego Doimo and Alberto Cazzaniga},
      year={2025},
      eprint={2412.06646},
      archivePrefix={arXiv},
      primaryClass={cs.CV},
      url={https://arxiv.org/abs/2412.06646}, 
}

@inproceedings{autodiscovery2024,
 title={Automatic Discovery of Visual Circuits}, 
      author={Achyuta Rajaram and Neil Chowdhury and Antonio Torralba and Jacob Andreas and Sarah Schwettmann},
      year={2024},
      eprint={2404.14349},
      archivePrefix={arXiv},
      primaryClass={cs.CV},
      url={https://arxiv.org/abs/2404.14349}, 
}

@inproceedings{sharedcircuits2024,
 title = "Towards Interpretable Sequence Continuation: Analyzing Shared Circuits in Large Language Models",
    author = "Lan, Michael  and
      Torr, Philip  and
      Barez, Fazl",
    editor = "Al-Onaizan, Yaser  and
      Bansal, Mohit  and
      Chen, Yun-Nung",
    booktitle = "Proceedings of the 2024 Conference on Empirical Methods in Natural Language Processing",
    month = nov,
    year = "2024",
    address = "Miami, Florida, USA",
    publisher = "Association for Computational Linguistics",
    url = "https://aclanthology.org/2024.emnlp-main.699/",
    doi = "10.18653/v1/2024.emnlp-main.699",
    pages = "12576--12601",
}

@article{embretson1983construct,
  title={Construct validity: Construct representation versus nomothetic span},
  author={Embretson, Susan E.},
  journal={Psychological Bulletin},
  volume={93},
  number={1},
  pages={179--197},
  year={1983},
  doi={10.1037/0033-2909.93.1.179}
}

@book{embretson2013item,
  title={Item Response Theory for Psychologists},
  author={Embretson, Susan E. and Reise, Steven P.},
  year={2013},
  publisher={Psychology Press},
  doi={10.4324/9781315807048}
}

@book{ekstrom1976kit,
  title={Kit of Factor-Referenced Cognitive Tests},
  author={Ekstrom, Ruth B. and French, John W. and Harman, Harry H. and Dermen, Diran},
  year={1976},
  publisher={Educational Testing Service},
  address={Princeton, NJ}
}

@article{vandenberg1978mental,
  title={Mental rotations, a group test of three-dimensional spatial visualization},
  author={Vandenberg, Steven G. and Kuse, Allan R.},
  journal={Perceptual and Motor Skills},
  volume={47},
  number={2},
  pages={599--604},
  year={1978},
  doi={10.2466/pms.1978.47.2.599}
}

@book{de2003theory,
  title={Explanatory Item Response Models: A Generalized Linear and Nonlinear Approach},
  author={De Boeck, Paul and Wilson, Mark},
  year={2003},
  publisher={Springer},
  doi={10.1007/978-1-4757-3799-0}
}

@article{yu2023seed,
  title={Seed-bench: Benchmarking multimodal llms with generative comprehension},
  author={Li, Bohao and Wang, Rui and Wang, Guangzhi and Ge, Yuying and Ge, Yixiao and Shan, Ying},
  journal={arXiv preprint arXiv:2307.16125},
  year={2023}
}

@inproceedings{li2023mmbench,
author = {Liu, Yuan and Duan, Haodong and Zhang, Yuanhan and Li, Bo and Zhang, Songyang and Zhao, Wangbo and Yuan, Yike and Wang, Jiaqi and He, Conghui and Liu, Ziwei and Chen, Kai and Lin, Dahua},
title = {MMBench: Is Your Multi-modal Model an All-Around Player?},
year = {2024},
isbn = {978-3-031-72657-6},
publisher = {Springer-Verlag},
address = {Berlin, Heidelberg},
url = {https://doi.org/10.1007/978-3-031-72658-3_13},
doi = {10.1007/978-3-031-72658-3_13},
booktitle = {Computer Vision – ECCV 2024: 18th European Conference, Milan, Italy, September 29–October 4, 2024, Proceedings, Part VI},
pages = {216–233},
numpages = {18},
location = {Milan, Italy}
}

@inproceedings{nie2020adversarial,
  title={Adversarial NLI: A New Benchmark for Natural Language Understanding},
  author={Nie, Yixin and Williams, Adina and Dinan, Emily and Bansal, Mohit and Weston, Jason and Kiela, Douwe},
  booktitle={Proceedings of the 58th Annual Meeting of the Association for Computational Linguistics},
  year={2020},
  pages={4885--4901},
  doi={10.18653/v1/2020.acl-main.441}
}

@inproceedings{zellers2019hellaswag,
  title={HellaSwag: Can a Machine Really Finish Your Sentence?},
  author={Zellers, Rowan and Holtzman, Ari and Bisk, Yonatan and Farhadi, Ali and Choi, Yejin},
  booktitle={Proceedings of the 57th Annual Meeting of the Association for Computational Linguistics},
  year={2019},
  pages={4791--4800},
  doi={10.18653/v1/P19-1472}
}

@book{carroll1993human,
  title={Human cognitive abilities: A survey of factor-analytic studies},
  author={Carroll, John B.},
  year={1993},
  publisher={Cambridge University Press}
}

@book{mcgrew2005cattell,
  title={The Cattell-Horn-Carroll theory of cognitive abilities},
  author={McGrew, Kevin S},
  year={2005},
  publisher={Springer}
}

@article{schneider2018intelligence,
  title={Intelligence in education: Cattell-Horn-Carroll theory and assessment},
  author={Schneider, W. Joel and McGrew, Kevin S.},
  journal={Psychology in the Schools},
  volume={55},
  number={1},
  pages={7--43},
  year={2018},
  publisher={Wiley Online Library}
}

@article{raven2000raven,
  title={Raven’s Progressive Matrices},
  author={Raven, John},
  journal={Handbook of Nonverbal Assessment},
  pages={223--237},
  year={2000},
  publisher={Springer}
}

@article{halford2010relational,
  title={Relational complexity and reasoning},
  author={Halford, Graeme S. and Wilson, William H. and Phillips, Steven},
  journal={Cognitive Science},
  volume={34},
  number={8},
  pages={1451--1476},
  year={2010},
  publisher={Wiley}
}

@book{newell1994unified,
  title={Unified theories of cognition},
  author={Newell, Allen},
  year={1994},
  publisher={Harvard University Press}
}

@inproceedings{vlmevalkit2023,
author = {Duan, Haodong and Yang, Junming and Qiao, Yuxuan and Fang, Xinyu and Chen, Lin and Liu, Yuan and Dong, Xiaoyi and Zang, Yuhang and Zhang, Pan and Wang, Jiaqi and Lin, Dahua and Chen, Kai},
title = {VLMEvalKit: An Open-Source ToolKit for Evaluating Large Multi-Modality Models},
year = {2024},
isbn = {9798400706868},
publisher = {Association for Computing Machinery},
address = {New York, NY, USA},
url = {https://doi.org/10.1145/3664647.3685520},
doi = {10.1145/3664647.3685520},
booktitle = {Proceedings of the 32nd ACM International Conference on Multimedia},
pages = {11198–11201},
numpages = {4},
location = {Melbourne VIC, Australia},
series = {MM '24}
}

@inproceedings{li2024blink,
author = {Fu, Xingyu and Hu, Yushi and Li, Bangzheng and Feng, Yu and Wang, Haoyu and Lin, Xudong and Roth, Dan and Smith, Noah A. and Ma, Wei-Chiu and Krishna, Ranjay},
title = {BLINK: Multimodal Large Language Models Can See but Not Perceive},
year = {2024},
isbn = {978-3-031-73336-9},
publisher = {Springer-Verlag},
address = {Berlin, Heidelberg},
url = {https://doi.org/10.1007/978-3-031-73337-6_9},
doi = {10.1007/978-3-031-73337-6_9},
booktitle = {Computer Vision – ECCV 2024: 18th European Conference, Milan, Italy, September 29–October 4, 2024, Proceedings, Part XXIII},
pages = {148–166},
numpages = {19},
location = {Milan, Italy}
}

@inproceedings{liu2024mmbench,
author = {Liu, Yuan and Duan, Haodong and Zhang, Yuanhan and Li, Bo and Zhang, Songyang and Zhao, Wangbo and Yuan, Yike and Wang, Jiaqi and He, Conghui and Liu, Ziwei and Chen, Kai and Lin, Dahua},
title = {MMBench: Is Your Multi-modal Model an All-Around Player?},
year = {2024},
isbn = {978-3-031-72657-6},
publisher = {Springer-Verlag},
address = {Berlin, Heidelberg},
url = {https://doi.org/10.1007/978-3-031-72658-3_13},
doi = {10.1007/978-3-031-72658-3_13},
booktitle = {Computer Vision – ECCV 2024: 18th European Conference, Milan, Italy, September 29–October 4, 2024, Proceedings, Part VI},
pages = {216–233},
numpages = {18},
location = {Milan, Italy}
}

@inproceedings{wei2022cot,
author = {Wei, Jason and Wang, Xuezhi and Schuurmans, Dale and Bosma, Maarten and Ichter, Brian and Xia, Fei and Chi, Ed H. and Le, Quoc V. and Zhou, Denny},
title = {Chain-of-thought prompting elicits reasoning in large language models},
year = {2022},
isbn = {9781713871088},
publisher = {Curran Associates Inc.},
address = {Red Hook, NY, USA},
booktitle = {Proceedings of the 36th International Conference on Neural Information Processing Systems},
articleno = {1800},
numpages = {14},
location = {New Orleans, LA, USA},
series = {NIPS '22}
}

@inproceedings{kojima2022zerocot,
author = {Kojima, Takeshi and Gu, Shixiang Shane and Reid, Machel and Matsuo, Yutaka and Iwasawa, Yusuke},
title = {Large language models are zero-shot reasoners},
year = {2022},
isbn = {9781713871088},
publisher = {Curran Associates Inc.},
address = {Red Hook, NY, USA},
booktitle = {Proceedings of the 36th International Conference on Neural Information Processing Systems},
articleno = {1613},
numpages = {15},
location = {New Orleans, LA, USA},
series = {NIPS '22}
}

@inproceedings{nie2020bongard,
author = {Nie, Weili and Yu, Zhiding and Mao, Lei and Patel, Ankit B. and Zhu, Yuke and Anandkumar, Animashree},
title = {BONGARD-LOGO: a new benchmark for human-level concept learning and reasoning},
year = {2020},
isbn = {9781713829546},
publisher = {Curran Associates Inc.},
address = {Red Hook, NY, USA},
booktitle = {Proceedings of the 34th International Conference on Neural Information Processing Systems},
articleno = {1382},
numpages = {13},
location = {Vancouver, BC, Canada},
series = {NIPS '20}
}

@inproceedings{zhang2019raven, 
    title={RAVEN: A Dataset for Relational and Analogical Visual rEasoNing}, 
    author={Zhang, Chi and Gao, Feng and Jia, Baoxiong and Zhu, Yixin and Zhu, Song-Chun}, 
    booktitle={Proceedings of the IEEE Conference on Computer Vision and Pattern Recognition (CVPR)}, 
    year={2019}
}

@misc{visulogic2024,
        title={VisuLogic: A Benchmark for Evaluating Visual Reasoning in Multi-modal Large Language Models}, 
      author={Weiye Xu and Jiahao Wang and Weiyun Wang and Zhe Chen and Wengang Zhou and Aijun Yang and Lewei Lu and Houqiang Li and Xiaohua Wang and Xizhou Zhu and Wenhai Wang and Jifeng Dai and Jinguo Zhu},
      year={2025},
      eprint={2504.15279},
      archivePrefix={arXiv},
      primaryClass={cs.CV},
      url={https://arxiv.org/abs/2504.15279}, 
}

@misc{gemmateam2025gemma3technicalreport,
      title={Gemma 3 Technical Report}, 
      author={Gemma Team and Aishwarya Kamath and Johan Ferret and Shreya Pathak and Nino Vieillard and Ramona Merhej and Sarah Perrin and Tatiana Matejovicova and Alexandre Ramé and Morgane Rivière and Louis Rouillard and Thomas Mesnard and Geoffrey Cideron and Jean-bastien Grill and Sabela Ramos and Edouard Yvinec and Michelle Casbon and Etienne Pot and Ivo Penchev and Gaël Liu and Francesco Visin and Kathleen Kenealy and Lucas Beyer and Xiaohai Zhai and Anton Tsitsulin and Robert Busa-Fekete and Alex Feng and Noveen Sachdeva and Benjamin Coleman and Yi Gao and Basil Mustafa and Iain Barr and Emilio Parisotto and David Tian and Matan Eyal and Colin Cherry and Jan-Thorsten Peter and Danila Sinopalnikov and Surya Bhupatiraju and Rishabh Agarwal and Mehran Kazemi and Dan Malkin and Ravin Kumar and David Vilar and Idan Brusilovsky and Jiaming Luo and Andreas Steiner and Abe Friesen and Abhanshu Sharma and Abheesht Sharma and Adi Mayrav Gilady and Adrian Goedeckemeyer and Alaa Saade and Alex Feng and Alexander Kolesnikov and Alexei Bendebury and Alvin Abdagic and Amit Vadi and András György and André Susano Pinto and Anil Das and Ankur Bapna and Antoine Miech and Antoine Yang and Antonia Paterson and Ashish Shenoy and Ayan Chakrabarti and Bilal Piot and Bo Wu and Bobak Shahriari and Bryce Petrini and Charlie Chen and Charline Le Lan and Christopher A. Choquette-Choo and CJ Carey and Cormac Brick and Daniel Deutsch and Danielle Eisenbud and Dee Cattle and Derek Cheng and Dimitris Paparas and Divyashree Shivakumar Sreepathihalli and Doug Reid and Dustin Tran and Dustin Zelle and Eric Noland and Erwin Huizenga and Eugene Kharitonov and Frederick Liu and Gagik Amirkhanyan and Glenn Cameron and Hadi Hashemi and Hanna Klimczak-Plucińska and Harman Singh and Harsh Mehta and Harshal Tushar Lehri and Hussein Hazimeh and Ian Ballantyne and Idan Szpektor and Ivan Nardini and Jean Pouget-Abadie and Jetha Chan and Joe Stanton and John Wieting and Jonathan Lai and Jordi Orbay and Joseph Fernandez and Josh Newlan and Ju-yeong Ji and Jyotinder Singh and Kat Black and Kathy Yu and Kevin Hui and Kiran Vodrahalli and Klaus Greff and Linhai Qiu and Marcella Valentine and Marina Coelho and Marvin Ritter and Matt Hoffman and Matthew Watson and Mayank Chaturvedi and Michael Moynihan and Min Ma and Nabila Babar and Natasha Noy and Nathan Byrd and Nick Roy and Nikola Momchev and Nilay Chauhan and Noveen Sachdeva and Oskar Bunyan and Pankil Botarda and Paul Caron and Paul Kishan Rubenstein and Phil Culliton and Philipp Schmid and Pier Giuseppe Sessa and Pingmei Xu and Piotr Stanczyk and Pouya Tafti and Rakesh Shivanna and Renjie Wu and Renke Pan and Reza Rokni and Rob Willoughby and Rohith Vallu and Ryan Mullins and Sammy Jerome and Sara Smoot and Sertan Girgin and Shariq Iqbal and Shashir Reddy and Shruti Sheth and Siim Põder and Sijal Bhatnagar and Sindhu Raghuram Panyam and Sivan Eiger and Susan Zhang and Tianqi Liu and Trevor Yacovone and Tyler Liechty and Uday Kalra and Utku Evci and Vedant Misra and Vincent Roseberry and Vlad Feinberg and Vlad Kolesnikov and Woohyun Han and Woosuk Kwon and Xi Chen and Yinlam Chow and Yuvein Zhu and Zichuan Wei and Zoltan Egyed and Victor Cotruta and Minh Giang and Phoebe Kirk and Anand Rao and Kat Black and Nabila Babar and Jessica Lo and Erica Moreira and Luiz Gustavo Martins and Omar Sanseviero and Lucas Gonzalez and Zach Gleicher and Tris Warkentin and Vahab Mirrokni and Evan Senter and Eli Collins and Joelle Barral and Zoubin Ghahramani and Raia Hadsell and Yossi Matias and D. Sculley and Slav Petrov and Noah Fiedel and Noam Shazeer and Oriol Vinyals and Jeff Dean and Demis Hassabis and Koray Kavukcuoglu and Clement Farabet and Elena Buchatskaya and Jean-Baptiste Alayrac and Rohan Anil and Dmitry and Lepikhin and Sebastian Borgeaud and Olivier Bachem and Armand Joulin and Alek Andreev and Cassidy Hardin and Robert Dadashi and Léonard Hussenot},
      year={2025},
      eprint={2503.19786},
      archivePrefix={arXiv},
      primaryClass={cs.CL},
      url={https://arxiv.org/abs/2503.19786}, 
}

@misc{lu2024mathvistaevaluatingmathematicalreasoning,
      title={MathVista: Evaluating Mathematical Reasoning of Foundation Models in Visual Contexts}, 
      author={Pan Lu and Hritik Bansal and Tony Xia and Jiacheng Liu and Chunyuan Li and Hannaneh Hajishirzi and Hao Cheng and Kai-Wei Chang and Michel Galley and Jianfeng Gao},
      year={2024},
      eprint={2310.02255},
      archivePrefix={arXiv},
      primaryClass={cs.CV},
      url={https://arxiv.org/abs/2310.02255}, 
}

@misc{zhang2024mathversedoesmultimodalllm,
      title={MathVerse: Does Your Multi-modal LLM Truly See the Diagrams in Visual Math Problems?}, 
      author={Renrui Zhang and Dongzhi Jiang and Yichi Zhang and Haokun Lin and Ziyu Guo and Pengshuo Qiu and Aojun Zhou and Pan Lu and Kai-Wei Chang and Peng Gao and Hongsheng Li},
      year={2024},
      eprint={2403.14624},
      archivePrefix={arXiv},
      primaryClass={cs.CV},
      url={https://arxiv.org/abs/2403.14624}, 
}

@misc{lu2022learnexplainmultimodalreasoning,
      title={Learn to Explain: Multimodal Reasoning via Thought Chains for Science Question Answering}, 
      author={Pan Lu and Swaroop Mishra and Tony Xia and Liang Qiu and Kai-Wei Chang and Song-Chun Zhu and Oyvind Tafjord and Peter Clark and Ashwin Kalyan},
      year={2022},
      eprint={2209.09513},
      archivePrefix={arXiv},
      primaryClass={cs.CL},
      url={https://arxiv.org/abs/2209.09513}, 
}

@inproceedings{johnson2017clevr,
  title={CLEVR: A Diagnostic Dataset for Compositional Language and Elementary Visual Reasoning},
  author={Johnson, Justin and Hariharan, Bharath and van der Maaten, Laurens and Fei-Fei, Li and Zitnick, C Lawrence and Girshick, Ross},
  booktitle={CVPR},
  year={2017}
}

@inproceedings{hudson2019gqa,
  title={{GQA}: A New Dataset for Real-World Visual Reasoning and Compositional Question Answering},
  author={Hudson, Drew A and Manning, Christopher D},
  booktitle={CVPR},
  year={2019}
}

@inproceedings{lu2023mathvista,
  author    = {Lu, Pan and Bansal, Hritik and Xia, Tony and Liu, Jiacheng and Li, Chunyuan and Hajishirzi, Hannaneh and Cheng, Hao and Chang, Kai-Wei and Galley, Michel and Gao, Jianfeng},
  title     = {MathVista: Evaluating Mathematical Reasoning of Foundation Models in Visual Contexts},
  booktitle={International Conference on Learning Representations (ICLR)},
  year      = {2024}
}

@inproceedings{yue2023mmmu,
            title={MMMU: A Massive Multi-discipline Multimodal Understanding and Reasoning Benchmark for Expert AGI},
            author={Xiang Yue and Yuansheng Ni and Kai Zhang and Tianyu Zheng and Ruoqi Liu and Ge Zhang and Samuel Stevens and Dongfu Jiang and Weiming Ren and Yuxuan Sun and Cong Wei and Botao Yu and Ruibin Yuan and Renliang Sun and Ming Yin and Boyuan Zheng and Zhenzhu Yang and Yibo Liu and Wenhao Huang and Huan Sun and Yu Su and Wenhu Chen},
            booktitle={Proceedings of CVPR},
            year={2024},
}

@InProceedings{li2023seed,
    author    = {Li, Bohao and Ge, Yuying and Ge, Yixiao and Wang, Guangzhi and Wang, Rui and Zhang, Ruimao and Shan, Ying},
    title     = {SEED-Bench: Benchmarking Multimodal Large Language Models},
    booktitle = {Proceedings of the IEEE/CVF Conference on Computer Vision and Pattern Recognition (CVPR)},
    month     = {June},
    year      = {2024},
    pages     = {13299-13308}
}

@InProceedings{li2023seed2,
      title={SEED-Bench-2: Benchmarking Multimodal Large Language Models}, 
      author={Bohao Li and Yuying Ge and Yixiao Ge and Guangzhi Wang and Rui Wang and Ruimao Zhang and Ying Shan},
      year={2023},
      eprint={2311.17092},
      archivePrefix={arXiv},
      primaryClass={cs.CV},
      url={https://arxiv.org/abs/2311.17092}, 

}

@article{shepard1971mental,
  title={Mental Rotation of Three-Dimensional Objects},
  author={Shepard, Roger N and Metzler, Jacqueline},
  journal={Science},
  year={1971}
}

@article{mindthegap2025,
  title={Mind the Gap: Benchmarking Spatial Reasoning in Vision‐Language Models},
  author={Stogiannidis, Ilias and McDonagh, Steven and Tsaftaris, Sotirios A.},
  journal={arXiv preprint arXiv:2503.19707},
  year={2025},
  url={https://arxiv.org/abs/2503.19707}
}

@inproceedings{li2023blip2,
  title={{BLIP-2}: Bootstrapping Language-Image Pre-training with Frozen Image Encoders and Large Language Models},
  author={Li, Junnan and Li, Dongxu and Xiong, Caiming and Hoi, Steven CH},
  booktitle={ICML},
  year={2023}
}

@article{mindthegap2024,
  title={{Mind the Gap}: Benchmarking Spatial Reasoning in Vision-Language Models},
  author={Stogiannidis, Ilias and McDonagh, Steven and Tsaftaris, Sotirios A},
  journal={arXiv preprint arXiv:2403.19707},
  year={2024}
}

@misc{doyouseeme2024,
   title={Do You See Me : A Multidimensional Benchmark for Evaluating Visual Perception in Multimodal LLMs}, 
      author={Aditya Kanade and Tanuja Ganu},
      year={2025},
      eprint={2506.02022},
      archivePrefix={arXiv},
      primaryClass={cs.CV},
      url={https://arxiv.org/abs/2506.02022},
}

@article{vandenberg1978mrt,
  title={Mental Rotations, a Group Test of Three-Dimensional Spatial Visualization},
  author={Vandenberg, Steven G and Kuse, Allan R},
  journal={Perceptual and Motor Skills},
  volume={47},
  number={2},
  pages={599--604},
  year={1978}
}

@book{cogat2009paperfolding,
  title={Cognitive Abilities Test (CogAT) Form 7, Paper Folding Items},
  author={Riverside Publishing},
  publisher={Riverside},
  year={2009},
  note={Psychometric test battery}
}

@book{hoffstadter1979godel,
  title={G{\"o}del, Escher, Bach: An Eternal Golden Braid},
  author={Hofstadter, Douglas R},
  year={1979},
  publisher={Basic Books}
}

@inproceedings{yu2023mmvet,
author = {Yu, Weihao and Yang, Zhengyuan and Li, Linjie and Wang, Jianfeng and Lin, Kevin and Liu, Zicheng and Wang, Xinchao and Wang, Lijuan},
title = {MM-Vet: evaluating large multimodal models for integrated capabilities},
year = {2024},
publisher = {JMLR.org},
booktitle = {Proceedings of the 41st International Conference on Machine Learning},
articleno = {2381},
numpages = {25},
location = {Vienna, Austria},
series = {ICML'24}
}

@inproceedings{gadz2024visulogic,
  title = {VisuLogic: A Benchmark for Evaluating Visual Reasoning in Multi-modal Large Language Models},
  author = {Xu, Weiye and Wang, Jiahao and Wang, Weiyun and Chen, Zhe and Zhou, Wengang and Yang, Aijun and Lu, Lewei and Li, Houqiang and Wang, Xiaohua and Zhu, Xizhou and Wang, Wenhai and Dai, Jifeng and Zhu, Jinguo},
  year = {2025},
  journal = {arXiv preprint arXiv:2504.15279}
}

@article{grit2023,
  title = {GRIT: Teaching MLLMs to Think with Images},
  author = {Fan, Yue and He, Xuehai and Yang, Diji and Zheng, Kaizhi and Kuo, Ching-Chen and Zheng, Yuting and Narayanaraju, Sravana~Jyothi and Guan, Xinze and Wang, Xin~Eric},
  year = {2025},
  journal = {arXiv preprint arXiv:2505.15879}
}

@article{arcagi2019,
  title={On the Measure of Intelligence},  
  author={Chollet, François},
  journal={arXiv preprint arXiv:1911.01547},
  year={2019},
  url={https://arxiv.org/abs/1911.01547}
}
\clearpage
\appendix

\section*{Appendix}
We present the following additional results and discussions which we could not include in the main paper owing to space constraints:
\begin{itemize}
    \item[\textbf{A}] Additional Results 
    \item[\textbf{B}] Extended Analysis
    \item[\textbf{C}] More about \textsc{Mind's Eye}
    \item[\textbf{D}] Detailed Benchmark Design
    \item[\textbf{E}] Evaluation Setup Details
    \item[\textbf{F}] Human Evaluation Protocol
    \item[\textbf{G}] Prompting Strategies and Styles
    \item[\textbf{H}] Full Prompt Templates
    \item[\textbf{I}] Analysis of CoT Reasoning Quality
    \item[\textbf{J}] Carroll's Three-Stratum Theory of Fluid Intelligence
    \item[\textbf{J}] Performance comparison of humans and models across cognitive subtasks by difficulty level.
    
\end{itemize}


\section{Additional Results}
\label{app:additional_results}

\begin{figure}[h!] 
    \centering
    \includegraphics[width=\columnwidth]{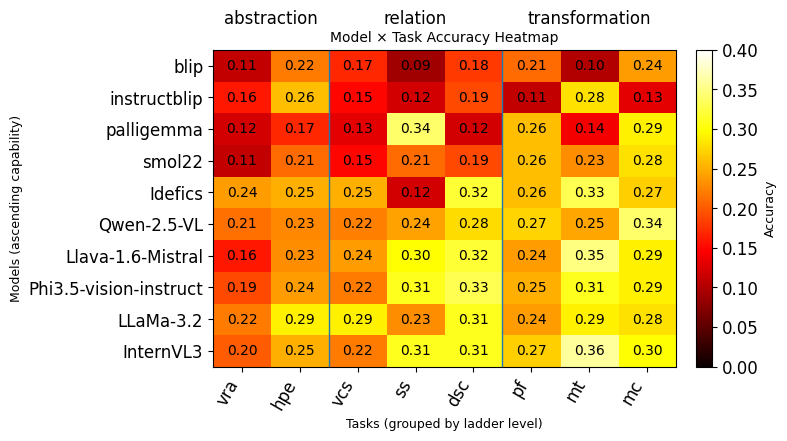}
    \vspace{-15pt}
    \caption{\textbf{Model performance across all tasks}: Heatmap of model performance across tasks, with rows denoting models and columns denoting tasks (color intensity represents accuracy). Models ordered by increasing capability (top to bottom), and tasks grouped by ART, revealing that even top-tier models have significant/varied weaknesses.
}
    \label{fig:model_task_heatmap}
\end{figure}

\begin{figure*}[t] 
    \centering
    \begin{subfigure}[t]{\textwidth}
        \centering
        \includegraphics[width=\linewidth]{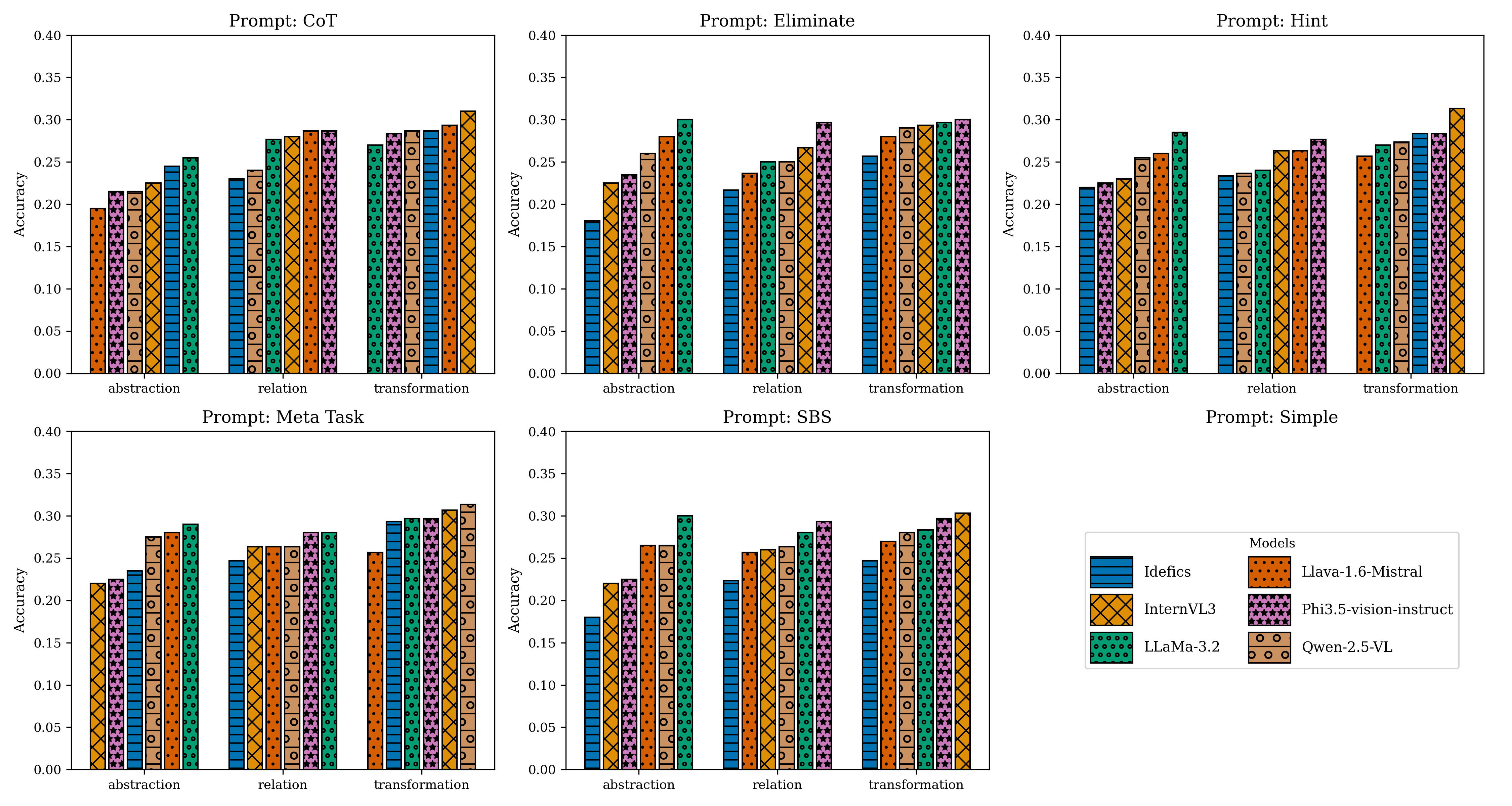}
        \caption{Skill-level bar chart showing average model performance across the three cognitive levels (Transformation, Relation, Abstraction). Tasks are grouped by level, and bars indicate per model averages.}
        \label{fig:skill_level_bar}
    \end{subfigure}
    \vspace{2em}
    \begin{subfigure}[t]{\textwidth}
        \centering
        \includegraphics[width=\linewidth]{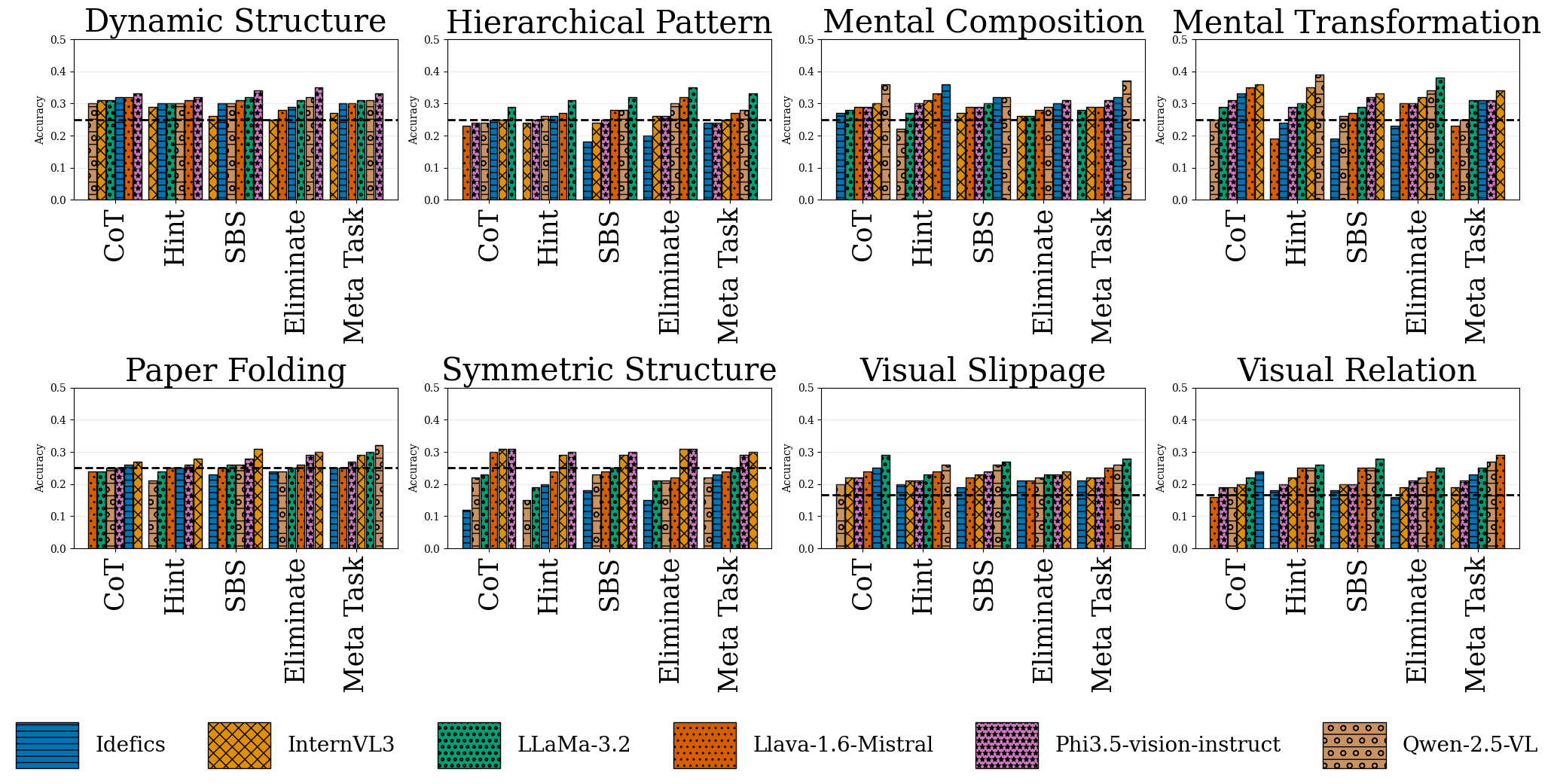}
        \caption{Effect of different prompting styles (CoT, Hint, Meta) on per task performance. The dotted line denotes the random choice baseline for each task.}
        \label{fig:prompt_effect}
    \end{subfigure}

    \caption{\textbf{Model Performance and Prompting Effects} : (a) Average model performance across cognitive skill levels. (b) Prompting style effects on task wise performance. Together, these plots illustrate the inconsistent and often modest impact of different prompting styles (CoT, Hint, Meta) on per task performance}
    \label{fig:combined_bars}
\end{figure*}

\begin{figure*}[t] 
    \centering
    \includegraphics[width=\columnwidth]{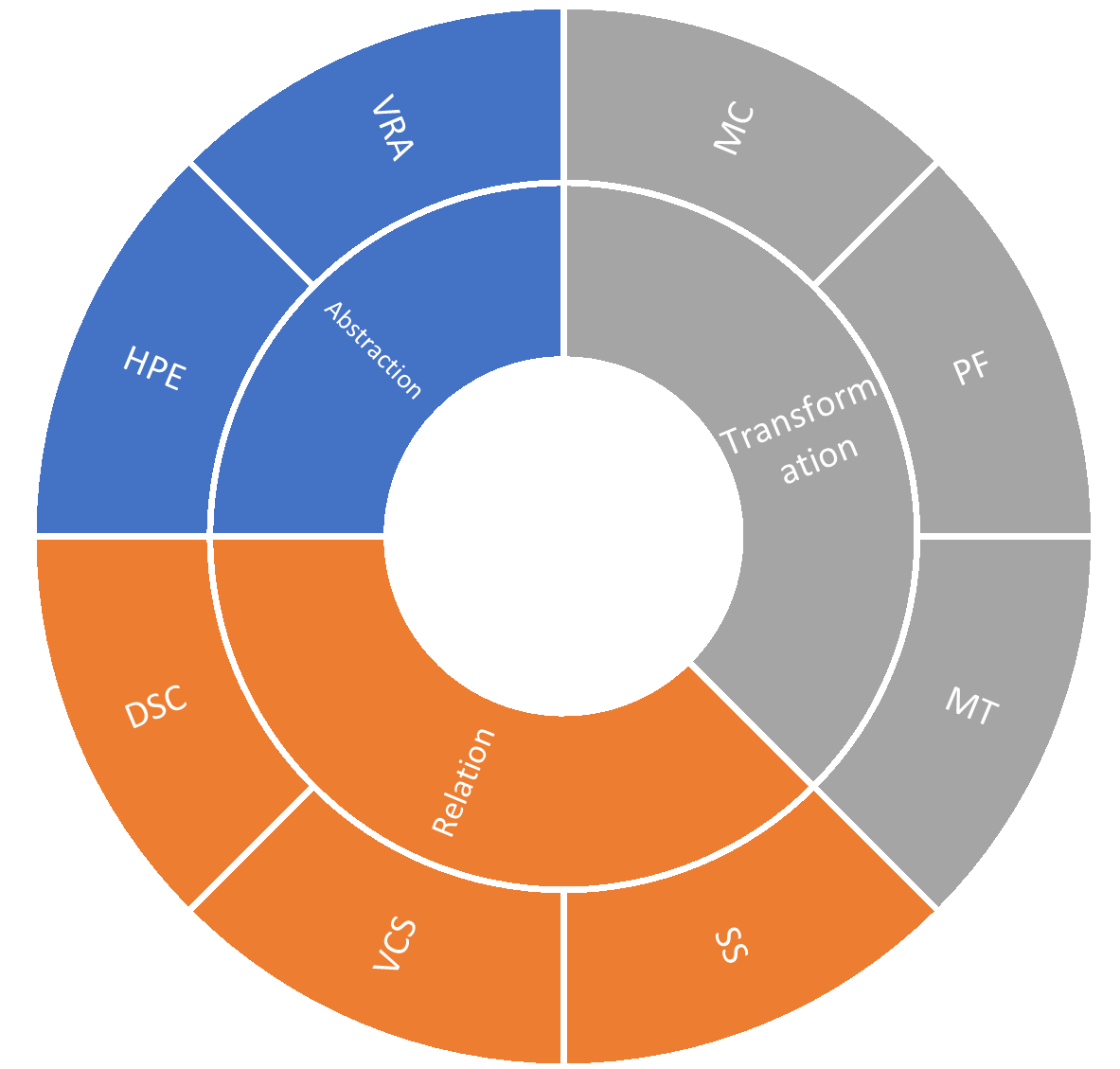}
    \caption{\textbf{Dataset Distribution as per the ART Framework}: The inner ring represents the three A-R-T cognitive categories, while the outer ring shows the eight specific tasks and their alignment within this framework.}
    \label{fig:dataset_distribution}
\end{figure*}
\textbf{Benchmark Distribution} Our benchmark consists of eight tasks: Dynamic Structural Correspondence, Hierarchical Pattern Equivalence, Mental Composition, Mental Transformation, Paper Folding, Visual Conceptual Slippage, Symmetric Structures, and Visual Relation Abstraction. These tasks are grouped into three categories that align with core dimensions of fluid intelligence: Pattern Abstraction (Visual Relation Abstraction, Hierarchical Pattern Equivalence), Relation (Dynamic Structural Correspondence, Visual Conceptual Slippage, Symmetric Structures), and Transformation (Mental Transformation, Paper Folding, Mental Composition). Each task is programmatically generated \footnote{Code: \url{https://anonymous.4open.science/r/Minds_Eye-0801/}}, allowing precise control over difficulty by varying task specific parameters. Figure \ref{fig:dataset_distribution} illustrates the dataset distribution employed for probing and evaluating model performance.

\begin{figure*}[t] 
\begin{subfigure}[t]{0.48\linewidth}
    \centering
    \includegraphics[width=\linewidth]{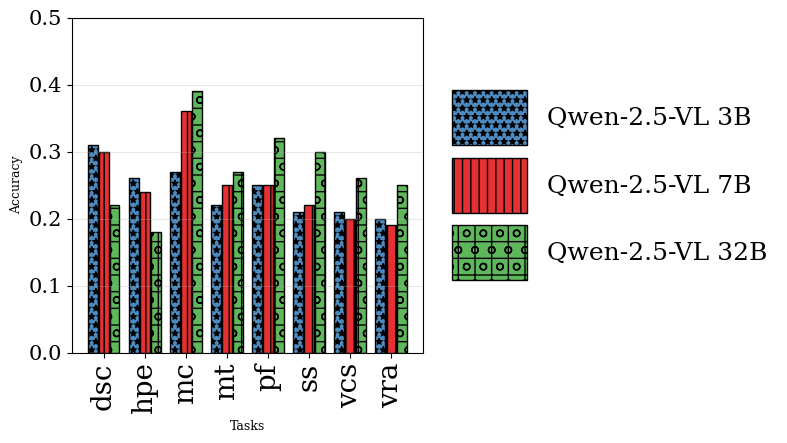}
    \caption{\textbf{Accuracy of Qwen-2.5-VL models (3B, 7B, 32B) across ART tasks} : The performance comparison of Qwen-2.5-VL models of three different sizes (3B, 7B, 32B) across the eight tasks. It highlights that scaling provides non uniform gains, with larger models improving on some tasks but not uniformly across all the tasks, reinforcing that scale alone is insufficient to overcome the reasoning deficits.
}
    \label{fig:qwenscaling}
\end{subfigure}
\hfill
\begin{subfigure}[t]{0.48\linewidth}
    \centering
    \includegraphics[width=\linewidth]{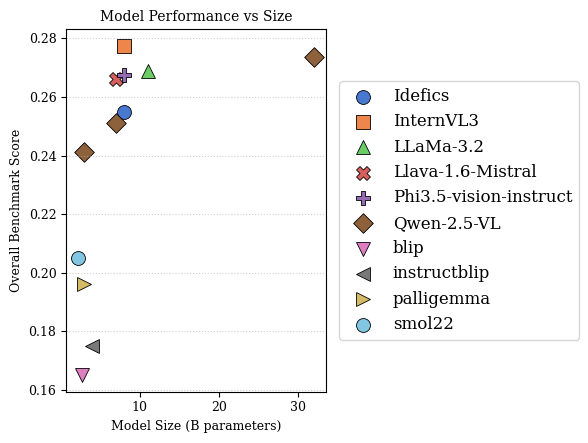}
    \caption{\textbf{Model performance versus size on our benchmark}. While larger models (e.g., Qwen-2.5-VL-32B) achieve strong results, several medium sized models (InternVL3, LLaMA-3.2, Phi-3.5) match or exceed them, indicating that scaling alone is insufficient and that training design and architecture critically influence cognitive reasoning performance.}
    \label{fig:scatter_score_vs_size}
\end{subfigure}
\caption{\textbf{Impact of Scale }: (a) and (b) provides a compelling evidence that scaling alone is not sufficient enough to improve performance on this benchmark}
\end{figure*}

We compare overall benchmark performance against model size (in billions of parameters) across a diverse set of multimodal models in Figure \ref{fig:scatter_score_vs_size}. Interestingly, performance does not scale monotonically with size: some medium scale models (e.g., InternVL3, LLaMA-3.2, Phi-3.5) achieve competitive or even superior performance relative to much larger counterparts, while smaller models (e.g., BLIP, InstructBLIP, PaliGemma) consistently underperform. Notably, Qwen-2.5-VL exhibits strong performance at both small and large scales, suggesting architectural and training choices play a larger role than raw parameter count. A correlation analysis confirms this observation, with Pearson’s $r \approx 0.62$, indicating only a moderately positive relationship between model size and benchmark performance. 
Taken together, these results highlight that \emph{scaling} yields non uniform gains across our tasks, suggesting that parameter growth alone may not suffice under this benchmark, and that improved training and architecture could be equally important.

\begin{figure*}[htbp] 
    \centering
    \includegraphics[width=2.2\columnwidth]{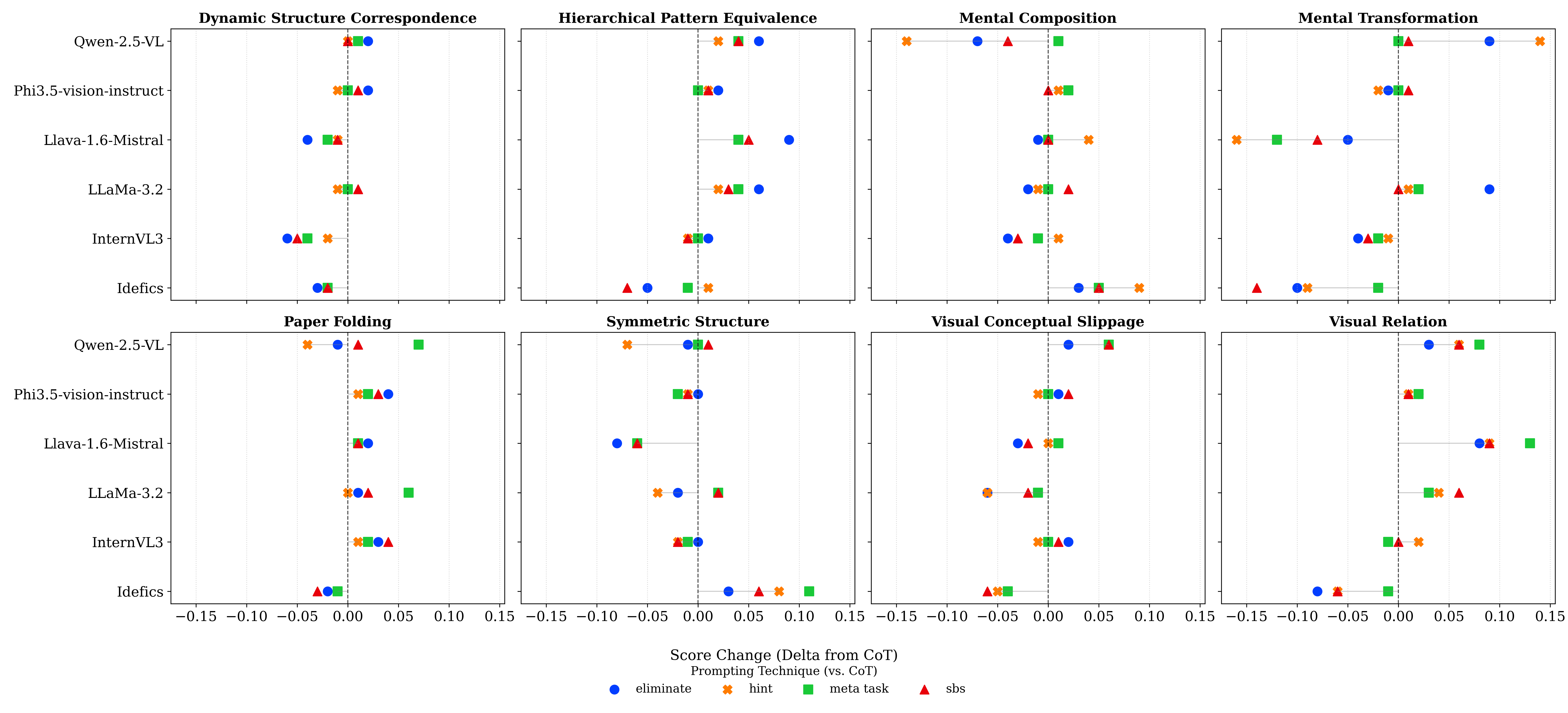}
            \caption{\textbf{Relative effect of prompting strategies versus chain-of-thought (CoT) across tasks}. Points to the left of the dashed line indicate performance deterioration, while those to the right indicate improvement. Meta-task and step-by-step (SBS)  prompts often improve tasks like \textit{Hierarchical Pattern Equivalence, Visual Relation, Paper Folding}, but abstraction tasks like \textit{Symmetric Structure, Mental Composition, Mental Transformation, Visual Conceptual Slippage} show consistent declines. Prompting strategies therefore exert strongly task dependent effects, with no universally reliable method for improving performance.}
    \label{fig:delta_per_task}
\end{figure*}

\textbf{Prompting strategies performance deltas :} We compare the effect of four prompting strategies (\textit{eliminate, hint, meta-task, step-by-step}) against chain-of-thought (CoT) across the eight benchmark tasks in Figure~\ref{fig:delta_per_task}. Each subplot shows the relative score change, where values to the left of the vertical dashed line indicate deterioration and values to the right indicate improvement. The results reveal a heterogeneous landscape: 
\begin{itemize}
    \item \textbf{Consistent improvements:} Tasks such as \textit{Hierarchical Pattern Equivalence} and \textit{Visual Relation} benefit from meta-task, eliminate and step-by-step prompts, which appear to help models engage in multistep reasoning.  
    \item \textbf{Mixed or task dependent effects:} Tasks like \textit{Dynamic Structural Correspondence},\textit{Visual  Conceptual Slippage} and \textit{Paper Folding} show both gains and regressions depending on the prompting strategy.  
    \item \textbf{Clear deterioration:} Cognition heavy tasks such as \textit{Mental Transformation} and \textit{Mental Composition} exhibit consistent performance drops across most prompting strategies relative to CoT.  
    \item \textbf{Instability of eliminate and hint:} These strategies occasionally yield benefits, but more often result in deterioration across tasks.  
\end{itemize}
Overall, the figure highlights that while prompting can produce gains in reasoning intensive tasks, it can also worsen performance in many tasks, underscoring the lack of a universally beneficial prompting strategy. 

Figure~\ref{fig:qwenscaling} shows the performance of Qwen-2.5-VL models of different sizes (3B, 7B, 32B) on the ART benchmark under chain-of-thought (CoT) prompting. Several patterns emerge. First, scaling does not yield uniform improvements across tasks: while the 32B variant outperforms the smaller models on conceptual relation and transformation heavy tasks such as Mental Transformation, Mental Composition and Paper Folding, the smaller 3B and 7B variants remain competitive or superior on temporal relation and abstraction oriented tasks like Dynamic Structural Correspondence and Hierarchical Pattern Equivalence. This reinforces that scale alone is insufficient to overcome reasoning deficits, and that certain tasks demand structured cognitive mechanisms rather than brute force capacity. Second, Mental Composition is a particularly challenging task, where both 7B and 32B improve substantially over 3B, yet overall accuracy remains low, reflecting the persistent difficulty of compositional reasoning. Finally, we note a strong dependency on perceptual similarity in some tasks: while larger models exploit surface level cues more effectively (in Paper Folding, Symmetric Structures, Visual Conceptual Slippage, Visual Relational Abstraction), they continue to fail when success requires internal simulation of transformations.

Taken together, these results highlight that while larger models can achieve gains in perception heavy reasoning tasks, smaller models sometimes generalize better in abstraction oriented settings. This suggests that scaling amplifies perceptual alignment but does not induce the higher level grouping or cognitive mechanisms required for robust visuo-cognitive reasoning.

\begin{figure}[H] 
\centering
    \centering
    \includegraphics[width=\columnwidth]{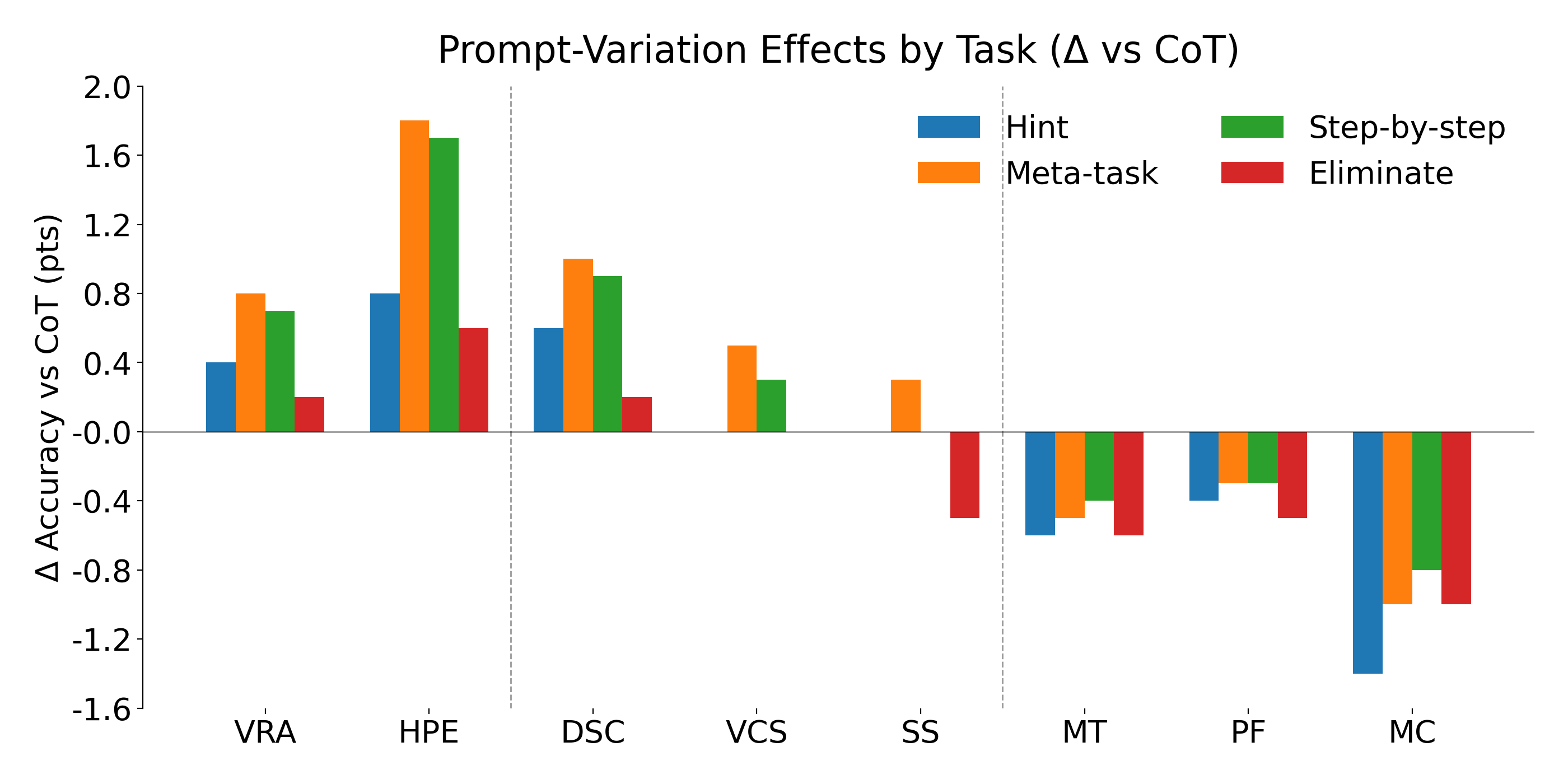}
    \caption{\textbf{Differential Effects of prompting across ART dimensions:} Average accuracy of models relative to CoT performance aggregated by Abstraction, Relation, and Transformation.
}
\label{fig:prompt_deltas_by_task}
\end{figure}

Figure \ref{fig:prompt_deltas_by_task} presents a task-level analysis of prompt variation effects relative to chain-of-thought (CoT), revealing substantial heterogeneity within each ART dimension. Within Abstraction, Hierarchical Pattern Equivalence (HPE) is the most prompt-sensitive task, with meta-task and step-by-step prompting yielding gains of approximately +1.8 pts and +1.7 pts, respectively, while Visual Relation Abstraction (VRA) shows smaller improvements (+0.4–0.8 pts) and remains substantially less sensitive to prompt framing. Relation tasks exhibit the greatest internal variability: Dynamic Structural Correspondence (DSC) benefits from structured prompting, achieving gains of up to +1.0 pt under meta-task prompting, whereas Visual Conceptual Slippage (VCS) remains near-invariant ($\leq$ +0.3 pts across all prompts), and Symmetric Structures (SS) shows mixed behavior, with a modest gain under meta-task prompting ($\approx$ +0.3 pts) but degradation under elimination-based prompting ($\approx$ -0.5 pts). In contrast, Transformation tasks are uniformly brittle to prompt variation. Mental Composition (MC) exhibits the largest and most consistent drops across all prompting strategies (-0.8 to -1.4 pts), followed by Mental Transformation (MT) ($\approx$ -0.4 to -0.6 pts), while Paper Folding (PF) shows comparatively smaller declines ($\approx$ -0.3 to -0.5 pts). Importantly, no transformation task shows systematic improvement under any alternative prompting strategy. These task-level dissociations indicate that prompting primarily benefits tasks requiring explicit symbolic rule induction (e.g., HPE, DSC), while consistently disrupting tasks that depend on multi-step internal visual simulation (e.g., MC, MT), reinforcing that prompt engineering modulates surface reasoning behavior but does not address the underlying transformation bottleneck identified by the ART framework.

\begin{figure}[H] 
    \centering
    \includegraphics[width=0.8\linewidth]{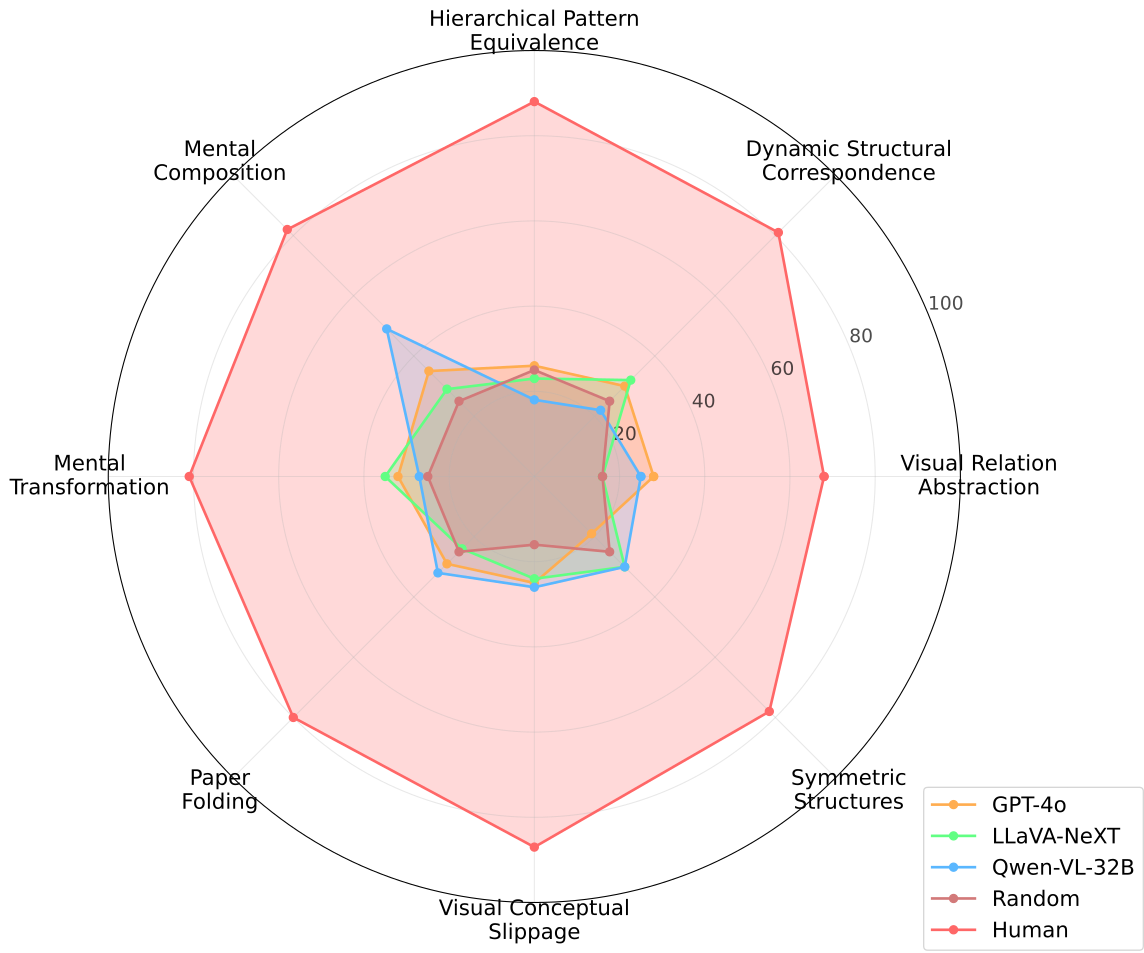}
    \caption{Accuracies of multimodal LLMs on \textsc{Mind's Eye} Benchmark. Please refer to \ref{tab:results} for more results and discussions.}
    \label{fig:spider}
\end{figure}

\paragraph{Effect of Image Resolution.}
A natural question is whether the performance gap we observe could be attributed to visual quality rather than reasoning limitations. To test this, we varied image resolution between 100\,DPI (600$\times$800\,px) and 300\,DPI (1024$\times$1024\,px) and evaluated Qwen-2.5-VL-7B across all eight tasks. As shown in Table~\ref{tab:resolution_ablation}, no statistically significant difference was observed at either resolution, suggesting that image quality is not a bottleneck for model performance on our benchmark. We note that all stimuli provided to models during evaluation are rendered as SVGs and exported at 1024$\times$1024\,px at 300\,DPI, ensuring that option labels and geometric details are fully legible at inference time.

\begin{table*}[t!]
\centering
\small
\caption{Resolution ablation on Qwen-2.5-VL-7B. No statistically significant difference is observed across the two settings.}
\label{tab:resolution_ablation}
\begin{tabular}{lcccccccc}
\toprule
Resolution & VRA & HPE & DSC & VCS & SS & MT & PF & MC \\
\midrule
100\,DPI / 600$\times$800 & 18.7{$\pm$0.02} & 24.4{\scriptsize$\pm$0.05} & 30.1{\scriptsize$\pm$0.08} & 22.1{\scriptsize$\pm$0.02} & 20.7{\scriptsize$\pm$0.05} & 25.2{\scriptsize$\pm$0.30} & 24.8{\scriptsize$\pm$0.10} & 36.1{\scriptsize$\pm$0.40} \\
300\,DPI / 1024$\times$1024 & 19.1{\scriptsize$\pm$0.01} & 24.2{\scriptsize$\pm$0.01} & 30.4{\scriptsize$\pm$0.01} & 22.7{\scriptsize$\pm$0.01} & 20.2{\scriptsize$\pm$0.04} & 25.7{\scriptsize$\pm$0.02} & 25.1{\scriptsize$\pm$0.02} & 36.4{\scriptsize$\pm$0.01} \\
\bottomrule
\end{tabular}
\end{table*}

\section{Extended Analysis}
\label{app:extended_analysis}
\subsection{Attention Maps}
\label{app:attentionmaps}

\begin{figure}[t] 
    \centering
    \includegraphics[width=\columnwidth]{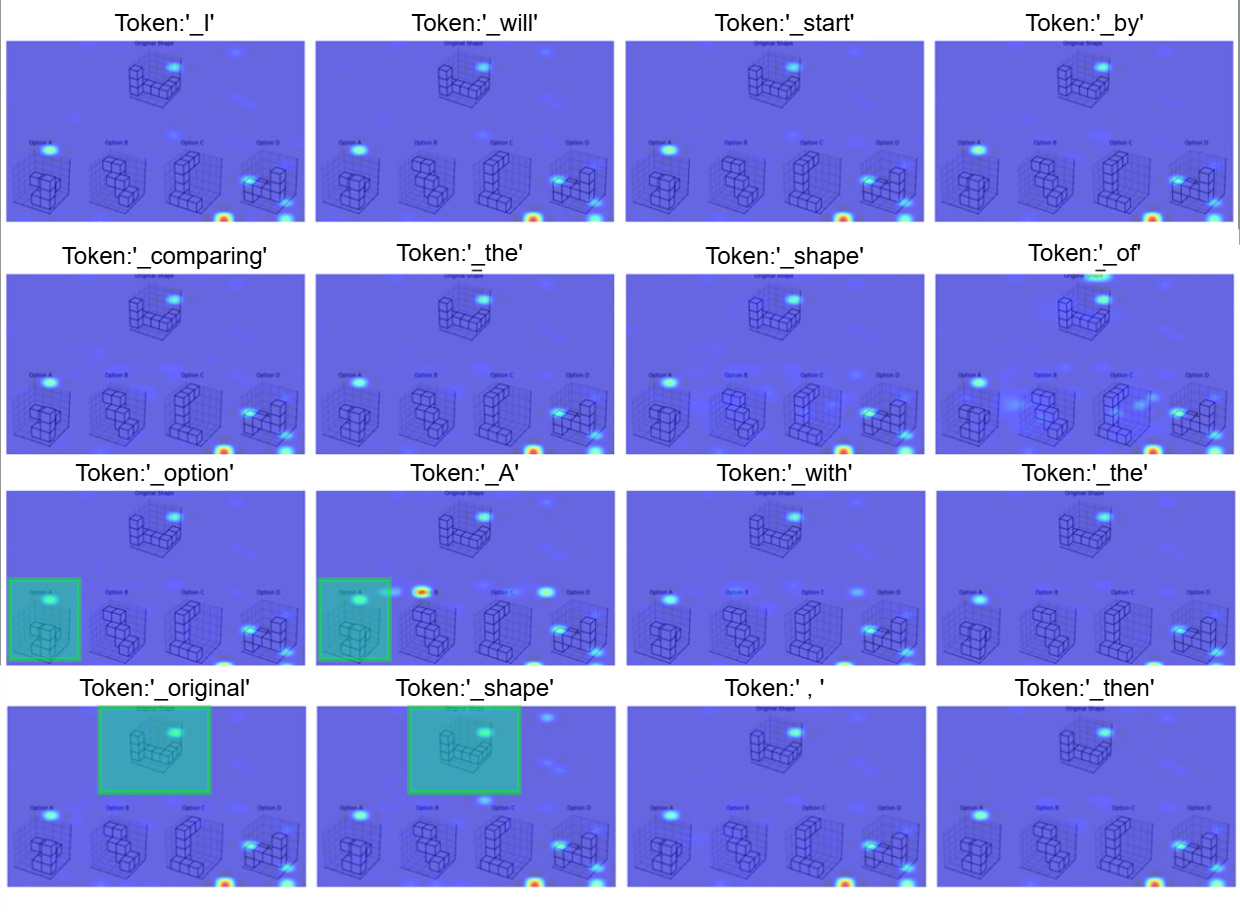}
    \caption{\textbf{Misplaced model attention} : Attention map of the LLaVa-7B model for the Mental Transformation Task. The green boxes shows the expected regions of attention.}
    \label{fig:attention_maps}
\end{figure}

\paragraph{Attention Heatmap Analysis.}  
To further probe the internal reasoning failures, we analyzed the attention heatmaps 
Surprisingly, when tokens explicitly referred to specific figures or options (e.g., ``option A'', ``shape'', ``comparing''), the model’s attention was not concentrated on the corresponding visual regions. Instead, attention was diffusely spread across background areas or unrelated parts of the image.  Across sampled instances ($N=20$ for each tasks, where N is the sampled CoT Traces), less than 20\% of the model’s normalized attention mass was directed toward the objects explicitly referenced in the reasoning trace.  
This failure mode highlights a limitation: although the model produces fluent chain-of-thought reasoning, the underlying attention does not ground the reasoning process in the visual input. In other words, query specific tokens fail to anchor attention to the corresponding figures, undermining the fidelity of the reasoning process. These findings reinforce our broader conclusion that current models may rely more on linguistic priors than grounded visual attention when tackling cognitive reasoning tasks.  

\paragraph{Grounded Attention Metric.}
We quantify grounding with \emph{Region Aligned Attention} (RAA):
$\text{RAA} = \frac{1}{T}\sum_{t=1}^{T} \sum_{p \in \mathcal{R}(t)} a_{t}(p)$,
where $a_t(p)$ is the normalized attention over pixels (or patches) at token $t$, and $\mathcal{R}(t)$ is the union of regions referenced by token $t$ (e.g., “option~A”, “shape”, “compare”).
We evaluate RAA on \textit{N}=$20$ tokens for which perception dependence is high (e.g. Option A, shape, color) across 20 items sampled stratified by task; mean RAA is \textbf{0.18}, corroborating the qualitative observation that the attention of query specific tokens often fails to align precisely with the corresponding figures. 

\subsection{Attention Performance Correlation Analysis}
\label{app:attention_performance}

\paragraph{Option-Specific Attention Score (OAS)}

For each item $i$ with $K$ options $\{O_1,\dots,O_K\}$, we define the \emph{Option-Specific Attention Score} (OAS) as the average normalized attention mass allocated to an option’s spatial region during reasoning token generation. Let $a_t(p)$ denote normalized attention at reasoning token $t$ to visual patch $p$, with $\sum_p a_t(p)=1$. Let $\mathcal{T}_{\text{reason}}$ denote reasoning tokens and $\mathcal{R}_k$ the patch set corresponding to option $O_k$. Then:
\begin{equation}
\mathrm{OAS}_k(i)=\frac{1}{|\mathcal{T}_{\text{reason}}|}\sum_{t\in\mathcal{T}_{\text{reason}}}\sum_{p\in\mathcal{R}_k} a_t(p).
\end{equation}
This metric quantifies the average proportion of attention allocated to option $k$ across all reasoning tokens. For our analysis, we compute three variants:
We analyze $\mathrm{OAS}_{\text{correct}}$, $\mathrm{OAS}_{\text{selected}}$, and mean $\mathrm{OAS}_{\text{distractors}}$.

\paragraph{Correlation and Trend Analysis}

Let $y_i\in\{0,1\}$ denote correctness. We compute the point-biserial correlation between $\mathrm{OAS}_{\text{correct}}$ and $y_i$:
\begin{equation}
r_{pb}=\frac{\bar{x}_1-\bar{x}_0}{s_x}\sqrt{\frac{n_1n_0}{n(n-1)}},
\end{equation}
where:
\begin{itemize}
    \item $\bar{x}_1 = \frac{1}{n_1} \sum_{i: y_i=1} \text{OAS}_{\text{correct}}(i)$ is the mean OAS for correct predictions
    \item $\bar{x}_0 = \frac{1}{n_0} \sum_{i: y_i=0} \text{OAS}_{\text{correct}}(i)$ is the mean OAS for incorrect predictions
    \item $s_x$ is the standard deviation of OAS across all items
    \item $n_1 = \sum_i y_i$ and $n_0 = n - n_1$ are the counts of correct and incorrect predictions
    \item $n$ is the total number of items analyzed
\end{itemize}

\paragraph{Paired Attention Comparisons}

For correct predictions, we test whether attention favors correct options over distractors:
\begin{equation}
H_0:\mathbb{E}[\mathrm{OAS}_{\text{correct}}-\mathrm{OAS}_{\text{distractors}}]=0,
\end{equation}
using paired $t$-tests. For incorrect predictions, we compare $\mathrm{OAS}_{\text{selected}}$ against $\mathrm{OAS}_{\text{correct}}$ to assess attention misallocation.

\paragraph{Implementation and Sampling}

Cross-attention weights are extracted from the final decoder layer and averaged across heads. Option regions are defined via fixed or programmatic bounding boxes depending on task layout. Analyses are conducted on $N=200$ items (25 per task).

\begin{table}[h]
\centering
\small
\begin{tabular}{lcc}
\toprule
\textbf{Analysis} & \textbf{Statistic} & \textbf{Result} \\
\midrule
Point-biserial correlation & $r_{pb}$ & 0.34 (p < 0.001) \\
\midrule
\multicolumn{3}{l}{\textit{Paired Comparisons (Correct Predictions, $n=87$)}} \\
\quad OAS$_{\text{correct}}$ & Mean & 0.24\scriptsize$\pm$0.08 \\
\quad OAS$_{\text{distractors}}$ & Mean & 0.16\scriptsize$\pm$0.06 \\
\quad Paired t-test & $t(86)$ & 4.32 (p $<$ 0.001) \\
\midrule
\multicolumn{3}{l}{\textit{Paired Comparisons (Incorrect Predictions, $n=113$)}} \\
\quad OAS$_{\text{selected}}$ & Mean & 0.18\scriptsize$\pm$0.07 \\
\quad OAS$_{\text{correct}}$ & Mean & 0.17\scriptsize$\pm$0.07 \\
\quad Paired t-test & $t(112)$ & 0.84 (p = 0.40) \\
\bottomrule
\end{tabular}
\caption{Statistical results from attention-performance correlation analysis. All p-values are two-tailed except where noted.}
\label{tab:attention_stats}
\end{table}

\subsection{Interpretation}

The positive point-biserial correlation ($r_{pb} = 0.34$) provide convergent evidence that attention alignment to correct options is a significant predictor of task performance. However, the modest effect size and low absolute accuracy even in the highest attention quartile (35.7\% vs. $>80\%$ human performance) indicate that attention is \textit{necessary but insufficient} for correct reasoning.

The paired comparison results reveal a critical asymmetry: when models answer correctly, they allocate significantly more attention to correct options than distractors (Cohen's $d = 1.15$, large effect). However, when models err, their attention to the selected (incorrect) option is statistically indistinguishable from attention to the correct option ($p = 0.40$), suggesting that errors arise from \textit{attention misallocation} rather than systematic biases away from correct answers. This pattern is consistent with a model that lacks robust visual grounding: it attends to plausible options without the cognitive mechanisms to reliably distinguish correctness from perceptual similarity.

\subsection{Pairwise Analysis of CoT vs Non CoT of Same Model Family}
\label{app:cotvncot}

\begin{figure}[H] 
    \centering
    \includegraphics[width=\columnwidth]{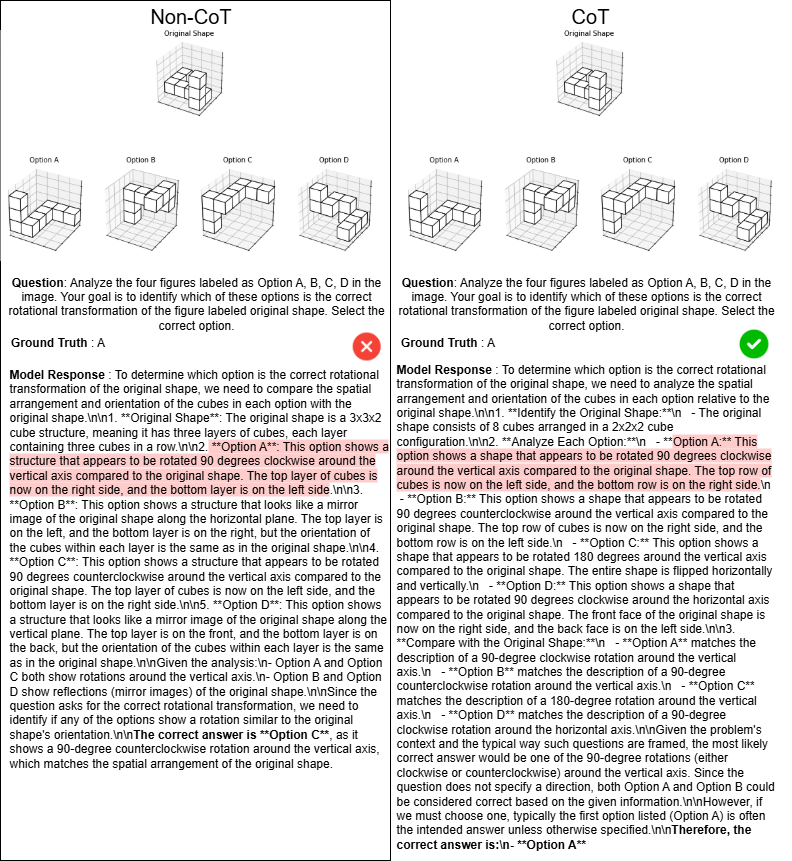}
    \caption{\textbf{Reasoning with prompt variations}: (Left) Response with Non-CoT Prompt, (Right) Response with CoT Prompt, for the same image-question pair. The highlighted section of the reasoning traces shows the superficial shift in reasoning of an option, without any principled justification.}
    \label{fig:cotvncot}
\end{figure}
Our analysis of reasoning traces (Figure~\ref{fig:cotvncot}) highlights two critical failure modes. First, we observe a clear perception error: the models often misinterpret the underlying structure of the figure. For instance, when reasoning about a shape, the model incorrectly encodes it as a 3$\times$2$\times$2 cube, indicating persistent misperception of visual structure. Second, we find a systematic instability in reasoning: altering the prompt does not induce substantive changes in the underlying reasoning process, but instead produces superficial shifts in response orientation. The traces provide no principled justification for why the answer changes, suggesting limited visuo-cognitive grounding and inconsistent reasoning explanations.

\subsection{Similar Answer Selection Propensity}
\label{app:similaranswers}

\begin{figure}[H] 
    \centering
    \includegraphics[width=\columnwidth]{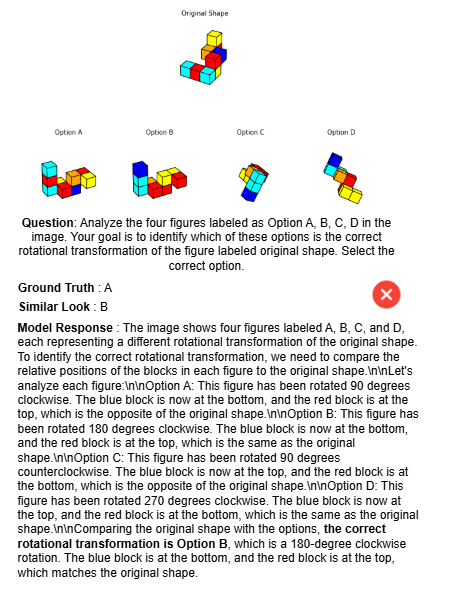}
    \caption{\textbf{Similar Answer Selection reasoning } : The figure illustrates cases where a distractor option closely resembles the correct answer. The model doesn't perform the necessary multistep reasoning and final disambiguation. The \textbf{bold text} shows the final answer selection }
    \label{fig:similaranswer}
\end{figure}

\begin{table}[htpb]
\resizebox{\columnwidth}{!}{%
\begin{tabular}{c|ccc|ccc}
\toprule
 & \multicolumn{3}{c|}{\textbf{Mental Transformation}} & \multicolumn{3}{c}{\textbf{Paper Folding}} \\
\textbf{Models} & Correct & Similar & Incorrect & Correct & Similar & Incorrect \\
\midrule
InternVL2.5     & 0.20 & 0.32 & 0.48 & 0.21 & 0.24 & 0.50 \\
InternVL3.5     & 0.34 & 0.30 & 0.36 & 0.18 & 0.34 & 0.48 \\
Qwen2.5-VL-7B   & 0.34 & 0.18 & 0.48 & 0.18 & 0.24 & 0.58 \\
Qwen2.5-VL-3B   & 0.28 & 0.24 & 0.48 & 0.25 & 0.25 & 0.50 \\
Qwen2.5-VL-32B  & 0.36 & 0.24 & 0.40 & 0.31 & 0.22 & 0.47 \\
LLaVa           & 0.26 & 0.20 & 0.54 & 0.15 & 0.20 & 0.65 \\
Idefics         & 0.28 & 0.22 & 0.50 & 0.30 & 0.25 & 0.45 \\
\bottomrule
\end{tabular}%
}
\caption{ \textbf{ Propensity for Similar Answer Selection} : The table reports the proportion of times the model selects the correct, incorrect, and visually similar (distractor) options for each task.}
\label{tab:6x2_sub}
\end{table}

In tasks where one of the distractor options closely resembles the correct answer, successful solving requires multi-step reasoning to disambiguate between the two. As shown in the figure \ref{fig:similaranswer}, the model seldom engages in such multi-step reasoning and final disambiguation step and instead falls to the wrong option uniformly as shown in Section \ref{app:model_consensus}, ultimately leading to systematic errors.

\subsection{Domain Knowledge Dependence}
\label{app:conceptmisunderstanding}

\begin{figure}[H] 
    \centering
    \includegraphics[width=\columnwidth]{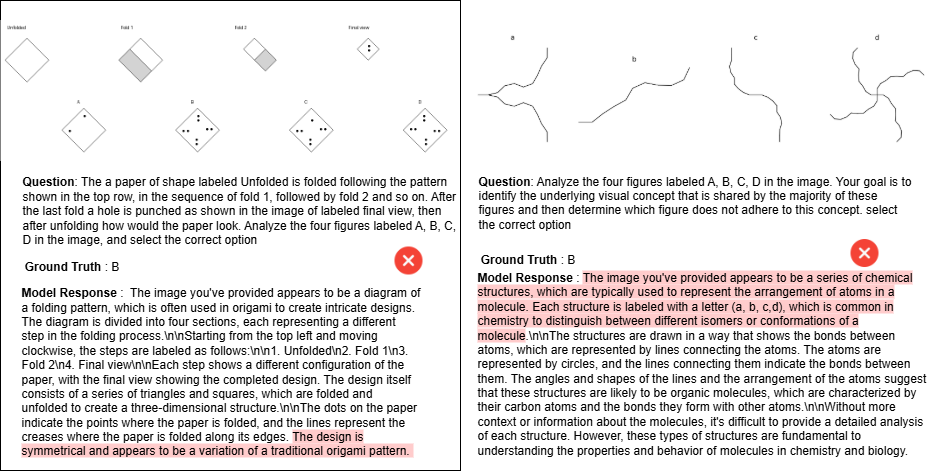}
    \caption{\textbf{ Concept Misunderstanding} : This showcases two examples of model failure due to applying inappropriate domain knowledge. The model misinterprets abstract symmetric structures as "chains of molecules" instead of reasoning about their geometric properties and misinterprets paper folding as a task related to Origami, leading to an incorrect conclusion.
    }
    \label{fig:conceptmiss}
\end{figure}
For many tasks, the models try to retrieve an answer from its domain of knowledge, which leads to error in understanding the underlying concepts of perception. Like for symmetric structures, it infers them as chain of molecules rather than trying to understand their underlying concepts, Fig \ref{fig:conceptmiss}. Models shows poor performance in understanding the underlying concept due to their heavy reliance on domain knowledge based interpretation. Like for symmetric figures, it thinks these are chains of molecules.

\subsection{Causal Intervention}
\label{app:causalintervention}

Causal intervention for circuit discovery in MLLM entails selectively ablating model components—such as attention heads, residual streams, or activations—to assess their functional role in a target task \citep{narrowgate2024, autodiscovery2024, sharedcircuits2024}. The primary objective of knockout based intervention is to causally identify subnetworks or circuits within the language tower that are responsible for specific behaviors or multimodal communication, by observing the disruption of model outputs when these components are ablated \citep{autodiscovery2024, sharedcircuits2024, narrowgate2024}.

Recent works have demonstrated the utility of knockout interventions for mechanistic discovery in MLLM. For instance, \citet{narrowgate2024} employs attention knockout to localize circuits mediating image-to-text transfer. Similarly, \citet{autodiscovery2024} utilizes cross layer attribution followed by activation knockout to validate discovered circuits, and \citet{sharedcircuits2024} investigates the causal impact of ablation on shared subnetworks. These methodologies collectively establish knockout intervention as a central paradigm for causal interpretability in MLLM.

Motivated by these techniques, we performed knockout interventions aligned with current methodology to determine whether circuits exist for a specific task within Qwen-7B. In contrast to prior findings, our knockout experiments did not reveal any functional circuit whose ablation affected model performance on the tested task. To verify this observation, we performed similar intervention across intra-family(Qwen-3B, Qwen-7B)y and across models (Qwen-7b and LlaVa-7B) for all the tasks. This negative result suggests that, for the task investigated, no distinct causal circuits could be isolated within the model using this approach.

\begin{figure}[H] 
    \centering
    \includegraphics[width=\columnwidth]{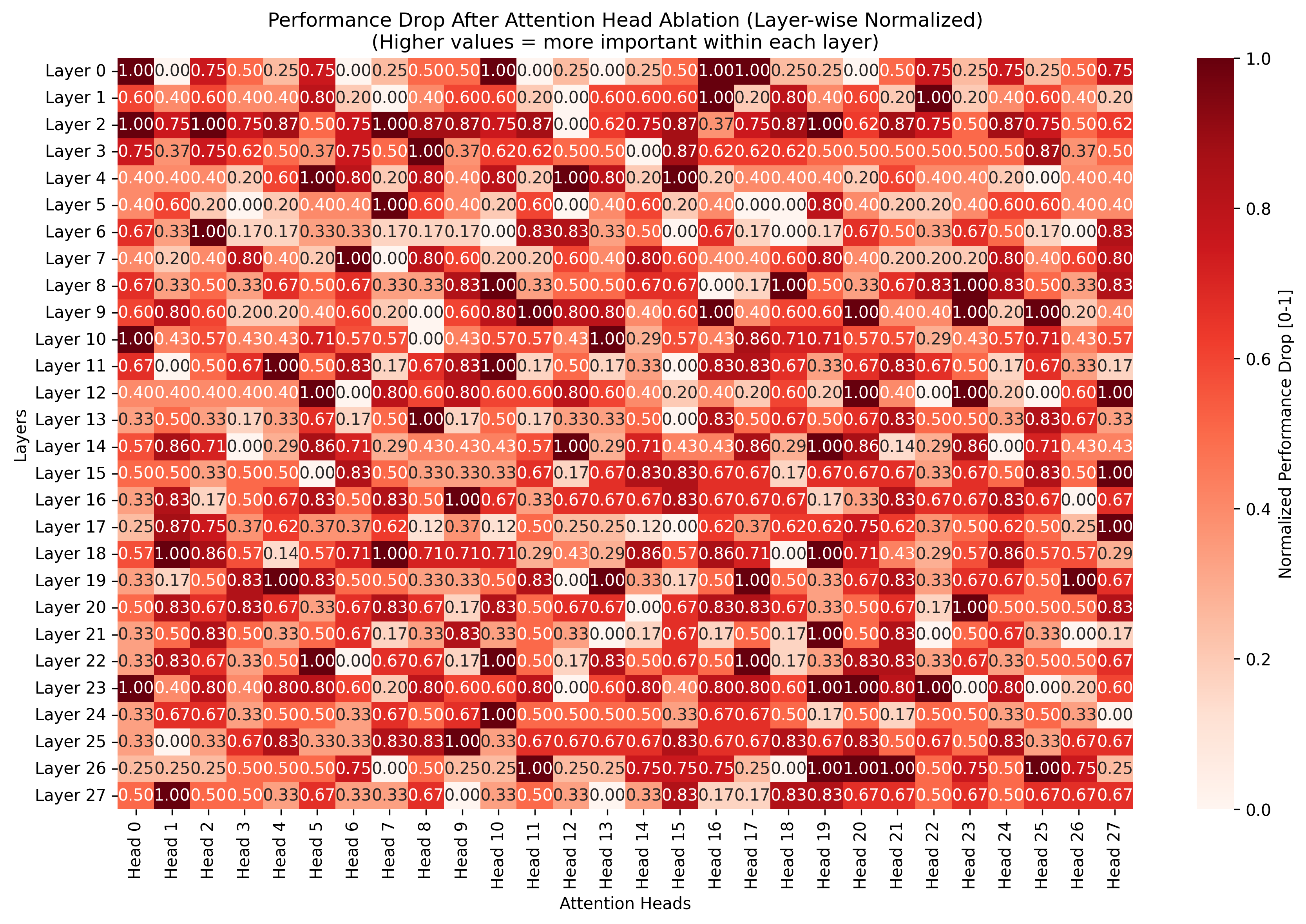}
    \caption{\textbf{Performance Variation from Attention Head Knockout}: A heatmap from a causal intervention experiment on the Qwen-7B model indicates that disabling individual attention heads did not cause a significant performance drop, suggesting the model lacks a specific, localized circuit for the Mental Composition task. }
    \label{fig:cin}
\end{figure}

\subsection{Model Consensus: Role of Distractors}
\label{app:model_consensus}
A key concern in the evaluation of these tasks is whether models are genuinely failing to reason about the underlying transformation, or merely being misled by distractors that resemble the correct answer. 

Formally, our null hypothesis states that when a model answers incorrectly, its choice is \emph{uniformly random} among the three available wrong options. The alternative hypothesis is that models exhibit a statistically significant bias toward the specified distractor.

we performed a $\chi^2$ goodness-of-fit test on the distribution of incorrect responses with respect to the designated distractor option. 
For the \textbf{Mental Transformation (MT)} subset, The chi-squared test yielded $\chi^2 = 0.59$, $p = 0.44$.  For the \textbf{Paper Folding (PF)} subset, The chi--squared statistic was $\chi^2 = 0.42$, $p = 0.52$.  

\paragraph{Human Performance.} To validate the distractor effects on human performance, we performed the same $\chi^2$ goodness-of-fit test on the distribution of incorrect responses with respect to the designated distractor option. For the \textbf{Mental Rotation Test (MRT)} subset, The chi-squared test yielded $\chi^2 = 3.21$, $p = 0.073$.  For the \textbf{Paper Folding (PF)} subset, The chi--squared statistic was $\chi^2 = 6.12$, $p = 0.014$. This indicates a trend toward distractor concentration among the wrong options.

In both cases, the $p$-values are far above the conventional significance threshold ($\alpha = 0.05$), meaning we \emph{fail to reject the null hypothesis}. This indicates that models are not disproportionately attracted to the annotated distractors. Instead, their errors appear uniformly spread across all incorrect alternatives. 

\paragraph{(ii) Per-item exact tests.}
To check whether any individual question disproportionately attracted errors to its distractor, we ran an \emph{exact binomial test} per item. This test compares the observed fraction of distractor errors against the null $1/3$ baseline. A small $p$--value would indicate that, for that item, models were systematically biased toward (or away from) the distractor.  

In both tasks, while a few items reached uncorrected $p<0.05$, none survived Holm-Bonferroni correction for multiple comparisons (PF: $5/50$ uncorrected, $0/50$ corrected; MRT: $3/25$ uncorrected, $0/25$ corrected). Thus, no item showed a reliable per-item distractor effect.

\paragraph{(iii) Mixed-effects logistic regression with clustered inference.}
Because responses to the same item are not independent, we fit a logistic generalized estimating equation (GEE) with clustering by question. This model tests whether the log-odds of choosing the distractor differ from the null logit $(1/3)$ baseline, while accounting for within question correlation.  

If the intercept were significantly different from zero, it would indicate a systematic shift toward (or away from) distractors across all items. In practice, both tasks showed non significant results. 
confirming the absence of such a bias.

\begin{center}
\begin{tabular}{lcccc}
\toprule
Task & $\hat\beta$ & SE & OR vs.\ null & $p$ \\
\midrule
MRT & $0.156$ & $0.266$ & $1.17$ & $0.556$ \\
PF  & $0.088$ & $0.170$ & $1.09$ & $0.605$ \\
\bottomrule
\end{tabular}
\end{center}

Both tasks show intercepts not different from zero; the estimated odds ratios relative to the null (1.00) are close to $1$ and non significant.

Across pooled $\chi^2$ tests, per-item exact tests with multiplicity control, and a mixed effects logistic model that accounts for within item dependence, we find \textbf{no statistical evidence} that models are disproportionately choosing the annotated distractors. Their wrong answers are distributed roughly uniformly across all incorrect options. These results suggest that models are not simply confused by visually similar foils, but may instead reflect a broader challenge in integration cognition with perception.

\subsection{Thought Anchors CoT Annotation Analysis}
\label{app:thoughtanchors}

Adopting the \citet{bogdan2025thoughtanchorsllmreasoning} framework, we categorized sentences in chain-of-thought traces into six reasoning stages.  
Our fine-grained analysis of Qwen-7B on the Mental Transformation task highlights that current MLLM lack the mechanisms required for genuine spatial reasoning. Despite generating detailed chain-of-thought (CoT) traces, the model consistently failed to align its reasoning with perceptual evidence. To systematically analyze the internal reasoning structure of the model, we adopt the framework in \citet{bogdan2025thoughtanchorsllmreasoning} to automatically labeled each sentence in the chain-of-thought (CoT) trace into one of six categories using an LLM based auto labeling procedure:

\begin{enumerate}
    \item \textbf{problem\_setup}: Parsing or rephrasing the problem statement, often reflecting initial comprehension.  
    \item \textbf{plan\_generation}: Stating or deciding on a plan of action (e.g., outlining steps of reasoning).  
    \item \textbf{option\_analysis}: Analyzing a specific option (A, B, C, or D) in detail with supporting reasoning.  
    \item \textbf{final\_answer\_emission}: Explicitly providing the predicted answer or sentences directly leading to the answer.  
    \item \textbf{self\_checking}: Verifying prior steps, double checking logic, or expressing re-confirmation of reasoning.  
    \item \textbf{unknown}: Reserved for sentences that do not fit any of the above, including purely stylistic or filler expressions.  
\end{enumerate}

This categorization enables us to disentangle where failures occur within the reasoning process, whether at the stage of problem comprehension, option level analysis, or final answer selection. 

Each example in our dataset is annotated with the ground-truth transformation and the correct option. This allows us to probe not only whether the model predicted the correct option, but also whether its reasoning steps were aligned with the ground truth transformation.

\paragraph{Near chance performance.} The model achieved an overall accuracy of 32.2\%, only marginally above random guessing among four options. In contrast, human participants on comparable tasks reliably achieve accuracies above 80\%. This gap underscores a fundamental inability to simulate mental transformations.

\paragraph{Axis-specific biases.} Performance varied strongly by the axis of rotation: 28.2\% for Y-axis, 32.4\% for X-axis, and 35.7\% for Z-axis. The samples generated were uniformly sampled across all the axis of rotations.Such anisotropy is inconsistent with human visuo-spatial reasoning, where performance is relatively robust across axes. This suggests that the model relies on superficial 2D heuristics rather than constructing flexible 3D representations.

\paragraph{Mis-binding of reasoning and answers.} In 61.1\% of cases, the model’s intermediate reasoning correctly described the ground truth transformation, but failed on two accounts \textbf{(1)} the final predicted option was incorrect. \textbf{(2)} The rotation angle across the axes were incorrect or misaligned. This indicates a systematic \emph{mis-binding error}: the model can verbalize the correct transformation but fails to ground it in the corresponding visual candidate, indicating a loose coupling between linguistic reasoning and visual perception.

\paragraph{Summary.} Together, these analyses suggest that current MLLMs exhibit limited evidence of embodied visuo-cognitive processes required for these tasks. Rather than performing internal perceptual transformations, they rely on shallow symbolic heuristics, leading to systematic and structured errors in mental rotation and transformation tasks.
For Qwen-7B, overall accuracy was 32.2\%, with 61.1\% of reasoning steps correctly describing the transformation but yielding incorrect final answers: a \textit{mis-binding} failure between verbal reasoning and visual grounding.  
Axis-specific results reveal anisotropy across rotation axes, reinforcing the lack of cognitive representation of perception.

\begin{table}[t]
\centering
\caption{Overall performance and reasoning--answer consistency on Mental Transformation (Qwen-7B).}
\label{tab:mt-overall}
\begin{tabular}{lcc}
\toprule
Metric & Value \\
\midrule
Overall Accuracy & 32.2\% \\
Mis-binding  & 61.1\% \\
\bottomrule
\end{tabular}
\end{table}

\begin{table}[t]
\centering
\caption{Accuracy by dominant rotation axis.}
\label{tab:mt-axis}
\begin{tabular}{lcc}
\toprule
Dominant axis & \# Items & Accuracy \\
\midrule
X & 68 & 32.4\% \\
Y & 39 & 28.2\% \\
Z & 42 & 35.7\% \\
\bottomrule
\end{tabular}
\end{table}

\begin{table}[t]
\centering
\caption{Accuracy by number of active axes in the ground truth transformation.}
\label{tab:mt-numaxes}
\begin{tabular}{lcc}
\toprule
\# Active axes & \# Items & Accuracy \\
\midrule
1 axis  & 87 & 29.9\% \\
2 axes  & 26 & 30.8\% \\
3 axes  & 36 & 38.9\% \\
\bottomrule
\end{tabular}
\end{table}

\subsection{Effect of Prompt Optimization on Performance}
\label{app:promptopt}

We further examined whether performance limitations could be attributed to prompt ambiguity or poor phrasing by applying the framework in \citep{agarwal2024promptwizard}, which iteratively refines instructions and examples through a feedback driven critique and synthesis process. For this experiment, we generated three optimized prompt variations for the Qwen2.5-VL model and evaluated them across representative tasks from our benchmark. Table~\ref{tab:promptwiz} reports the comparison between the baseline prompt and optimized variations.

While the optimized prompts yielded small but consistent gains across tasks (e.g., +0.08 on Visual Conceptual Slippage, +0.08 on Dynamic Structural Correspondence, +0.07 on Mental Transformation, and +0.06 on Hierarchical Pattern Equivalence), the overall improvements remained modest. These differences, though positive, do not substantially alter the performance profile of the model. 

This suggests that the observed errors cannot be explained away as artifacts of ambiguous prompt wording. Instead, the persistence of core error patterns across both baseline and optimized prompts indicates that the primary bottleneck lies in the model’s inherent reasoning limitations rather than surface level prompt design. Thus, prompt optimization serves to confirm that the challenges exposed by our benchmark are fundamentally model driven rather than prompt driven.

This observation reinforces our conclusion that the benchmark exposes genuine deficiencies in visuo-cognitive reasoning, rather than artifacts of prompt design.

\begin{table}[t]
\centering
\caption{\textbf{ Prompt Optimization }.This table compares the performance of Qwen 2.5-VL:7B with a baseline prompt versus an optimized version on four tasks. The modest improvements demonstrate that while better phrasing helps, it does not fix the core limitations of the model. All deltas \textless 0.10 absolute.}
\label{tab:promptwiz}
\begin{tabular}{lcccc}
\toprule
Prompt & VCS & DCS & MRT & HPE \\
\midrule
Baseline & 0.30 & 0.32 & 0.35 & 0.24 \\
Variation  & 0.38 & 0.40 & 0.42 & 0.30 \\
\bottomrule
\end{tabular}
\end{table}




\section{More about \textsc{Mind's Eye}}
\label{app:benchmark}

\begin{table*}[t]
\caption{\textbf{Closest benchmarks vs.\ \textsc{Mind's Eye} along diagnostic axes} : A comparative evaluation of \textsc{Mind's Eye} against other benchmarks on key diagnostic criteria such as Parametric Control, Distractor Quality, and the presence of a Human Baseline. It highlights the unique features that make \textsc{Mind's Eye} a more controlled and diagnostic tool for assessing fluid intelligence. 
\ding{51}=explicit support; \ding{114}=partial; \ding{55}=absent.}
\resizebox{\textwidth}{!}{
\begin{tabular}{lccccccc}
\toprule
Dataset & Parametric & Cognitive & Distractors & No Knowledge & Format & Multi-Pass \\
 & Control & Factor (ART) & Keyed to Confounds & Reliance & (MCQ/Open) &  Human Evaluation \\
\midrule
CLEVR-like & \ding{51} & \ding{114} (Abstraction) & \ding{114} & \ding{51} & Open & \ding{55} \\
Bongard-LOGO & \ding{51} & \ding{51} (Abstraction/Relation) & \ding{114} & \ding{51} & MCQ & \ding{55} \\
RPM (RAVEN/I-RAVEN) & \ding{114} & \ding{51} (Abstraction/Relation) & \ding{114} & \ding{51} & MCQ & \ding{55} \\
Mega-bench (MMMU/SEED/...) & \ding{55} & \ding{114} (Mixed) & \ding{55} & \ding{55} & Mixed & \ding{55} \\
\rowcolor{green!20}
\textbf{Mind's Eye (ours)} & \textbf{\ding{51}} & \textbf{\ding{51} (A/R/T)} & \textbf{\ding{51}} & \textbf{\ding{51}} & \textbf{MCQ} & \textbf{\ding{51}} \\
\bottomrule
\end{tabular}}
\label{tab:rw-contrast}
\end{table*}

We construct the \textsc{Mind’s Eye} benchmark by procedurally generating eight families of visuospatial reasoning tasks. Each family implements a well-defined cognitive operation (rotation, folding, composition, abstraction, etc.) and produces itemized question–answer pairs with explicit metadata (answer key, violation type, difficulty). Table~\ref{tab:task_controls} summarizes the controllable parameters, visual layout, and answer annotations for all tasks.

\begin{table*}[t]
\centering
\resizebox{\textwidth}{!}{
\begin{tabular}{l|p{4cm}|p{4.5cm}|p{4cm}}
\toprule
\textbf{Task} & \textbf{Controlled Parameters} & \textbf{Image Layout} & \textbf{Answer Key / Annotation} \\
\midrule
Visual Conceptual Slippage & Concept type (spacing, alignment, number, enclosure, symmetry, topology, border, hollowness, word symmetry), variation seeds & $2 \times 3$ grid of six labeled panels (A–F), one violates the rule & Index of violating panel; concept label; violation description \\
Visual Relation Abstraction & Shape attributes (convexity, symmetry, straight lines, angles, closure, regularity), positive vs.\ negative sets & $2 \times 3$ grid of six figures (A–F), one negative embedded among positives & Label of negative sample; reasoning string decomposed from attributes \\
Mental Transformation & Shape identity, difficulty level (single vs.\ multi-axis rotation), cube colors & Top row: ``Original'' 3D polycube; bottom row: four rotated candidates (A–D) & Correct option among A–D, rotation angles, difficulty tag \\
Mental Composition & Net type (cube, tetrahedron, prism, pyramid, cone, etc.), color assignments, difficulty (easy vs.\ hard nets) & Left: 2D net; right: four 3D candidate solids (A–D) & Correct option matching the folded solid; net/shape pair metadata \\
Paper Folding & Paper polygon size, fold sequence (V/H, diagonal), number and placement of punched holes & Top row: folding sequence; bottom row: four unfolded candidates (A–D) & Correct option label; fold sequence metadata \\
Dynamic Structural Correspondence & Shape type (triangle, square, pentagon, hexagon, diamond), transformation pair from library (rotate, shear, pulsate, bounce, etc.), time steps & $2 \times 4$ grid: first row shows transformation sequence; second row shows four candidate continuations (A–D) & Correct continuation label (fifth frame) with transformation description \\
Symmetric Structures & Symmetry type (vertical, horizontal, rotational), path complexity & $1 \times 4$ grid of line drawings (A–D), three symmetric and one asymmetric & Asymmetric label; annotation ``lacks symmetry'' \\
Hierarchical Pattern Equivalence & Structure type (nested circles, fractal trees, Sierpinski, L-system, etc.), violation injection & $2 \times 2$ grid of hierarchical drawings (A–D), one random violation & Label of violating structure; hierarchical function metadata \\
\bottomrule
\end{tabular}}
\caption{\textbf{Overview of task generation} : The technical blueprint for the benchmark, detailing the specific controlled parameters, image layout, and answer annotations for each of the eight procedurally generated tasks.
}
\label{tab:task_controls}
\end{table*}

\paragraph{Visual Conceptual Slippage.}
We adapt classical “odd-one-out’’ paradigms to probe sensitivity to abstract visual relations. Each item draws six panels arranged in a $2\times3$ grid. Five panels conform to a chosen concept (e.g., equidistant spacing, global symmetry, enclosure of one shape by another).Exactly one panel is designated as violating the concept. For word-symmetry items, a random uppercase string is rendered and mirrored to induce or break bilateral symmetry. Controlled parameters include concept type, variation seeds, and (for word-symmetry) word length. Random seeds are set to ensure reproducibility. The metadata records the violating option, the concept type, and, in word trials, the sampled word.

\paragraph{Visual Relation Abstraction.} 
Visual Relation Abstraction items follow the Bongard problem style as in \citep{nie2020bongard}. Using curated shape attributes (e.g., convexity, line crossings, polygonal regularity), we generate six figures: five positives sharing an attribute and one negative. Images are arranged in a $2\times3$ grid with randomized positions. The annotation records the negative label and a decomposed textual reason string (e.g., “others are convex closed shapes; this one is not”).

\paragraph{Mental Transformation.} Mental Transformation tests are generated from polycube assemblies. Building on the mental rotation subtask introduced in \cite{mindthegap2025}, we extend it along an additional reasoning dimension to evaluate the model’s capacity for multistep reasoning as well. Each item shows a 3D “Original Shape’’ above four candidate rotations. Controlled factors are (i) shape identity, (ii) difficulty (single-axis vs.\ multi-axis rotation), and (iii) cube coloring (monochrome vs.\ varied). The correct answer is the candidate that matches the rotated original; metadata includes applied angles and difficulty level.

\paragraph{Mental Composition.}
This task probes net-to-solid reasoning. A 2D net (cube, prism, pyramid, cone, etc.) is rendered alongside four 3D candidate solids. Nets are chosen from a mapping (cube, cuboid, prism, pyramid, cone; harder items also include octahedron, dodecahedron, icosahedron). For easy items, nets are restricted to simple solids with uniform coloring; for hard items, complex polyhedra with confounding colorings are used. The net is drawn in the top-left of a $2\times4$ grid, and candidate solids are rendered in the bottom row with distinct colors. The correct candidate is the folded realization of the net. Annotations store net identity, correct solid, distractors, color assignments, and difficulty.

\paragraph{Paper Folding.}
We simulate folding and hole punching on polygonal sheets. The sheet is a square or hexagon, represented as a polygon with vertices. A sequence of two folds is sampled either from vertical/horizontal reflections or from diagonal reflections. After folding, a single hole is punched at a random valid coordinate inside the polygon. The algorithm recursively unfolds the sheet and computes the mirrored hole positions. The final composite image shows: (i) initial unfolded sheet, (ii) two intermediate folds, (iii) the final folded sheet with hole, and (iv) four candidate unfolded sheets (A–D), one correct and three foils generated by removing, mirroring, or randomizing holes. The annotation records the fold group and the correct label. The task is to infer the unfolded hole pattern. Images show the fold sequence (top row) and four candidate unfolded sheets (bottom row, A–D). The correct option reproduces the true unfolded hole distribution.

\paragraph{Dynamic Structural Correspondence.}
Dynamic isomorphism tasks evaluate extrapolation of geometric motion. Two shapes undergo distinct continuous transformations (e.g., rotate-back-and-forth, bounce, wiggle, pulsate, swirl, shear, compress-and-stretch). The top row shows their trajectories at $t\in\{0.0,0.25,0.5,0.75\}$. The bottom row contains four candidate continuations for $t=1.0$, with one true continuation and three distractors (e.g., using mismatched functions or perturbed times). Parameters control shape identities, transformation pair, and time discretization. Annotations specify the correct continuation and textual explanation of which transformation applied to each shape.

\paragraph{Symmetric Structures.}
This task probes symmetry detection in line drawings. Each item shows four connected-path drawings: three exhibiting a chosen symmetry (vertical, horizontal, or rotational) and one lacking it. We generate random line paths by chaining ten short segments with random turns. Symmetry is imposed by reflection (vertical/horizontal) or rotation of order $k\in\{2,4\}$. The layout is a $1\times4$ grid (A–D), and the answer is the asymmetric panel.

\paragraph{Hierarchical Pattern Equivalence.}
Hierarchical reasoning is tested using recursively defined drawings (nested circles, concentric hexagons, fractal trees, L-systems, Sierpinski gaskets, Pythagoras trees, etc.). Each $2\times2$ grid shows three valid hierarchical constructions and one violation consisting of random disconnected strokes. A random seed per panel ensures reproducible but varied instantiations. Parameters include which hierarchical generator is sampled and the seed for randomness. The correct answer is the violating panel.

\section{Benchmark Design}
\label{app:benchmark_design}
The stimuli in \textsc{Mind's Eye} are generated using scalable vector graphics (SVG) to keep a tight control over the geometric properties of the generated figures. Our design follows established principles from cognitive psychometrics \citep{embretson2013item, de2003theory} and recent best practices in multimodal evaluation \citep{yu2023seed, li2023mmbench}.

\begin{table*}[t]
\centering
\begin{minipage}{0.48\textwidth}
\centering
\rowcolors{2}{gray!20}{white}
\resizebox{\textwidth}{!}{
\begin{tabular}{lcccccc}
\toprule
\textbf{Task} & \textbf{\makecell{S1: \\ Mental \\ Rotation}} & \textbf{\makecell{S2: \\ Folding / \\ Topology}} & \textbf{\makecell{S3: \\ Relational \\ Mapping}} & \textbf{\makecell{S4: \\ Symmetry / \\ Group Actions}} & \textbf{\makecell{S5:\\ Composition / \\ Decomposition  }} & \textbf{\makecell{S6: \\ Slippage / \\ Robustness}} \\
\midrule
MT            & \cmark &  &  & \cmark &  &  \\
PF                    &  & \cmark &  & \cmark &  &  \\
DSCS              & \cmark &  & \cmark &  &  &  \\
HPE         &  &  & \cmark &  & \cmark &  \\
VRA               &  &  & \cmark & \cmark &  &  \\
SS              &  &  &  & \cmark &  &  \\
MC               &  &  &  &  & \cmark &  \\
VCS               &  &  & \cmark &  &  & \cmark \\
\bottomrule
\end{tabular}}
\caption{\textbf{Q–matrix blueprint.} Each task is mapped to a vector of latent visuo-cognitive skills. This matrix operationalizes the benchmark’s construct coverage and supports multi–trait psychometric modeling \citep{de2003theory, embretson2013item}.}
\label{tab:qmatrix}
\end{minipage}
\hfill
\begin{minipage}{0.48\textwidth}
\centering
\resizebox{\textwidth}{!}{
\begin{tabular}{l>{\columncolor{green!15}}ccccc}
\toprule
\textbf{\makecell{Subtasks / \\ Dimensions}} & \textbf{\makecell{Ours}} & \textbf{\makecell{Mind \\ the Gap}} & \textbf{\makecell{Bongard \\ Logo}} & \textbf{Visulogic} & \textbf{\makecell{Bongard \\ Hoi}} \\
\midrule
\multicolumn{6}{l}{\textbf{Abstraction}} \\
VRA & \checkmark & \checkmark &  &  &  \\
HPE  & \checkmark &  & \checkmark &  & \checkmark \\
\midrule
\multicolumn{6}{l}{\textbf{Relation}} \\
DSC            & \checkmark &  & \checkmark &  &  \\
VCS       & \checkmark &  &  &  &  \\
SS       & \checkmark &  &  &  &  \\
\midrule
\multicolumn{6}{l}{\textbf{Transformation}} \\
MT    & \checkmark &  &  &  &  \\
PF     & \checkmark &  &  & \checkmark &  \\
MC           & \checkmark & \checkmark &  &  &  \\
\bottomrule
\end{tabular}}
\caption{\textbf{Comparison of subtask coverage across benchmarks}: Comparison of \textsc{Mind's Eye} to other cognitive reasoning benchmarks. under the Abstraction-Relation-Transformation (A-R-T) framework. }
\label{table:other_dataset_coverage}
\end{minipage}
\caption{ Abstraction: VRA (Visual Relation Abstraction), HPE (Hierarchical Pattern Equivalence). Relation: DSC (Dynamic Structural Correspondence), VCS (Visual Conceptual Slippage), SS (Symmetric Structures). Transformation: MT (Mental Transformation), PF (Paper Folding), MC (Mental Composition).}
\end{table*}

\paragraph{Content blueprint and construct coverage.}
Each task targets a distinct visuospatial construct, e.g. axis–aware 3D rotation (MRT) or relational structure preservation (Analogies). We developed a \emph{q–matrix} mapping items to latent skills, ensuring coverage across multiple reasoning domains while avoiding construct underrepresentation \citep{embretson2013item}. This blueprint is intended to ensure broad cognitive coverage rather than overfitting to a narrow skill domain.

\paragraph{Factorial item generation.}
\label{app:factorial}
To minimize annotation artifacts and superficial shortcuts, we implemented \emph{parametric, factorial generators} for all tasks. Each generator independently randomizes \emph{structural factors} (e.g.\ fold sequence length in Paper Folding, transformation chain length in Dynamic Isomorphism) and \emph{nuisance factors} (rendering styles, color schemes, layout jitter). Orthogonal variation across these factors ensures item variety while balancing distractor plausibility. This design philosophy draws inspiration from adversarial benchmark construction in NLP and vision \citep{nie2020adversarial, zellers2019hellaswag}.

\paragraph{Difficulty calibration.}
\label{app:difficulty_calibration}
Difficulty levels were calibrated both \emph{a priori}, by manipulating structural complexity (e.g.\ rotation angle magnitude, hierarchy depth, color confounders), consistent with psychometric principles of item design where structural manipulations systematically affect item difficulty \citep{embretson1983construct, embretson2013item, ekstrom1976kit, vandenberg1978mental}.

\textbf{Difficulty Calibration via Human Consensus}: To establish empirically grounded difficulty levels for each item, we leverage human performance data collected during our evaluation study (Appendix \ref{humanproto}). Each item was independently evaluated by exactly five randomly sampled participants from our cohort of 30 non-expert adults. We operationalize difficulty through a consensus based framework rooted in inter-annotator agreement principles: items are classified as Easy if all five annotators provide the correct response (perfect agreement, $\kappa$ = 1.0 for that item), Hard if all five annotators respond incorrectly (perfect agreement on failure, $\kappa$ = 1.0) or any one answers correctly, and Medium if annotators exhibit split judgments with 2-3 correct responses (partial agreement, 0.33 $\leq \kappa \leq$  0.67). This approach aligns with established psychometric practice where item difficulty is calibrated against empirical response patterns rather than a priori structural complexity alone \cite{landis1977measurement}. Formally, for item $i$ with human responses $\{r_1, r_2, r_3,r_4,r_5\} \in \{0,1\}$, we assign difficulty d(i) as: d(i) = Easy if $\sum r_j = 5$; Hard if $\sum r_j \leq 1$; Medium otherwise. To quantify the reliability of these assignments, we computed Fleiss' kappa across all items within each task family, yielding moderate to substantial agreement ($\kappa$ = 0.71 across tasks), confirming that our difficulty categories capture stable individual differences rather than measurement noise. This consensus-driven calibration ensures that difficulty labels reflect actual human performance distributions and provides a principled basis for stratified analysis of model performance across varying levels of cognitive demand. The distribution of difficulty levels across our benchmark is: Easy (32\%), Medium (45\%), Hard (23\%), ensuring sufficient representation of all difficulty strata for robust evaluation.

\paragraph{Distractor taxonomy.}
\label{app:distractors}

We construct options via confound keyed templates: 
\emph{Transformation} — For MT the confounding dimensions were mirrored objects,varying color sequence of blocks. For MC the distractors were mirrored folds, parity punches, total punch holes reduced by 1. For MC colors and similar number of faces of the 3D object were the distractors.
\emph{Relation} — For DSC similar transformations applied to opposite shape and similar shapes with different transformation applied were used. For VCS shapes,colours and counts were used for the distractors.
\emph{Abstraction} — superficial feature match of the figures and motif substitution were the distractor's features.
Each item’s options include exactly one ground truth and one confounds sampled from distinct templates to avoid ambiguity and remaining wrong options. Table \ref{tab:task_controls} shows the control parameters for each task more in detail.

\paragraph{Answer encoding and randomization.}
\label{app:encoding}

Each item is a 1{+}4 panel (query + options A–D) except the tasks VCS and VRA. These tasks are 1{+}6 (query + options A-F). Option order is uniformly randomized; keys are uniformly distributed across all options. Unless stated, each subtask contains 100 items (balanced across difficulty bins), yielding 800 items total for the main suite.

\paragraph{Item specification.}
\label{app:item_specification}

We author in SVG and export to PNG at 1024$\times$1024 px (300~DPI) with fixed stroke widths and sans-serif labels; background is uniform. Panels use consistent gutters and margins to minimize layout cues. A schematic is shown in Fig.~\ref{fig:overview}.

\paragraph{Benchmark properties.}
\label{app:benchmark_properties}
The Q-matrix in Table~\ref{tab:qmatrix} specifies the mapping between each benchmark task and the three core processes of fluid reasoning : \textbf{Abstraction}, \textbf{Relation}, and \textbf{Transformation}. In psychometric terms, the Q-matrix operationalizes our construct blueprint \citep{de2003theory,embretson2013item}, serving as an explicit hypothesis about the latent skills each item requires. For example, \emph{Visual Relation Abstraction} is coded purely under Abstraction, while \emph{Hierarchical Pattern Equivalence} loads on both Abstraction and Relation, since it demands generalization of hierarchical patterns and recognition of their structural equivalence. Similarly, \emph{Dynamic Structural Correspondence} and \emph{Symmetric Structures} are placed at the Relation-Transformation intersection, as they require both analogical mapping and mental manipulation of visual forms. Pure Transformation tasks such as \emph{Mental Rotation}, \emph{Paper Folding}, and \emph{Mental Composition} emphasize dynamic visuospatial manipulation without strong abstraction demands.  

This structured mapping justifies that the benchmark covers a broad range of reasoning processes identified as central to fluid intelligence \citep{carroll1993human,schneider2018intelligence,mcgrew2005cattell}. Moreover, it allows us to move beyond raw accuracy by fitting multi trait IRT or cognitive diagnostic models, thereby diagnosing which cognitive processes (A, R, T) different models succeed or fail on. In effect, the Q-matrix both grounds our task design theoretically and provides the statistical scaffold for psychometric calibration and analysis.

\section{Evaluation Setup}
\label{ap:eval_setup}
To ensure a fair comparison across models, all systems are evaluated with identical visual inputs and standardized textual prompts. Since modern MLLMs often generate extended free form outputs, conventional rule based extraction is brittle and prone to errors \citep{vlmevalkit2023,li2024blink,lu2022learnexplainmultimodalreasoning}. Following recent practice \citep{lu2024mathvistaevaluatingmathematicalreasoning,zhang2024mathversedoesmultimodalllm}, we adopt an expert LLM–based evaluation protocol.  

The procedure consists of three stages.
\begin{enumerate}
    \item \textbf{Input Presentation:}  
    The model under evaluation receives both the image and textual question in a fixed prompt template designed to minimize variation across models.  

    \item \textbf{Answer Extraction:}  
    We employ Gemma-3 \citep{gemmateam2025gemma3technicalreport} as the judging model to parse free form answers into concise responses.  
    This method builds on prior work showing that large LLMs can perform semantic normalization of outputs with high reliability \citep{liu2024mmbench,li2024blink}.  

    \item \textbf{Label Standardization:}  
    Extracted responses are mapped to task specific discrete labels (e.g., multiple choice option identifiers or numeric values).  
    Accuracy is then computed against the ground truth key for each of the eight subtasks.
\end{enumerate}

\paragraph{Prompting strategies.}  
Since multimodal reasoning is highly sensitive to prompt design \citep{wei2022cot,kojima2022zerocot}, we explore four prompting strategies designed to elicit cognitive reasoning rather than shallow pattern matching:  
\begin{itemize}
    \item \textbf{Meta-task Framing.}  
    Before presenting a question, the prompt explicitly describes the type of reasoning required. For example:\emph{“This is a mental transformation test. You need to imagine folding or rotating the shape in 3D.”} \emph{“This is a paper folding puzzle. At the end, identify which option shows the holes in the unfolded paper.”}Such framing aligns the model’s reasoning pathway with the intended cognitive faculty, similar to task oriented prompting used in prior cognitive benchmarks \citep{nie2020bongard,zhang2019raven}.
    
    \item \textbf{Step-by-Step Instruction Prompts.} Models are encouraged to reason structurally by decomposing problems:  
    \emph{“First, describe the shapes. Then, identify the transformation (rotation, reflection, folding, symmetry). Finally, choose the answer.”} This mirrors structured reasoning templates shown effective in prior chain-of-thought prompting work \citep{wei2022cot}.
    \item \textbf{Hints via Concept Tags.}  
    To reduce ambiguity about the task type, we prepend task specific tags. For example:  
    \emph{“[Task: Mental Transformation] Which option matches the rotated version of the shape?”} \emph{“[Task: Symmetry Detection] Which figure preserves the symmetry of the original?”} Such concept scaffolding helps models focus on execution rather than inferring task intent, following recent evaluations of role tagged prompting in multimodal reasoning \citep{liu2024mmbench,visulogic2024}.

    \item \textbf{Chain-of-Thought Anchors.}  
    Instead of generic “think step by step,” we provide explicit anchors to guide reasoning stages:  
    \emph{“Step 1: Identify the primitive shapes. Step 2: Detect how they move or fold. Step 3: Eliminate mismatched answers.”}  
    This builds on structured CoT prompting approaches \citep{wei2022cot,kojima2022zerocot} and ensures models engage in interpretable intermediate reasoning rather than shortcutting to an answer.
\end{itemize}

\paragraph{Hardware} : The computational experiments presented in this paper were executed using a server equipped with four NVIDIA RTX A6000 graphics processing units, each providing 48 GB of dedicated memory to support the inferential and analytical demands of the evaluated models.

\paragraph{Closed source} We use the following setup of OpenAI API for evaluation:

\begin{verbatim}
OpenAI model name: o3-2025-04-16
response = client.responses.create(
    model="gpt-o3",
    reasoning={"effort": "medium"},
    input=[
        {
            "role": "user", 
            "content": prompt
        }
    ],
    max_output_tokens=500,
)
\end{verbatim}

\section{Human Evaluation Protocol}
\label{humanproto}
\begin{figure*}[H] 
    \centering
    \includegraphics[width=0.8\textwidth]{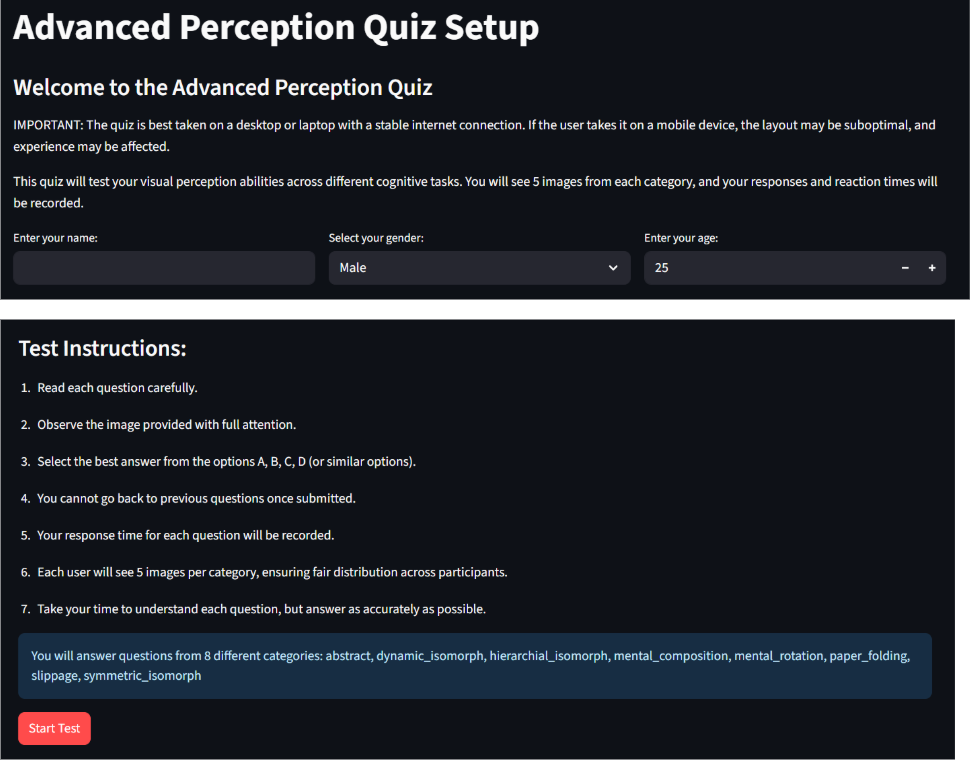}
    \caption{\textbf{Human Performance Benchmarking}:  The web interface used for the human evaluation study, showing the ) Information Collected in the Test and Test instructions provided to all participants}
    \label{fig:spider}
\end{figure*}
To establish a meaningful baseline and to ground our benchmark in psychometric validity, we conducted a controlled human evaluation study. A total of $N=30$ participants were recruited through university mailing lists and professional networks. All participants were adults between the ages of 20 and 40 years, ensuring that the sample represents a cognitively mature population while minimizing potential confounds associated with either adolescent development or age related decline in visuospatial processing. The cohort comprised 17 male and 13 female participants. None of the participants reported any prior expertise with the specific tasks used in our benchmark, and all provided informed consent.

Each participant completed the full battery of eight task families. The tasks were presented in randomized order to control for ordering effects, and items within each task family were also randomized to reduce learning or memorization effects. The evaluation was administered under standardized conditions: participants were given detailed instructions at the beginning of each task type. Responses were collected digitally through a custom interface that mirrors the image based multiple choice format used for multimodal language models.

The total testing time for each participant was approximately 60 minutes, which included both task instructions and the full set of items across all task families. This duration was sufficient to collect reliable performance data while avoiding participant fatigue. The resulting dataset of human responses provides not only an upper bound reference for model comparison but also enables us to quantify item difficulty and discrimination ability through psychometric analysis.

\begin{table}[t]
\centering
\small
\caption{Human participant statistics.}
\label{tab:human-stats}
\resizebox{\columnwidth}{!}{
\begin{tabular}{l r}
\toprule
\textbf{Metric} & \textbf{Value} \\
\midrule
Sample size ($N$)           & 30 \\
Age range (years)           & 20--40 \\
Mean age (years)            & 25.3 \\
Male : Female               & 17 : 13 \\
Prior task expertise        & None (self reported) \\
Recruitment                 & University lists, professional networks \\
Consent                     & Informed consent obtained \\
\bottomrule
\end{tabular}}
\end{table}


\section{Prompt Style Performance}

The performance under four different prompting strategies was evaluated to understand model sensitivity to instructions. Detailed results are presented for Hint-Based prompting (Table \ref{tab:results_hint}), Elimination-Based prompting (Table \ref{tab:results_eliminate}), Meta-Task prompting (Table \ref{tab:results_meta_task}), and Step-by-Step prompting (Table \ref{tab:results_sbs}).

\label{app:prompt_style_performance}
\begin{table*}[t]
\centering
\rowcolors{2}{gray!20}{white}
\resizebox{0.7\textwidth}{!}{
\begin{tabular}{l|cc|ccc|ccc}
\toprule
\hiderowcolors 
 & \multicolumn{2}{c|}{\textbf{Abstraction}} 
 & \multicolumn{3}{c|}{\textbf{Relation}} 
 & \multicolumn{3}{c}{\textbf{Transformation}} \\
 & VRA & HPE 
 & DSC & VCS & SS 
 & MT & PF & MC \\
 & (100) & (100) 
 & (100) & (100) & (100) 
 & (100) & (100) & (100) \\
\midrule
\showrowcolors
Idefics - 8B & 0.24 & 0.27 & 0.35 & 0.26 & 0.16 & 0.34 & 0.27 & 0.27 \\
InternVL3 - 8B & 0.22 & 0.28 & 0.31 & 0.25 & 0.30 & 0.37 & 0.28 & 0.28 \\
LLaMa-3.2 - 11B & 0.22 & 0.28 & 0.31 & 0.26 & 0.23 & 0.30 & 0.25 & 0.28 \\
Llava-1.6-Mistral - 7B & 0.18 & 0.24 & 0.33 & 0.25 & 0.31 & 0.36 & 0.25 & 0.30 \\
Phi3.5-vision-instruct - 8B & 0.21 & 0.26 & 0.32 & 0.24 & 0.30 & 0.33 & 0.26 & 0.30 \\
Qwen-2.5-VL - 7B & 0.25 & 0.26 & 0.30 & 0.26 & 0.15 & 0.39 & 0.21 & 0.22 \\
\bottomrule
\end{tabular}
}
\caption{\textbf{Performance on the task splits using Hint prompts}. Abstraction: VRA (Visual Relation Abstraction), HPE (Hierarchical Pattern Equivalence). Relation: DSC (Dynamic Structural Correspondence), VCS (Visual Conceptual Slippage), SS (Symmetric Structures). Transformation: MT (Mental Transformation), PF (Paper Folding), MC (Mental Composition).}
\label{tab:results_hint}
\end{table*}

\begin{table*}[h]
\centering
\rowcolors{2}{gray!20}{white}
\resizebox{0.7\textwidth}{!}{
\begin{tabular}{l|cc|ccc|ccc}
\toprule
\hiderowcolors 
 & \multicolumn{2}{c|}{\textbf{Abstraction}} 
 & \multicolumn{3}{c|}{\textbf{Relation}} 
 & \multicolumn{3}{c}{\textbf{Transformation}} \\
 & VRA & HPE 
 & DSC & VCS & SS 
 & MT & PF & MC \\
 & (100) & (100) 
 & (100) & (100) & (100) 
 & (100) & (100) & (100) \\
\midrule
\showrowcolors 

Idefics - 8B & 0.23 & 0.28 & 0.34 & 0.27 & 0.14 & 0.34 & 0.27 & 0.28 \\
InternVL3 - 8B & 0.23 & 0.29 & 0.32 & 0.25 & 0.30 & 0.37 & 0.28 & 0.29 \\
LLaMa-3.2 - 11B & 0.23 & 0.29 & 0.32 & 0.26 & 0.23 & 0.31 & 0.25 & 0.29 \\
Llava-1.6-Mistral - 7B & 0.18 & 0.25 & 0.33 & 0.25 & 0.31 & 0.36 & 0.25 & 0.30 \\
Phi3.5-vision-instruct - 8B & 0.22 & 0.27 & 0.33 & 0.25 & 0.30 & 0.33 & 0.26 & 0.30 \\
Qwen-2.5-VL - 7B & 0.27 & 0.28 & 0.31 & 0.26 & 0.22 & 0.25 & 0.32 & 0.37 \\
\bottomrule
\end{tabular}
}
\caption{\textbf{Performance on the task splits using Meta Task prompts}. Abstraction: VRA (Visual Relation Abstraction), HPE (Hierarchical Pattern Equivalence). Relation: DSC (Dynamic Structural Correspondence), VCS (Visual Conceptual Slippage), SS (Symmetric Structures). Transformation: MT (Mental Transformation), PF (Paper Folding), MC (Mental Composition).}
\label{tab:results_meta_task}
\end{table*}

\begin{table*}[h]
\centering
\rowcolors{2}{gray!20}{white}
\resizebox{\textwidth}{!}{
\begin{tabular}{l|cc|ccc|ccc}
\toprule
\hiderowcolors 
 & \multicolumn{2}{c|}{\textbf{Abstraction}} 
 & \multicolumn{3}{c|}{\textbf{Relation}} 
 & \multicolumn{3}{c}{\textbf{Transformation}} \\
 & VRA & HPE 
 & DSC & VCS & SS 
 & MT & PF & MC \\
 & (100) & (100) 
 & (100) & (100) & (100) 
 & (100) & (100) & (100) \\
\midrule
\showrowcolors 

Idefics - 8B & 0.24 & 0.28 & 0.35 & 0.27 & 0.15 & 0.35 & 0.27 & 0.28 \\
InternVL3 - 8B & 0.23 & 0.29 & 0.32 & 0.26 & 0.31 & 0.38 & 0.28 & 0.29 \\
LLaMa-3.2 - 11B & 0.23 & 0.29 & 0.32 & 0.26 & 0.23 & 0.31 & 0.25 & 0.29 \\
Llava-1.6-Mistral - 7B & 0.19 & 0.25 & 0.33 & 0.26 & 0.31 & 0.37 & 0.25 & 0.30 \\
Phi3.5-vision-instruct - 8B & 0.22 & 0.27 & 0.33 & 0.26 & 0.31 & 0.34 & 0.26 & 0.30 \\
Qwen-2.5-VL - 7B & 0.25 & 0.28 & 0.30 & 0.26 & 0.23 & 0.26 & 0.26 & 0.32 \\
\bottomrule
\end{tabular}
}
\caption{\textbf{Performance on the task splits using SBS prompts. Abstraction }: VRA (Visual Relation Abstraction), HPE (Hierarchical Pattern Equivalence). Relation: DSC (Dynamic Structural Correspondence), VCS (Visual Conceptual Slippage), SS (Symmetric Structures). Transformation: MT (Mental Transformation), PF (Paper Folding), MC (Mental Composition).}
\label{tab:results_sbs}
\end{table*}

\begin{table*}[h]
\centering
\rowcolors{2}{gray!20}{white}
\resizebox{\textwidth}{!}{
\begin{tabular}{l|cc|ccc|ccc}
\toprule
\hiderowcolors 
 & \multicolumn{2}{c|}{\textbf{Abstraction}} 
 & \multicolumn{3}{c|}{\textbf{Relation}} 
 & \multicolumn{3}{c}{\textbf{Transformation}} \\
 & VRA & HPE 
 & DSC & VCS & SS 
 & MT & PF & MC \\
 & (100) & (100) 
 & (100) & (100) & (100) 
 & (100) & (100) & (100) \\
\midrule
\showrowcolors
Idefics - 8B & 0.23 & 0.27 & 0.34 & 0.26 & 0.15 & 0.34 & 0.27 & 0.28 \\
InternVL3 - 8B & 0.22 & 0.28 & 0.32 & 0.25 & 0.31 & 0.37 & 0.28 & 0.28 \\
LLaMa-3.2 - 11B & 0.22 & 0.28 & 0.32 & 0.25 & 0.23 & 0.30 & 0.25 & 0.29 \\
Llava-1.6-Mistral - 7B & 0.18 & 0.24 & 0.33 & 0.25 & 0.31 & 0.36 & 0.25 & 0.30 \\
Phi3.5-vision-instruct - 8B & 0.21 & 0.26 & 0.33 & 0.25 & 0.31 & 0.33 & 0.26 & 0.29 \\
Qwen-2.5-VL - 7B & 0.22 & 0.30 & 0.32 & 0.22 & 0.21 & 0.34 & 0.24 & 0.29 \\
\bottomrule
\end{tabular}
}
\caption{\textbf{Performance on the task splits using Eliminate prompts}. Abstraction: VRA (Visual Relation Abstraction), HPE (Hierarchical Pattern Equivalence). Relation: DSC (Dynamic Structural Correspondence), VCS (Visual Conceptual Slippage), SS (Symmetric Structures). Transformation: MT (Mental Transformation), PF (Paper Folding), MC (Mental Composition).}
\label{tab:results_eliminate}
\end{table*}

\section{Prompts}

\begin{figure*}[p] 
    \centering
    \includegraphics[width=0.7\textwidth]{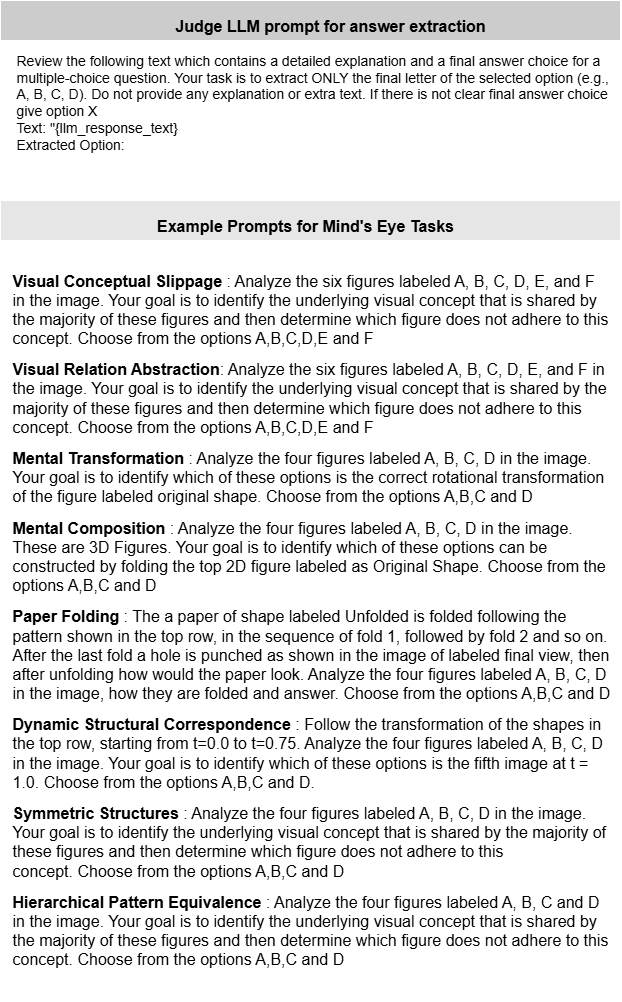}
    \caption{\textbf{Prompts}: (Top) The judge LLM prompt is used for extracting selected options form the free following answers of the model. (Bottom) The question prompt for each task of the benchmark }
    \label{fig:qprompts}
\end{figure*}

\begin{figure*}[p] 
    \centering
    \includegraphics[width=0.9\linewidth]{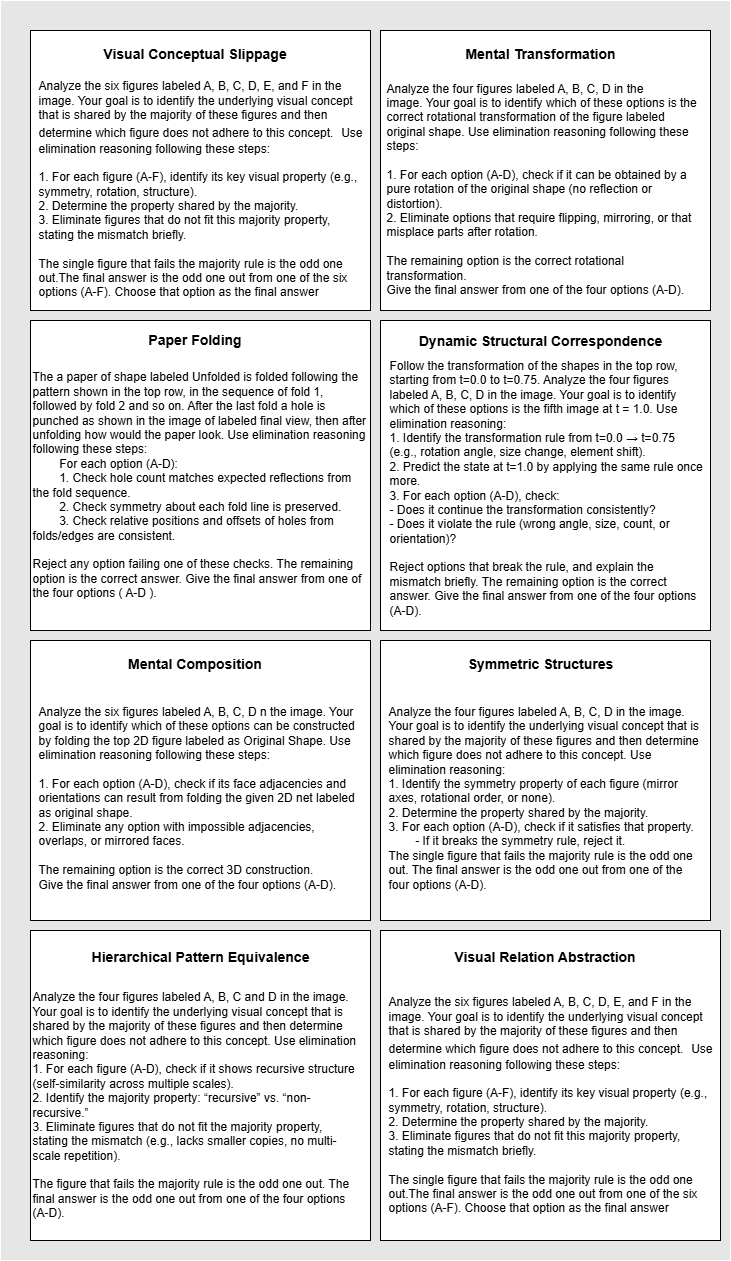}
    \caption{\textbf{Elimination Prompt} for all the tasks.}
    \label{fig:elimination_prompts}
\end{figure*}

\begin{figure*}[p] 
    \centering
    \includegraphics[width=0.9\linewidth]{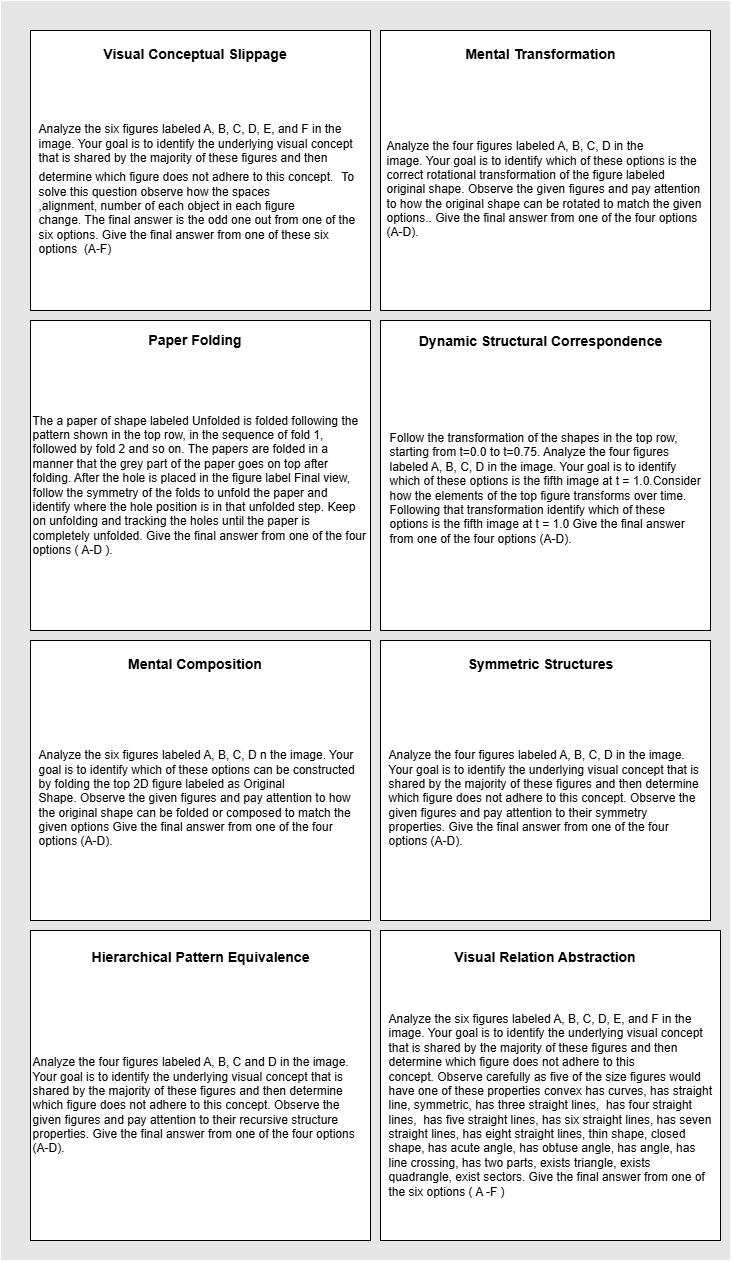}
    \caption{\textbf{Hint Prompt} for all the tasks.}
    \label{fig:hint_prompts}
\end{figure*}

\begin{figure*}[p] 
    \centering
    \includegraphics[width=0.55\linewidth]{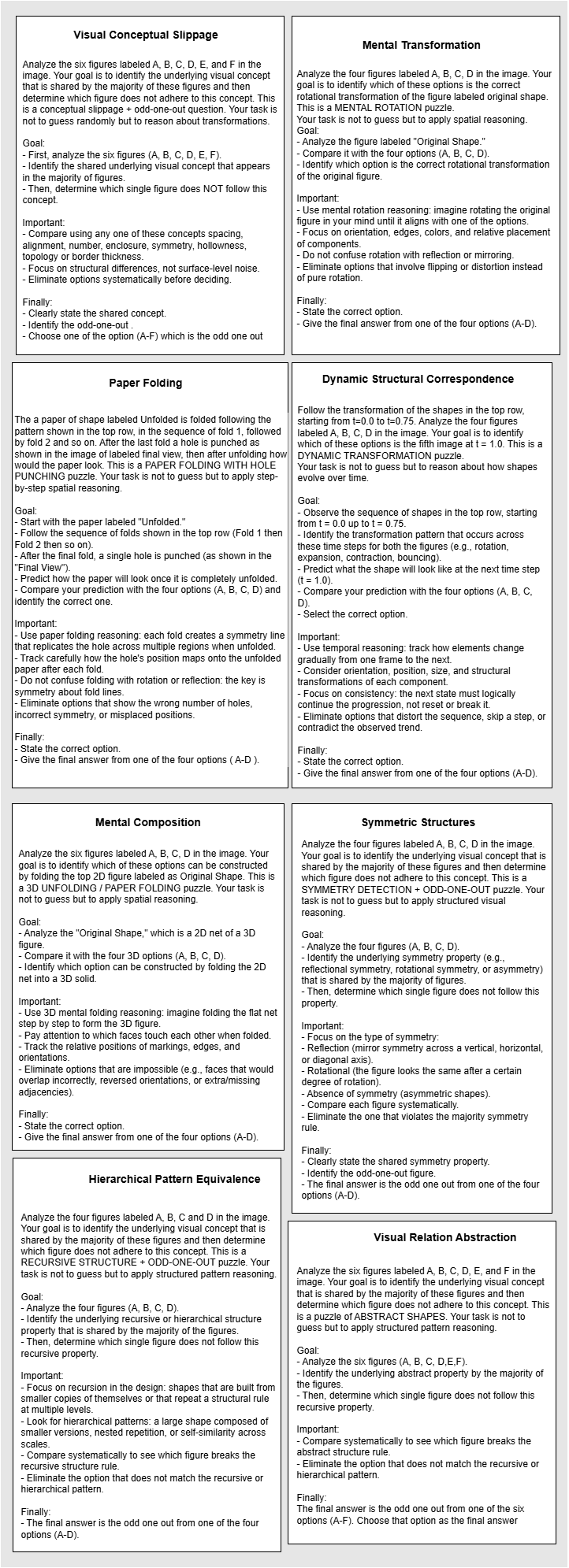}
    \caption{\textbf{Meta Task Prompt} for all the tasks.}
    \label{fig:meta_task_prompts}
\end{figure*}

Following the overview of evaluation strategies in Appendix \ref{ap:eval_setup}, this section presents the specific prompt templates used in our experiments. Figure \ref{fig:qprompts} illustrates the general question structure applied to each task. The detailed templates for our three primary prompting methods—Elimination, Hint-Based, and Meta-Task—are provided in Figures \ref{fig:elimination_prompts}, \ref{fig:hint_prompts}, and \ref{fig:meta_task_prompts}, respectively.

\label{app:prompt_strategies}

\section{Qualitative CoT Analysis}
\label{app:cotqualanalysis}
To further understand the internal reasoning behavior of multimodal large language models, we qualitatively analyzed the reasoning traces produced by GPT-4o across representative tasks from the \textsc{Mind’s Eye} benchmark. Across these tasks, the reasoning traces reveal a consistent pattern: although the models often provide syntactically coherent explanations and occasionally arrive at the correct answer, their reasoning is largely surface level and perceptually driven. Rather than performing the required cognitive operations of perception, such as mental transformation, folding/unfolding, or abstraction of relational structure, the model tends to depend on low level visual heuristics (e.g., color matching, spatial alignment, or visual distinctiveness among options).

Together, these analyses indicate that model reasoning traces often rely on heuristic visual cues rather than systematic cognitive reasoning.

\begin{figure*}[t] 
    \centering
    \includegraphics[width=0.8\linewidth]{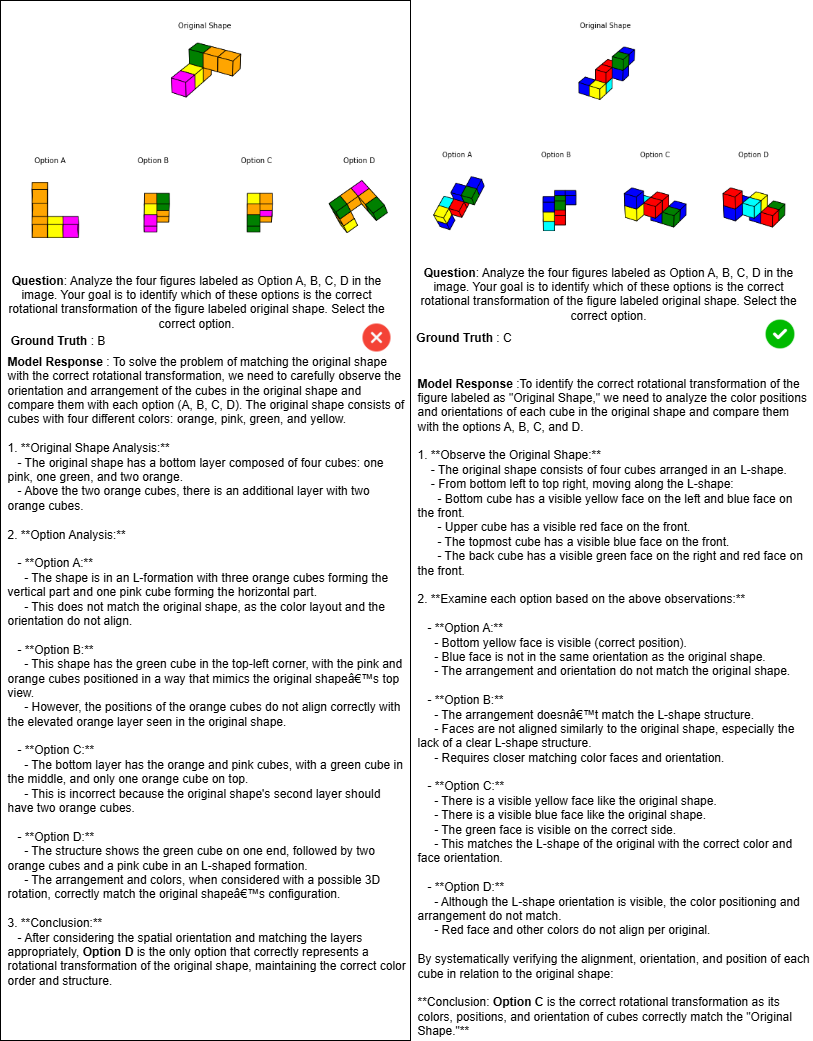}
    \caption{\textbf{Reasoning Trace Analysis for GPT-4o on Mental Transformation Task} : \textbf{(Left)} Incorrect Answer, \textbf{(Right)} Correct Answer. In the \textit{conclusion} of the reasoning traces, the final answer selection is done. Analyzing the reasoning traces for GPT-4o for the Mental Transformation Task (MT) shows that the models are relying on color as heuristic to try to match the options with the original shape. This reasoning traces suggests that the model's functional accuracy may not be consistent with the mechanistic equivalent of the capabilities required to reason about these solutions and reach the correct answer}
    \label{fig:mrtgpt4o}
\end{figure*}

\begin{figure*}[t] 
    \centering
    \includegraphics[width=0.8\linewidth]{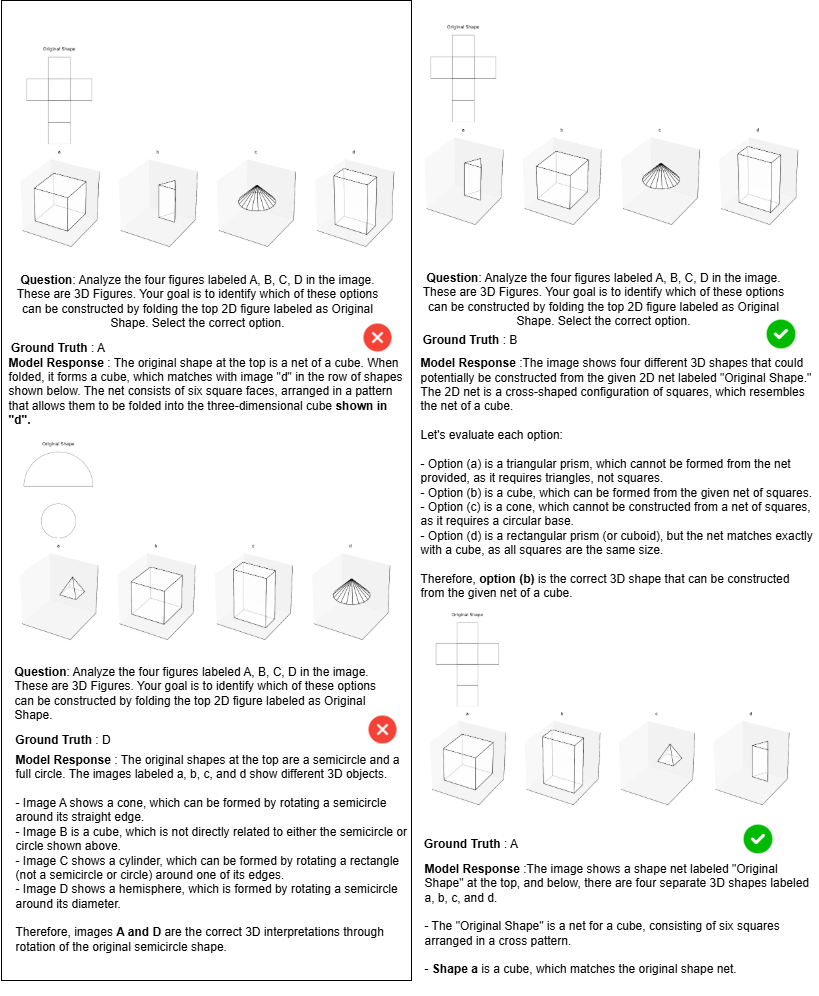}
    \caption{
    \textbf{Reasoning Trace Analysis for GPT-4o on Mental Composition Task}: \textbf{(Left)} Incorrect Answer, \textbf{(Right)} Correct Answer. The reasoning trace shows that when GPT-4o correctly identified the unfolded figure as the cube’s net, it was able to infer the correct folded shape and select the right answer. However, in cases where it failed to recognize the net structure, the model could not mentally simulate the folding operation, leading to incorrect predictions.}
    \label{fig:mcgpt4o}
\end{figure*}

\begin{figure*}[t] 
    \centering
    \includegraphics[width=0.8\linewidth]{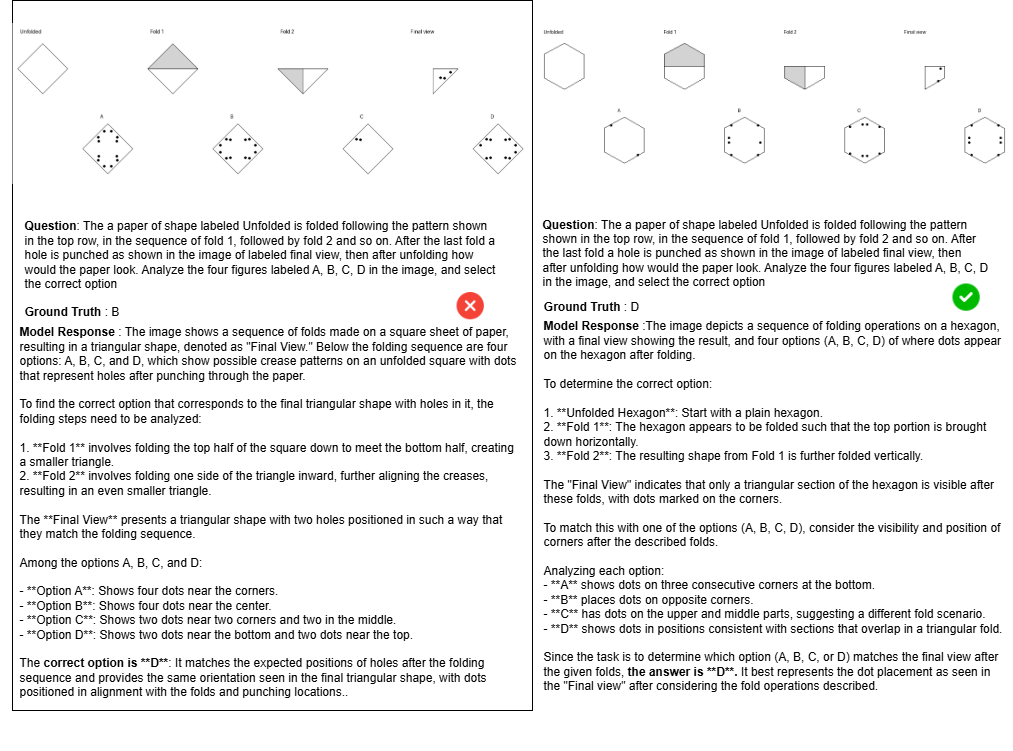}
    \caption{\textbf{Reasoning Trace Analysis for GPT-4o on Paper Folding Task}:\textbf{(Left)} Incorrect Answer, \textbf{(Right)} Correct Answer. Analysis of the reasoning traces shows that while the model correctly identifies how the paper is folded, its option analysis and final answer selection provide no evidence of tracking the holes through the unfolding process. Instead, the model appears to rely on superficial spatial matching between hole positions in the folded and unfolded states, rather than mentally simulating the unfolding operations to derive the correct answer. }
    \label{fig:pfgpt4o}
\end{figure*}

\begin{figure*}[t] 
    \centering
    \includegraphics[width=0.8\linewidth]{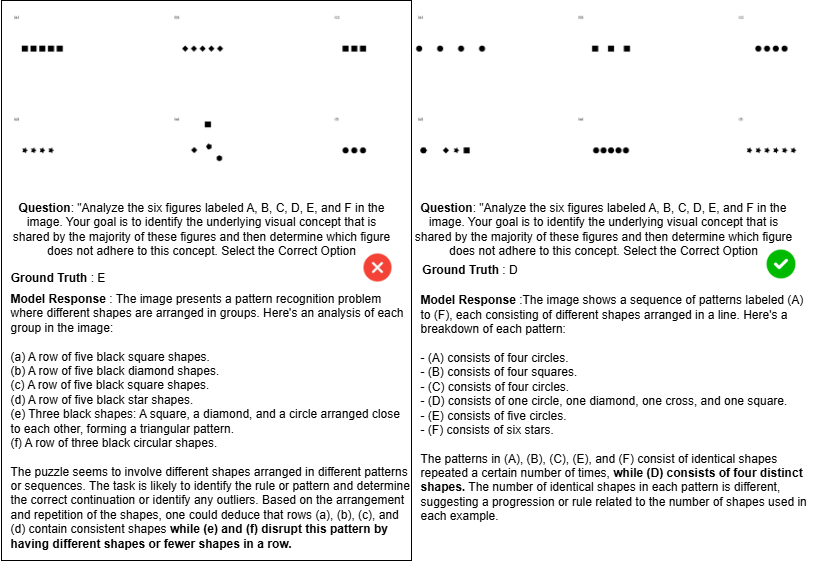}
    \caption{\textbf{Reasoning Trace Analysis for GPT-4o on Visual Conceptual Slippage Task}: \textbf{(Left)} Incorrect Answer, \textbf{(Right)} Correct Answer. Analysis of the reasoning trace suggests that the model relies primarily on superficial visual cues and perceptual artifacts when evaluating the options, rather than grasping the underlying abstract relations shared across the figures. The model arrives at the correct answer only because the correct option exhibits a distinct visual difference, not due to genuine conceptual understanding.}
    \label{fig:slipgpt4o}
\end{figure*}


\section{Caroll's Fluid Intelligence to ART Framework}
\begin{figure*}[ht]
    \centering
    \includegraphics[width=\textwidth]{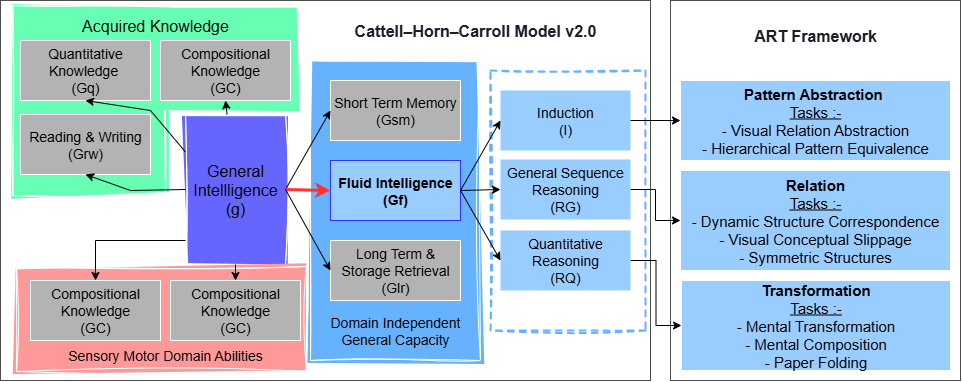}
    \caption{\textbf{Mapping Carroll’s Three Stratum Theory to the Mind’s Eye ART taxonomy}: The figure illustrates how \textbf{fluid intelligence (Gf)} : a core construct in Carroll’s Three Stratum Theory of cognitive abilities—corresponds to the three visuocognitive dimensions evaluated in \textbf{Mind’s Eye}: \textbf{Abstraction}, \textbf{Relation}, and \textbf{Transformation (ART)}. Arrows denote the conceptual linkage from fluid reasoning to these visual faculties, grounding the benchmark’s task design in established psychometric theory.}
    \label{fig:carol_fi_art}
\end{figure*}

Figure \ref{fig:carol_fi_art} illustrates the alignment between the constructs of Fluid Intelligence from the Cattell-Horn-Carroll (CHC) framework and our proposed Abstraction, Relation, and Transformation (A-R-T) taxonomy.

\section{Difficulty Analysis}
\label{app:difficulty}

The performance patterns across the eight cognitive subtasks reveal several critical insights into the visual reasoning capabilities of current MLLMs compared to human performance.

\textbf{Human Performance Sensitivity to Difficulty}: Human participants demonstrate the expected sensitivity to difficulty calibration, with performance systematically declining across difficulty levels. On Easy items, humans achieve accuracies of 0.85-0.95, consistent with the calibration criterion (all 5 annotators correct). Performance drops to 0.55-0.65 on Medium items (2-3 annotators correct), and further declines to 0.10-0.25 on Hard items (0-1 annotators correct). This graded degradation validates our difficulty manipulation and demonstrates that humans engage genuinely with increasing cognitive demands. The decline is particularly pronounced in Transformation tasks (Mental Composition: 0.94 → 0.14) and Abstraction tasks (Hierarchical Pattern Equivalence: 0.92 → 0.18), where spatial manipulation and pattern abstraction become progressively more demanding.

\textbf{The Model Performance Gap}: Both closed-source and open-source models exhibit substantially lower performance compared to humans, with accuracy typically ranging between 0.2-0.5 across tasks. This performance gap is consistent across all eight subtasks, indicating systematic limitations in visual-cognitive reasoning rather than isolated weaknesses. The gap is particularly pronounced in Transformation (Mental Composition, Paper Folding, Mental Transformation) and Abstraction (Visual Relation Abstraction, Hierarchical Pattern Equivalence) dimensions of the ART framework.

\textbf{Flat Model Difficulty Curves: A Critical Divergence}
In stark contrast to humans, both model categories show minimal sensitivity to task difficulty, with performance remaining relatively flat (typically varying by only 0.02-0.08 accuracy points) across Easy, Medium, and Hard conditions. This flat difficulty curve reveals a fundamental limitation: while humans struggle progressively more with harder instantiations of genuine spatial reasoning, models appear unable to perform the core cognitive operations at any difficulty level. The models' consistent low performance (0.20-0.45) regardless of difficulty suggests they lack the foundational visual-cognitive mechanisms required for these tasks. 

\textbf{Closed-Source vs. Open-Source Models}
Closed-source models consistently outperform open-source models across all tasks and difficulty levels, though both remain substantially below human performance and exhibit similarly flat difficulty curves. The performance advantage of closed-source models is most pronounced in Transformation tasks, where they achieve 0.30-0.45 accuracy compared to 0.25-0.35 for open-source models. However, this advantage narrows in Relation tasks (Symmetric Structures, Visual Conceptual Slippage) and Abstraction tasks (Hierarchical Pattern Equivalence, Dynamic Structural Correspondence), suggesting that certain types of visual reasoning present fundamental challenges even for state-of-the-art proprietary models. Critically, neither model category shows the systematic performance degradation across difficulty levels that characterizes human performance.

These results suggest that current MLLMs may lack fundamental visual-cognitive capabilities that humans deploy effortlessly. The divergence between human difficulty sensitivity and flat model performance curves provides compelling evidence that models are not merely worse at these tasks—they are solving them through fundamentally different (and inadequate) mechanisms. While humans engage in genuine visuospatial reasoning that scales with task complexity, models appear to rely on shallow heuristics that fail uniformly across difficulty levels. This suggests that bridging the human-model gap will require architectural innovations that enable true perceptual transformation and cognitive simulation, rather than simply scaling existing approaches.
\begin{figure*}[t] 
    \centering
    \includegraphics[width=\textwidth]{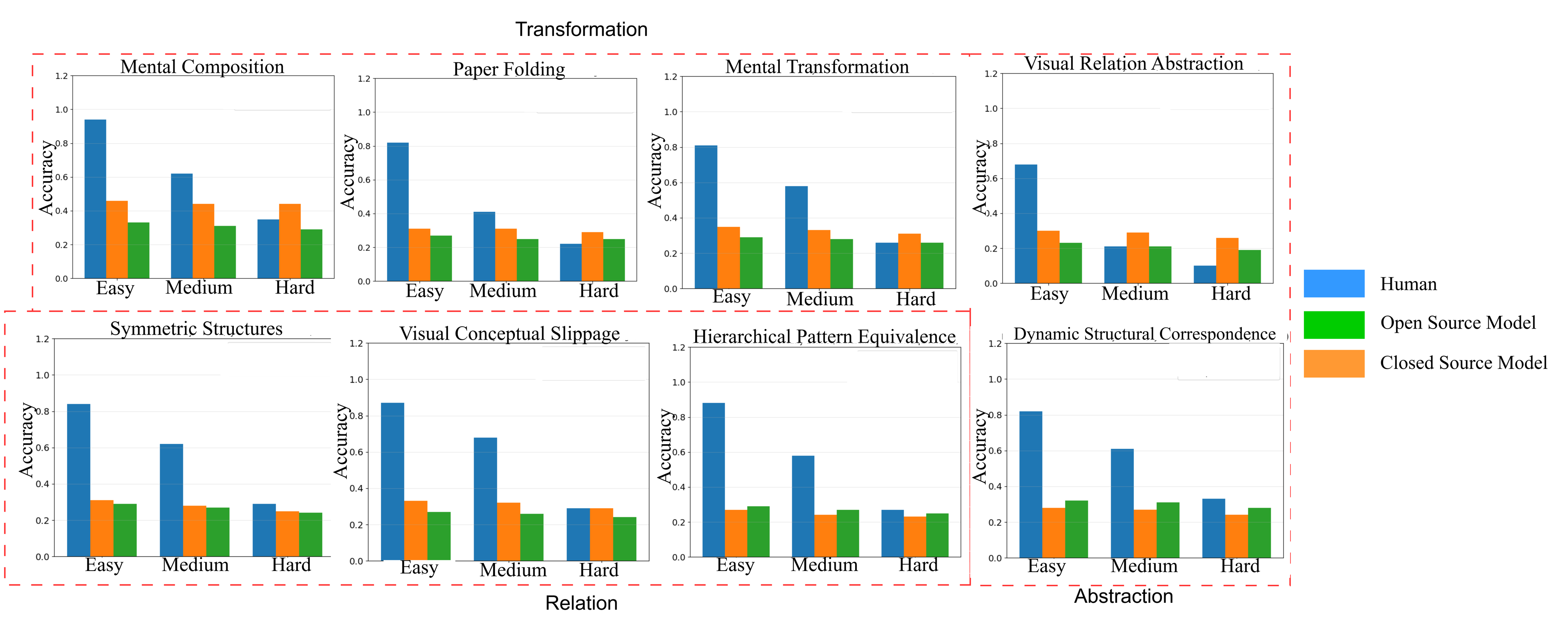}
    \caption{\textbf{Performance across ART dimensions by difficulty level.}Both closed-source and open-source
models struggle across all dimensions with flat difficulty curves (0.20-0.45 accuracy), while human experts maintain robust performance ($>$0.80) across easy, medium, and hard tasks. Each bar represents the macro-average accuracy for a task across all models in that category (see Table~\ref{tab:results}).}
    \label{fig:difficulty_distribution}
\end{figure*}

\end{document}